\documentclass{article}

\PassOptionsToPackage{numbers, compress}{natbib}

\usepackage[final]{neurips_2020}

\usepackage[utf8]{inputenc} 
\usepackage[T1]{fontenc}  
\usepackage{hyperref}    
\hypersetup{hidelinks}
\usepackage{booktabs}    
\usepackage{nicefrac}    
\usepackage{microtype}   

\usepackage{array}
\usepackage{textcomp}
\usepackage{stfloats}
\usepackage{verbatim}
\usepackage{graphicx}

\usepackage{psfrag,epsfig,subfigure,amsmath}
\usepackage{enumerate}
\usepackage{amsfonts,amssymb,amsmath,bm,paralist,theorem,ifthen,color}
\usepackage{listings}

\usepackage{algorithm}
\usepackage{algpseudocode}
\usepackage{amsmath}
\usepackage{multicol}
\usepackage{makecell}
\usepackage{dsfont}
\usepackage{ragged2e}
\usepackage{flushend}
\usepackage{url}

\usepackage{threeparttable}
\usepackage{multirow}

\newtheorem{theorem}{Theorem}
\newtheorem{lemma}{Lemma}

\newtheorem{remark}{Remark}
\newtheorem{assumption}{Assumption}
\newtheorem{property}{Property}
\newtheorem{definition}{Definition}
\newtheorem{observation}{Observation}

\makeatletter
\renewcommand{\maketag@@@}[1]{\hbox{\m@th\normalsize\normalfont#1}}%
\makeatother

\allowdisplaybreaks[4]

\title{Why Batch Normalization Damage Federated Learning \\ on Non-IID Data?}

\author{
Yanmeng~Wang,
Qingjiang~Shi,
and~Tsung-Hui~Chang,~\emph{Fellow,~IEEE}
\thanks{Y. Wang and T.-H. Chang are with the School of Science and Engineering, The Chinese University of Hong Kong, Shenzhen
518172, China, and also with the Shenzhen Research Institute of Big
Data, Shenzhen 518172, China (e-mail: hiwangym@gmail.com, tsunghui.chang@ieee.org).
Q. Shi is with the School of Software Engineering, Tongji University,
Shanghai 201804, China, and also with the Shenzhen Research Institute of Big
Data, Shenzhen 518172, China (e-mail: shiqj@tongji.edu.cn). \emph{(Corresponding author: Tsung-Hui Chang.)}}}

\makeatletter
\def\thanks#1{\protected@xdef\@thanks{\@thanks
    \protect\footnotetext{#1}}}
\makeatother

\begin{document}

\maketitle

\begin{abstract}
As a promising distributed learning paradigm, federated learning (FL) involves training deep neural network (DNN) models at the network edge while protecting the privacy of the edge clients.
To train a large-scale DNN model, batch normalization (BN) has been regarded as a simple and effective means to accelerate the training and improve the generalization capability.
However, recent findings indicate that BN can significantly impair the performance of FL in the presence of non-i.i.d. data.
While several FL algorithms have been proposed to address this issue, their performance still falls significantly when compared to the centralized scheme.
Furthermore, none of them have provided a theoretical explanation of how the BN damages the FL convergence.
In this paper, we present the first convergence analysis to show that under the non-i.i.d. data, the mismatch between the local and global statistical parameters in BN causes the gradient deviation between the local and global models, which, as a result, slows down and biases the FL convergence.
In view of this, we develop a new FL algorithm that is tailored to BN, called \texttt{FedTAN}, which is capable of achieving robust FL performance under a variety of data distributions via iterative layer-wise parameter aggregation.
Comprehensive experimental results demonstrate the superiority of the proposed \texttt{FedTAN} over existing baselines for training BN-based DNN models.
\end{abstract}

\section{Introduction}

\begin{figure}[t]
\centering
\subfigure[\scriptsize{Centralized Learning}]{
\includegraphics[width= 1.6 in ]{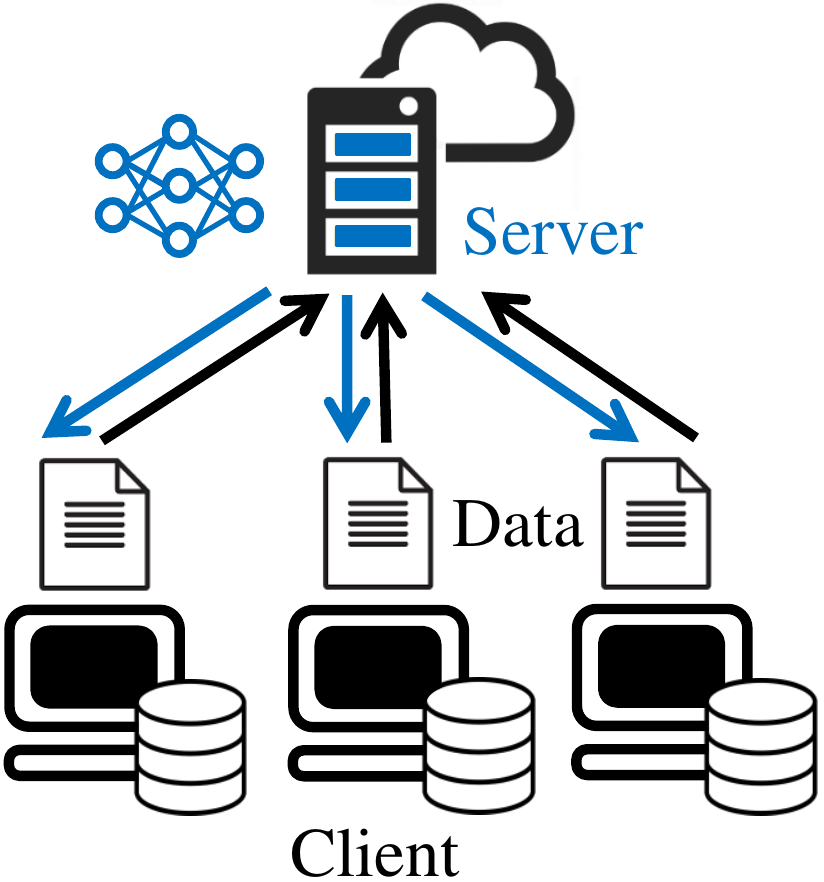}}
\hspace{0.5cm}
\subfigure[\scriptsize{Federated Learning}]{
\includegraphics[width= 1.6 in ]{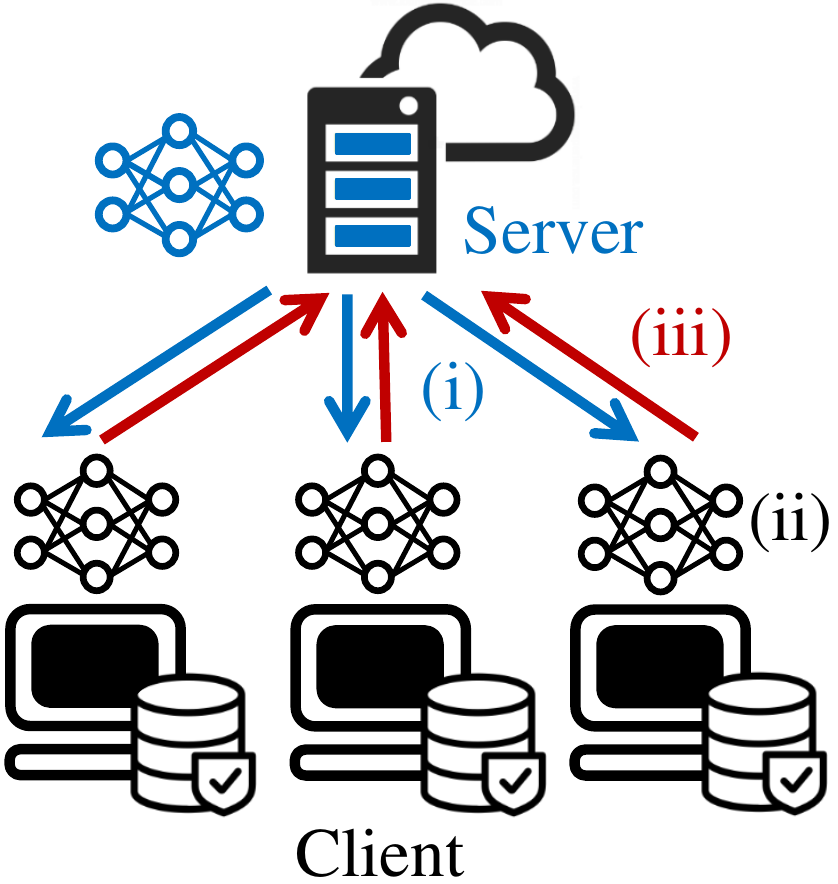}}
\caption{Centralized learning v.s. Federated Learning.}
\label{fig:CL VS FL}
\end{figure}

The traditional centralized learning systems, as illustrated in Fig. \ref{fig:CL VS FL}(a), involves collecting data samples from edge clients and generating a global dataset at the server.
This creates a huge transmission cost as well as privacy concerns.
In order to overcome these issues, federated learning (FL) in Fig. \ref{fig:CL VS FL}(b) has been proposed as a technique for training deep neural networks (DNNs) over distributed edge clients without accessing their raw data \cite{luo2021cost, wang2022quantized,jiang2022model}.
However, due to heterogeneous local datasets in practice, FL is facing the inter-client variance problem, and various FL algorithms have been proposed to address this issue.
For example, \texttt{FedProx} in \cite{li2020federated} introduces a regularization term in local objective functions to control model divergence.
\texttt{SCAFFOLD} in \cite{karimireddy2020scaffold} corrects drift in local models using control variates.
\texttt{FedALRC} in \cite{wei2022non} incorporates a regularization scheme to constrain the local Rademacher complexity, enhancing the generalization ability of FL.
Moreover, the work in \cite{sattler2019robust} proposes sparse ternary compression, which relies on frequent communication of weight updates between clients to mitigate divergence among local models.
However, most of these studies primarily concentrated on training small to moderate-sized DNN models.
Despite the challenge posed by the extensive computations and substantial data access required for training large-scale DNN models,
the rapid advancements in computing power in recent years have sparked a growing interest in utilizing FL networks for training such models \cite{wang2022quantized, lu2022eta, thompson2022importance, zheng2020design, chai2020fedat}.
In order to accelerate the training process and improve the generalization capability of large-scale DNN models, data normalization techniques such as batch normalization (BN) and group normalization (GN) are used \cite{santurkar2018does,wu2018group}.
It is interesting to note that recent studies have suggested that BN may negatively impact FL \cite{wang2022quantized,zheng2020design, chai2020fedat}.
For instance, the work in \cite{wang2022quantized} found that when clients have i.i.d. data, the ResNet-20 model trained with BN \cite{he2016deep} can achieve a testing accuracy of 85\%, but when the local data are non-i.i.d., the accuracy drops dramatically to 60\%.
This is because that with BN, the parameters of every DNN layer are determined by the statistics associated with its local input.
Thus, when the local datasets at the clients are non-i.i.d., there is a mismatch between the local data statistics and the global data statistics, which makes the gradients computed by the local datasets greatly differ from the global gradient computed by the global dataset.
In view of the fact that BN has demonstrated its success in the DNN training and has become a fundamental component of many state-of-the-art DNN models \cite{li2021fedbn}, it is crucial to fix the issue and develop efficient FL algorithms with BN.

\subsection{Related Works}\label{sec:related works}

In recent years, several attempts have been made toward this direction.
Firstly, some works replaced BN with alternative normalization mechanisms that do not rely on batch samples.
For example, to recover the accuracy loss caused by BN in the non-i.i.d. setting, the works \cite{hsieh2020non} and \cite{du2022rethinking} replace BN with Group Normalization (GN) and Layer Normalization (LN), respectively.
However, these alternative normalization mechanisms have some disadvantages.
Specifically, GN is an instance-based method of normalization, which is highly sensitive to the noise in data samples, while LN assumes that the inputs to all neurons within the same DNN layer contribute equally to the final inference \cite{hsieh2020non,du2022rethinking}.
In this regard, BN is likely to outperform GN and LN for some applications in terms of a faster convergence rate and higher testing accuracy.

Secondly, some studies adopted the local BN technique to reduce the gradient drift of local models under non-i.i.d. data.
For example, \texttt{FedBN} in \cite{li2021fedbn} trains all BN parameters (including batch mean, batch variance, scale parameter, and shift parameter) locally in clients based on only the local datasets, while \texttt{SiloBN} \cite{andreux2020siloed} updates only the batch mean and batch variance locally in clients but sharing other BN parameters including the scale and shift parameters with the server for global aggregation.
Different from \cite{andreux2020siloed}, the work in \cite{mills2021multi} suggests that updating both scale and shift parameters locally while aggregating batch mean and batch variance globally would achieve faster FL convergence.
However, all of the above methods would still perform unsatisfyingly in a general non-i.i.d. scenario where the clients have different label distributions (e.g., the data of each client contains only a subset of labels and they are different to each other).
Therefore, to guarantee FL convergence under such general non-i.i.d. data cases, it is insufficient to aggregate only part of the DNN model parameters at the server while keeping some BN parameters updated in clients.

In view of this, several recent FL works have considered aggregating all DNN model parameters at the server while modifying the mode of updating BN parameters.
Some studies considered modifying the aggregation scheme at the server side.
For instance, \texttt{FedBS} in \cite{idrissi2021fedbs} introduces a heuristic weighted aggregation scheme on local model parameters and BN parameters, by giving a greater weight to clients with a higher local loss function value.
\texttt{FedDNA} in \cite{duan2021feddna} adopts an importance weighting method to aggregate local batch means and batch variances, but requires the server to train an additional adversarial model to calculate the weighting coefficients.
Some works considered modifying the local model updating steps at the client side.
As an example, the work in \cite{vsajina2023peer} utilizes a fine-tuning technique to train BN parameters in local models, where if the validation accuracy of a local model does not improve, only its BN parameters are updated and the remaining network coefficients are frozen.
The above methods maintain tracking of the moving average of batch mean and batch variance throughout the entire training process and also utilizes it for model inference as a common technique in BN.
Unlike the previous approaches, \texttt{HeteroFL} in \cite{diao2020heterofl} proposed a static BN approach that does not track the moving average but instead makes use of the instantaneous output statistics of DNN layers in model inference.
Even so, the performance of the above works still drop significantly under the general non-i.i.d. data case. Besides, none of them have examined the theoretical impact of BN on the FL convergence speed.

Furthermore, the FL works above only observe that under the non-i.i.d. data, the mismatch between the local and global \emph{statistical parameters} (including batch mean and batch variance) obtained in the forward propagation procedure would adversely affect the FL performance.
However, they overlooked that the mismatch between the local and global gradients with respect to (w.r.t.) statistical parameters computed during the backward propagation procedure also harms FL convergence.

\subsection{Contributions}

This paper investigates why BN damages FL on the non-i.i.d. data.
We highlight that there are two types of mismatches that jointly influence the convergence of FL with BN: one is the mismatch between the local and global statistical parameters, and the other is that between the related local and global gradients w.r.t. statistical parameters.
Specifically, in the previous research aimed at improving the FL performance with BN, only the statistical parameters obtained in the forward propagation are noted, whereas the gradients w.r.t. statistical parameters computed in the backward propagation are ignored.
Unfortunately, the gradients w.r.t. statistical parameters are intermediate variables in calculating the gradients w.r.t. the DNN model parameters in the training process, and thereby have a significant impact on the convergence of FL.
This is the fundamental reason why the previous methods still suffer performance loss.

In view of this, we present the first theoretical analysis of how the above two types of mismatches caused by BN and non-i.i.d data degrade the convergence of \texttt{FedAvg} \cite{mcmahan2017communication}.
To overcome this effect, we further propose a new FL algorithm, named \texttt{FedTAN} (\underline{fed}erated learning algorithm \underline{t}ailored for b\underline{a}tch \underline{n}ormalization), which eliminates the two types of mismatches above by layer-wise statistical parameter aggregation and can achieve robust FL performance under different data distributions.
In particular, our main contributions include:

\begin{enumerate}
\item[(1)]
\textbf{Convergence analysis of FL with BN:}
We consider the general non-convex FL problems.
To the best of our knowledge, we are the first to analyze theoretically how BN affects the convergence rate of \texttt{FedAvg}.
The derived theoretical results indicate that the gradient deviation would occur in local models if the local statistical parameters in BN and the related gradients are inconsistent with the global ones obtained by the global dataset.
As a consequence, the resulting gradient deviation not only slows down, but also biases the convergence of FL.

\item[(2)]
\textbf{Relation analysis between local and global statistical parameters as well as their gradients:}
The formal mathematical expressions of both the i.i.d. and the non-i.i.d. data cases are defined.
Based on this, we conclude that the i.i.d. data leads to the same local and global statistical parameters as well as their gradients, whereas the non-i.i.d. data produces the mismatches.
Also, we find that merely eliminating the mismatch between local and global statistical parameters, without ensuring consistency between local and global gradients w.r.t. statistical parameters, cannot guarantee FL convergence.

\item[(3)]
\textbf{\texttt{FedTAN:}}
The above observations suggest that if all clients in FL have the same statistical parameters and related gradients as the global ones, then undesirable slowdown caused by gradient deviation under BN can be eliminated.
Inspired by this, we develop a new FL algorithm, \texttt{FedTAN}, which leverages layer-wise aggregation to match not only the local and global statistical parameters during forward propagation but also their associated gradients during backward propagation.

\item[(4)]
\textbf{Experiments:}
\texttt{FedTAN} is applied to classify color images in the CIFAR-10 dataset.
Experimental results indicate that \texttt{FedTAN} has promising performance and outperforms benchmark schemes.
In particular, we show that even though \texttt{FedTAN} requires extra message exchanges between the clients and the server, it is still capable of achieving a satisfying training performance and faster convergence rate under different data distributions.
\end{enumerate}

\textbf{Synopsis:}
Section \ref{sec:system model} introduces the FL algorithm on a DNN model with BN layers.
Section \ref{sec:convergence analysis} analyzes the convergence of FL with BN, while Section \ref{sec:relation S delta S} examines the relations between the local and global statistical parameters as well as their gradients, under different data distributions.
Based on the theoretical results, Section \ref{Sec:FedTAN} develops an FL algorithm tailored for BN.
Experimental results are presented in Section \ref{sec:experiment results}, and finally, Section \ref{sec:conclusion} concludes this paper.

\section{Federated Learning with Batch Normalization}\label{sec:system model}

\subsection{Batch Normalization}\label{subsec:Batch Normalization}

\begin{figure}[t]
\centering
\includegraphics[width= 3.8 in ]{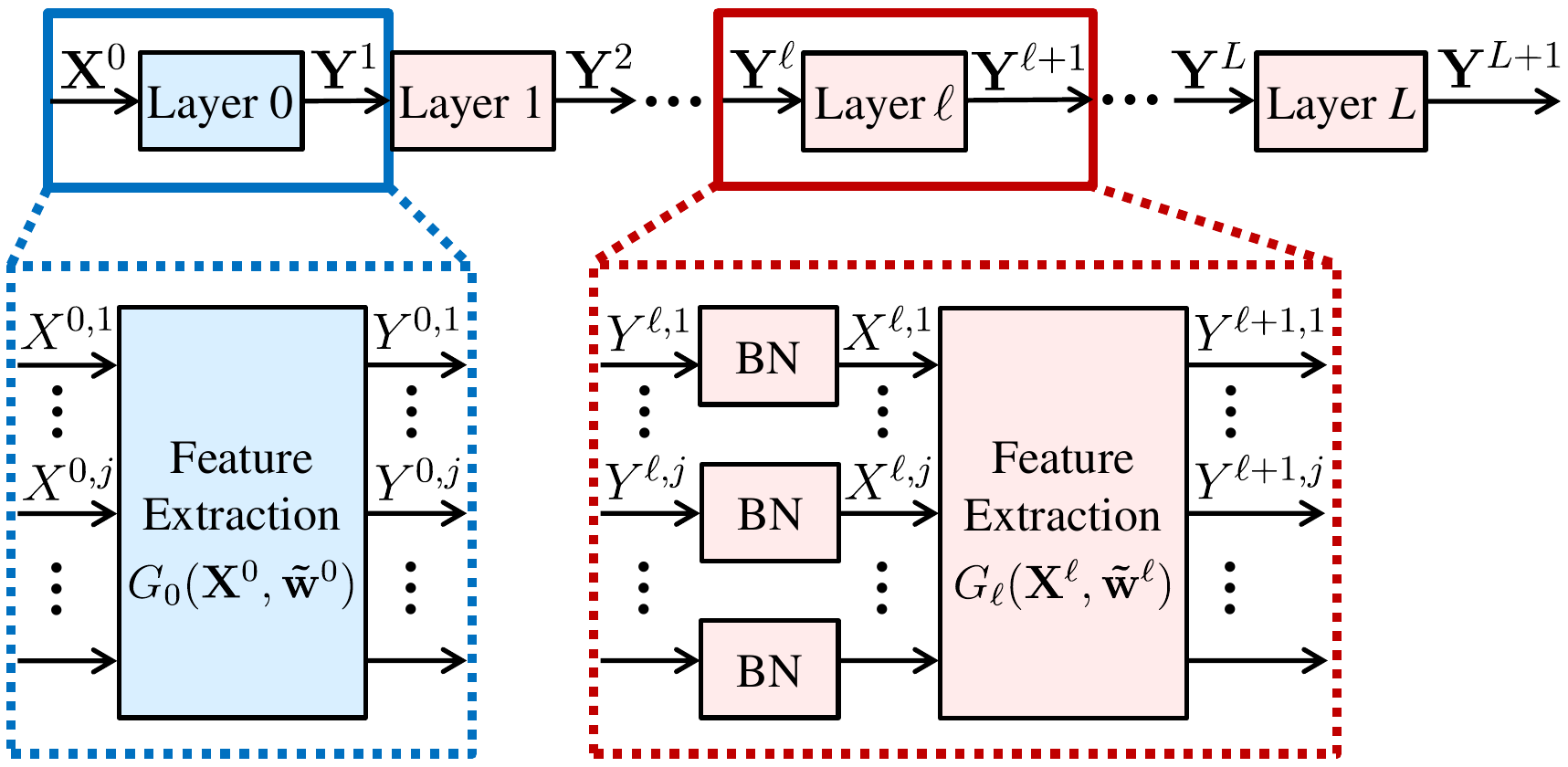}
\caption{An example of DNN model with BN layers. Each feature extraction layer can be comprised of convolution layer, fully-connected layer, pooling layer, flatten layer, activation layer, etc.}
\label{fig:plain DNN}
\end{figure}

As shown in Fig. \ref{fig:plain DNN}, we follow the similar architecture as the plain layer-by-layer network in \cite{he2016deep} and consider a DNN with $L$ BN layers.
Each feature extraction layer except the input layer (i.e., the 0-th layer) is preceded by a BN layer, and each feature extraction layer can be composed of a convolution layer, a fully-connected layer, a pooling layer, a flatten layer, or an activation layer, etc.

Let $\mathbf{Y}^{\ell}$ represent the input to the $\ell$-th BN layer, for $\ell=1,\ldots, L$.
Then, each $j$-th element of $\mathbf{Y}^{\ell}$ is normalized as
\begin{equation}\label{eq:batch output}
X^{\ell, j} = \gamma^{\ell,j} \underbrace{\frac{Y^{\ell, j}-\mu^{\ell, j}}{\sqrt{(\sigma^{\ell,j})^{2}+\epsilon}}}_{\triangleq {\hat Y}^{\ell, j}} + \beta^{\ell,j}
,
\end{equation}
where $\gamma^{\ell,j}$ is a scale parameter, $\beta^{\ell,j}$ is a shift parameter, and $\epsilon$ is a small positive value.
Here, batch mean $\mu^{\ell,j}$ and batch variance $(\sigma^{\ell,j})^{2}$ are obtained by
\begin{subequations}\label{eq:mu sigma ellj}
\begin{equation}\label{eq:mu ellj}
\mu^{\ell,j} =
\mathbb{E}_{\mathcal{D}}[Y^{\ell, j}]
,
\end{equation}
\begin{equation}\label{eq:sigma ellj}
(\sigma^{\ell,j})^{2} =
\mathbb{E}_{\mathcal{D}}[ ( Y^{\ell, j} - \mu^{\ell,j} )^2 ]
,
\end{equation}
\end{subequations}
where $\mathbb{E}_{\mathcal{D}}[\cdot]$ is the expectation over the data samples $\mathcal{D}$.
After that, the output of the BN layer, $\mathbf{X}^{\ell}$, is fed into its subsequent feature extraction layer with the computation function as
\begin{equation}\label{eq:batch input}
\mathbf{Y}^{\ell+1} = G_{\ell} (\mathbf{X}^{\ell}, \mathbf{\tilde w}^{\ell})
, \,
\ell=0, \ldots, L
,
\end{equation}
where $\mathbf{\tilde w}^{\ell}$ is the network coefficients and $\mathbf{Y}^{\ell+1}$ is the output.

For a DNN with BN, the model parameters include both the \emph{gradient parameters} (denoted by $\mathbf{w}$) which are commonly contained in all layers and updated by model gradients, and the \emph{statistical parameters} (denoted by $\mathcal{S}$) which are solely contained in BN layers and updated by the statistics of intermediate outputs in DNN \cite{duan2021feddna}.
The gradient parameters $\mathbf{w}$ consist of both the network coefficients $\{ \mathbf{\tilde w}^{\ell} \}$ as well as the scale parameters $\bm{\gamma}^{\ell} ( = \{\gamma^{\ell,j}\} )$ and shift parameters $\bm{\beta}^{\ell} ( = \{\beta^{\ell,j}\} )$, while the statistical parameters are batch mean $\bm{\mu}^{\ell} ( = \{\mu^{\ell,j}\})$ and batch variance $(\bm{\sigma}^{\ell})^2 (= \{ (\sigma^{\ell,j})^2 \})$.
Thus, the \emph{model parameters} in a DNN are $\mathbf{W} = \{\mathbf{w}, \mathcal{S}\}$.
It is important to note that $\mathbf{w}$ and $\mathcal{S}$ are interdependent.

In order to train a DNN to minimize a particular cost function $F({\mathbf{W}})$, the gradient of $F$ w.r.t. the gradient parameters $\mathbf{w}$, $\nabla_{\mathbf{w}} F$, must be computed.
However, when using BN, $\nabla_{\mathbf{w}} F$ is a function not only of the gradient parameters $\mathbf{w}$, but also of the statistical parameters $\mathcal{S}$ and their gradients $\Delta \mathcal{S} \triangleq \{ \nabla_{\bm{\mu}^{\ell}} F, \nabla_{(\bm{\sigma}^{\ell})^2} F \}$.
Specifically, $\mathcal{S}$ and $\Delta \mathcal{S}$ are used in the forward and backward propagations, respectively, to calculate $\nabla_{\mathbf{w}} F$ \cite{ioffe2015batch}.
Thus, we write $\nabla_{\mathbf{w}} F$ explicitly as $\nabla_{\mathbf{w}} F(\mathbf{w}; \mathcal{S}, \Delta \mathcal{S})$.

\subsection{Federated Learning Algorithm}\label{subsec:FL algorithm}

As shown in Fig. \ref{fig:CL VS FL}(b), the FL scheme involves a server coordinating $N$ clients to solve the following learning problem
\begin{equation}\label{objective function}
\min \limits_{{\mathbf{W}} } \; F({\mathbf{W}}) = \sum_{i = 1}^N p_i F_i({\mathbf{W}})
,
\end{equation}
where $p_i = {|\mathcal{D}_i|}/{|\mathcal{D}|}$, $\mathcal{D}_i$ is the local dataset in client $i$, $\mathcal{D} = \bigcup_{i = 1}^N \mathcal{D}_i$ is the global dataset, $F({\mathbf{W}}) = \mathbb{E}_{\mathcal{D}}[\mathcal{L}({\mathbf{W}};\mathcal{D})]
$ is the global cost function on $\mathcal{D}$ and $F_i({\mathbf{W}}) = \mathbb{E}_{\mathcal{D}_i}[\mathcal{L}({\mathbf{W}};\mathcal{D}_i)]$ is the local one on $\mathcal{D}_{i}$ in which $\mathcal{L}$ is the loss function of data samples.
This paper considers the celebrated \texttt{FedAvg} algorithm \cite{mcmahan2017communication}, which executes the following three steps every ${r}$-th iteration:
\begin{enumerate}[(i)]
\item \textbf{Initialization}: All clients receive the global model $\{ {\bar {\mathbf{w}}}_{r-1}, {\bar {\mathcal{S}}}_{r-1} \}$ in the last iteration from the server.
\item \textbf{Local model updating}: Each client $i \in [N] \triangleq \{ 1,\ldots,N \}$ updates its local gradient parameters through $E$ successive steps of local gradient descent\footnote{ In the case of mini-batch stochastic gradient descent (SGD), the expectation for computing the statistical parameters in (2) 
   is w.r.t. the mini-batch samples from each local dataset.
   For ease of illustration, we consider the full gradient descent for theorem development but use mini-batch SGD for experiments.}.
   Specifically, initialize the local model by \eqref{eq:local SGD_a} and perform \eqref{eq:local SGD_b} for $t \in [E]$.
\begin{subequations}\label{eq:local SGD}
\begin{align}
& \{ \mathbf{w}^{r,0}_{i}, {\bar {\mathcal{S}}}^{r,0}_{\mathcal{D}_i} \} = \{ {\bar {\mathbf{w}}}_{r-1}, {\bar {\mathcal{S}}}_{r-1} \}
,
\label{eq:local SGD_a}\\
&
\mathbf{w}^{r,t}_{i} = {\mathbf{w}}^{r,t-1}_{i} - \gamma \nabla_{\mathbf{w}} F_{i}( {\mathbf{w}}^{r,t-1}_{i} ; {\mathcal{S}}^{r,t}_{\mathcal{D}_i}, \Delta {\mathcal{S}}^{r,t}_{\mathcal{D}_i}),
\label{eq:local SGD_b}
\end{align}
\end{subequations}
where $\gamma>0$ is the learning rate, $E$ is the number of local updating steps, and ${\mathcal{S}}^{r,t}_{\mathcal{D}_i}$ is the statistical parameters computed via \eqref{eq:mu sigma ellj} with the local model ${\mathbf{w}}^{r,t-1}_{i}$ and the local dataset $\mathcal{D}_{i}$.
During this time, each client updates local statistical parameters for model inference by moving average as \cite{vsajina2023peer}
\begin{equation}\label{eq:moving avg}
{\bar {\mathcal{S}}}^{r,t}_{\mathcal{D}_i} = (1-\rho) {\bar {\mathcal{S}}}^{r,t-1}_{\mathcal{D}_i} + \rho {\mathcal{S}}^{r,t}_{\mathcal{D}_i}, \,
t \in [E]
,
\end{equation}
where $\rho$ is the momentum of moving average.
\item \textbf{Aggregation}:
The local model $\{ \mathbf{w}_i^{r,E}, {\bar {\mathcal{S}}}^{r,E}_{\mathcal{D}_i}\}$ in each client $i \in [N]$ is uploaded to the server for aggregating a new global model by
\begin{equation}\label{eq:global model}
\{ {\bar {\mathbf{w}}}_{r}, {\bar {\mathcal{S}}}_{r} \} = \sum_{i = 1}^N p_i \{ \mathbf{w}^{r,E}_{i}, {\bar {\mathcal{S}}}^{r,E}_{\mathcal{D}_i} \}
.
\end{equation}
\end{enumerate}

\begin{figure}[t]
\centering
\subfigure[The i.i.d. data case.]{
\includegraphics[width= 2 in ]{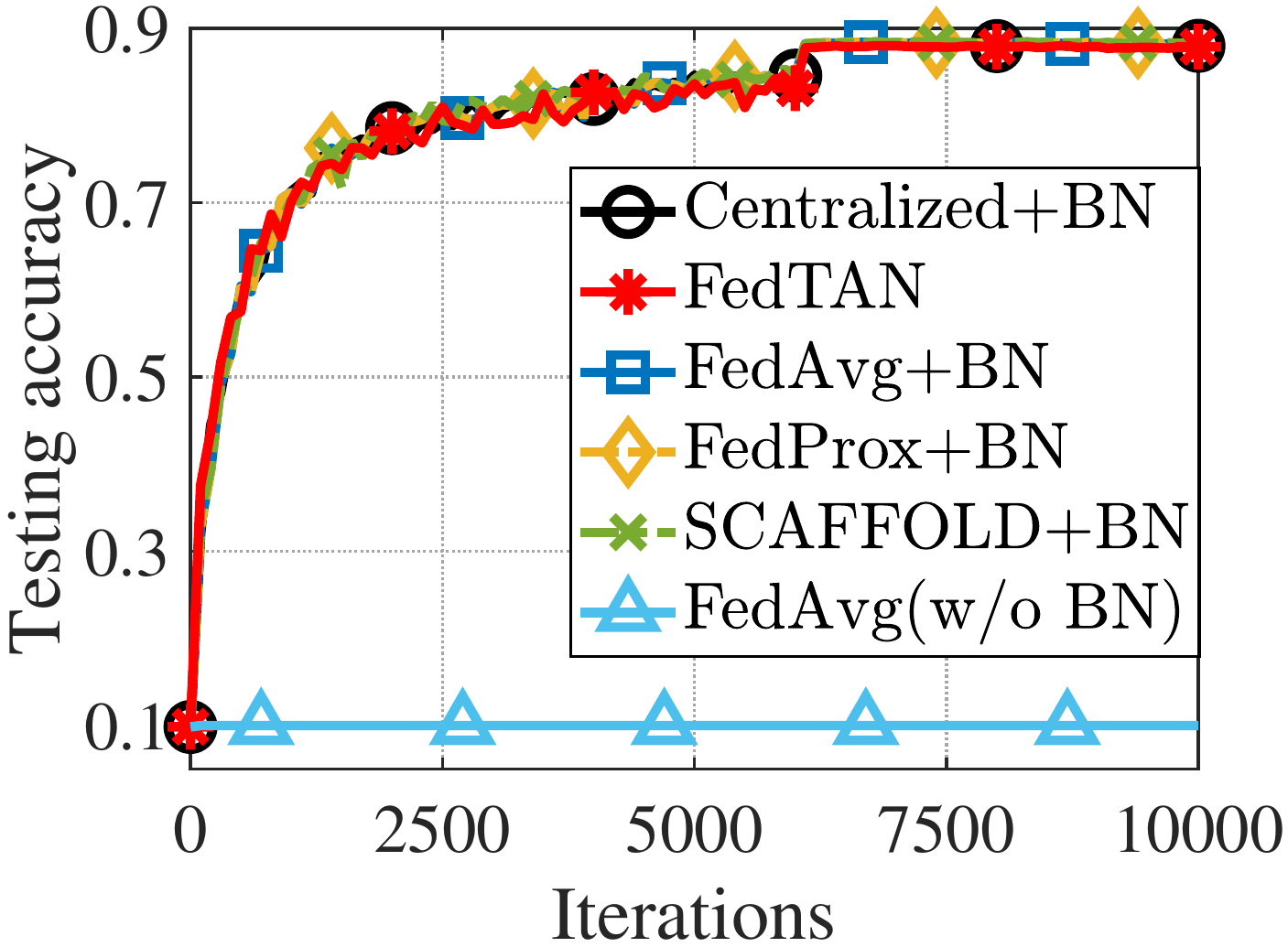}}
\subfigure[The non-i.i.d. data case.]{
\includegraphics[width= 2 in ]{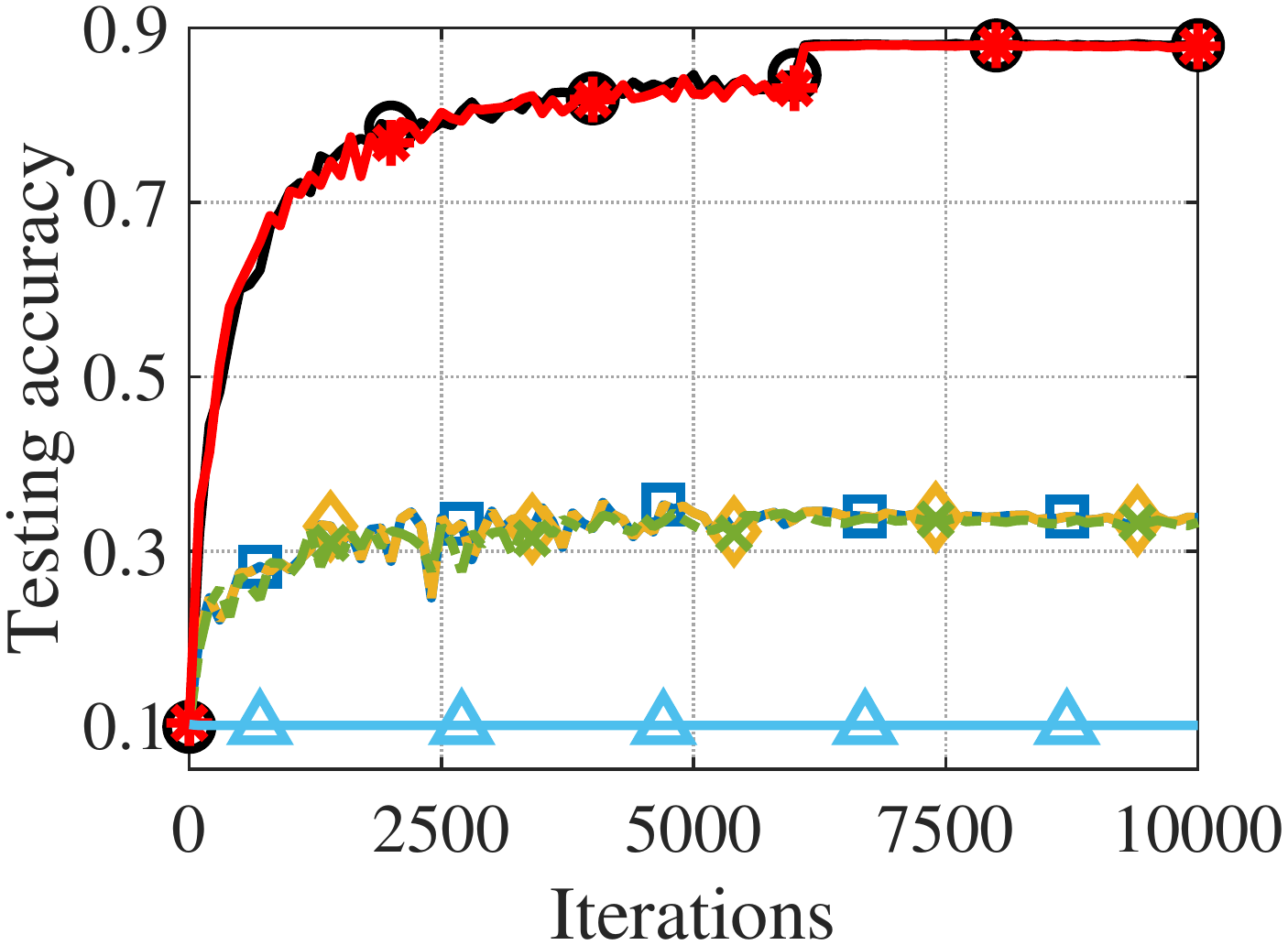}}
\caption{FL performance with BN and $E=1$. ResNet-20 is trained by CIFAR-10 dataset.}
\label{fig:performance E1}
\end{figure}

It is found that FL with BN performs much worse under the non-i.i.d. local datasets, even when $E=1$ \cite{wang2022quantized,zheng2020design, chai2020fedat}.
As an example, Fig. \ref{fig:performance E1} compares the testing accuracy of different FL schemes on training ResNet-20 \cite{he2016deep} under different data distributions, where $E = 1$ and detailed settings are presented in Section \ref{sec:experiment results}.
Firstly, according to Fig. \ref{fig:performance E1}(a), \texttt{FedAvg} is unable to train such a large model effectively without BN, whereas BN greatly improves performance when applied.
Secondly, based on the theory on FL convergence in the current research \cite{li2019convergence,reisizadeh2020fedpaq,yu2019parallel}, \texttt{FedAvg} with $E = 1$ should perform similarly to the centralized learning with a larger batch size.
However, Fig. \ref{fig:performance E1}(b) shows that \texttt{FedAvg+BN} suffers serious degradation when the data becomes non-i.i.d.
Additionally, the performance of advanced FL algorithms like \texttt{FedProx} \cite{li2020federated} and \texttt{SCAFFOLD} \cite{karimireddy2020scaffold} is also poor.
As a result, it is necessary to understand how the BN damages FL performance under non-i.i.d data and propose a remedy.

\section{How BN damages FL?}\label{sec:convergence analysis}

\subsection{Gradient Deviation under BN}

The reason why BN damages FL is that under the distributed setting, the local statistical parameters and their gradients differ from the global ones, resulting in an inconsistency between the global gradient and the average of local gradients computed by clients.
On general non-convex optimization problems, it appears from \eqref{objective function} and the analyses in \cite{wang2020tackling, yu2019parallel, li2020federated, karimireddy2020scaffold} that, the FL convergence is particularly sensitive to the following condition
\begin{equation}\label{eq:nabla Fw = sum pi Fi}
\nabla_{\mathbf{w}} F({\mathbf{W}}) = \sum_{i = 1}^N p_i \nabla_{\mathbf{w}} F_i({\mathbf{W}})
.
\end{equation}
However, when BN is used, $\mathbf{W}$ includes both gradient parameters and statistical parameters.
Consequently, \eqref{eq:nabla Fw = sum pi Fi} holds only if the left-hand side (LHS) and the right-hand side (RHS) of \eqref{eq:nabla Fw = sum pi Fi} are determined by the same $\mathcal{S}$ and $\Delta \mathcal{S}$, i.e.,
\begin{equation}\label{ineq:nabla F Fi SD}
\nabla_{\mathbf{w}} F({\mathbf{w}};\mathcal{S}_{\mathcal{D}}, \Delta \mathcal{S}_{\mathcal{D}}) = \sum_{i = 1}^N p_i \nabla_{\mathbf{w}} F_i ({\mathbf{w}};\mathcal{S}_{\mathcal{D}}, \Delta \mathcal{S}_{\mathcal{D}})
,
\end{equation}
where $\mathcal{D} = \bigcup_{i = 1}^N \mathcal{D}_i$ refers to the global dataset.

Unfortunately, in the FL setting, each client $i$ can only access its local dataset $\mathcal{D}_i$ and, as a result, can only compute $\nabla_{\mathbf{w}} F_i({\mathbf{w}}; \mathcal{S}_{\mathcal{D}_i}, \Delta \mathcal{S}_{\mathcal{D}_i})$.
\begin{observation}
In the non-i.i.d. data case, the local dataset $\mathcal{D}_i$ and global dataset $\mathcal{D}$ generally cause $\mathcal{S}_{\mathcal{D}_i} \neq \mathcal{S}_{\mathcal{D}}$ and $\Delta \mathcal{S}_{\mathcal{D}_i} \neq \Delta \mathcal{S}_{\mathcal{D}}$.
As a result, we have
\begin{equation}\label{ineq:nabla Fi SD SDi}
\nabla_{\mathbf{w}} F_i ({\mathbf{w}};\mathcal{S}_{\mathcal{D}_i}, \Delta \mathcal{S}_{\mathcal{D}_i})
\neq
\nabla_{\mathbf{w}} F_i ({\mathbf{w}};\mathcal{S}_{\mathcal{D}}, \Delta \mathcal{S}_{\mathcal{D}})
.
\end{equation}
\end{observation}
A formal discussion about this observation will be provided in Section \ref{sec:relation S delta S noniid}.
Then, substituting \eqref{ineq:nabla Fi SD SDi} into \eqref{ineq:nabla F Fi SD}, we have
\begin{equation}
\nabla_{\mathbf{w}} F({\mathbf{w}};\mathcal{S}_{\mathcal{D}}, \Delta \mathcal{S}_{\mathcal{D}})
\neq
\sum_{i = 1}^N p_i \nabla_{\mathbf{w}} F_i ({\mathbf{w}};\mathcal{S}_{\mathcal{D}_i}, \Delta \mathcal{S}_{\mathcal{D}_i})
,
\end{equation}
which contradicts the default assumption in \eqref{eq:nabla Fw = sum pi Fi}.
Hence, to characterize the gradient deviation resulting from both the mismatched statistical parameters and the mismatched gradients w.r.t. statistical parameters, we introduce Assumption \ref{assumption:BN disturbance}.
\begin{assumption}\label{assumption:BN disturbance}
The gradient deviation under BN satisfies $\| \nabla_{\mathbf{w}} F_i({\bar {\mathbf{w}}}_{r-1}; \mathcal{S}_{\mathcal{D}_i}^{r,1}, \Delta \mathcal{S}_{\mathcal{D}_i}^{r,1}) - \nabla_{\mathbf{w}} F_i ({\bar {\mathbf{w}}}_{r-1}; \mathcal{S}_{\mathcal{D}}^{r,1}, \Delta \mathcal{S}_{\mathcal{D}}^{r,1})\|^2 $ $\leq B_i^2$, $\forall i \in [N]$.
\end{assumption}

It is important to note that Assumption \ref{assumption:BN disturbance} considers the discrepancy between $\{\mathcal{S}_{\mathcal{D}_i}^{r,1}, \Delta \mathcal{S}_{\mathcal{D}_i}^{r,1}\}$ and $\{\mathcal{S}_{\mathcal{D}}^{r,1}, \Delta \mathcal{S}_{\mathcal{D}}^{r,1}\}$ at the beginning of each iteration $r$ only.
In fact, according to the convergence analysis presented in Theorem \ref{theorem:1} below, Assumption \ref{assumption:BN disturbance} is sufficient to capture the effects of BN.

In general, the value of $B_i$ in Assumption \ref{assumption:BN disturbance} depends on both the architecture of DNN and the distribution of training data.
Without BN or similar normalization techniques like Decorrelated Batch Normalization \cite{huang2018decorrelated} and Batch Group Normalization \cite{zhou2020batch}, statistical parameter $\mathcal{S}$ and its gradient $\triangle \mathcal{S}$ are not present.
In this case, model gradients for each data sample are independent, leading to $B_i = 0$ irrespective of data distribution\footnote{
For instance, GN divides adjacent neurons into groups of a predetermined size and calculates the mean and variance for each group.
As a result, the GN output for each sample is computed independently and does not rely on mini-batch statistics, leading to $B_i = 0$.
}.
However, with BN in a DNN model, $B_i = 0$ when the data is i.i.d. due to $\mathcal{S}_{\mathcal{D}_i} = \mathcal{S}_{\mathcal{D}}$ and $\Delta \mathcal{S}_{\mathcal{D}_i} = \Delta \mathcal{S}_{\mathcal{D}} $.
Conversely, for non-i.i.d. data, $B_i > 0$ because of $\mathcal{S}_{\mathcal{D}_i} \neq \mathcal{S}_{\mathcal{D}}$ and $\Delta \mathcal{S}_{\mathcal{D}_i} \neq \Delta \mathcal{S}_{\mathcal{D}}$.
The formal discussion on the relation between local and global statistical parameters, as well as between their gradients, will be presented in Section \ref{sec:relation S delta S}.

\subsection{Convergence Analysis of \texttt{FedAvg} with BN}

We also make the following assumptions.

\begin{assumption}\label{assumption:lowered bounded}
Global function $F$ is lowered bounded, i.e., $F({\mathbf{w}};\mathcal{S}_{\mathcal{D}}, \Delta \mathcal{S}_{\mathcal{D}}) \geq \underline{F} > - \infty$.
\end{assumption}
\begin{assumption}\label{assumption:L continuous BN}
Local functions $F_i$, are differentiable whose gradients are Lipschitz continuous with constant $L$: $\forall$${\mathbf{w}}$ and ${\mathbf{w}'}$, $F_i({\mathbf{w}'};\mathcal{S}'_{\mathcal{D}_i}, \Delta \mathcal{S}'_{\mathcal{D}_i}) \leq F_i({\mathbf{w}};\mathcal{S}_{\mathcal{D}_i}, \Delta \mathcal{S}_{\mathcal{D}_i}) + ({\mathbf{w}'} - {\mathbf{w}})^T \nabla_{\mathbf{w}} F_i({\mathbf{w}};\mathcal{S}_{\mathcal{D}_i}, \Delta \mathcal{S}_{\mathcal{D}_i}) + \frac{L}{2} \| {\mathbf{w}'} - {\mathbf{w}} \|_2^2$.
\end{assumption}
\begin{assumption}\label{assumption:data var}
It holds that $\| \nabla_{\mathbf{w}} F_i({\mathbf{w}}; \mathcal{S}_{\mathcal{D}}, \Delta \mathcal{S}_{\mathcal{D}}) - \nabla_{\mathbf{w}} F ({\mathbf{w}}; \mathcal{S}_{\mathcal{D}}, \Delta \mathcal{S}_{\mathcal{D}})\|^2 \leq V_i^2$, $\forall i \in [N]$,
which measures the heterogeneity of local datasets.
\end{assumption}

Note that $B_i$ in Assumption \ref{assumption:BN disturbance} results from the mismatch between the local and global statistical parameters as well as that between their gradients, while $V_i$ in Assumption \ref{assumption:data var} measures the difference between global and local function gradients caused by the heterogeneous local datasets across clients like \cite{wang2022quantized,lian2017can}.
Since $F({\mathbf{W}}) = \mathbb{E}_{\mathcal{D}}[\mathcal{L}({\mathbf{W}};\mathcal{D})]
$ and $F_i({\mathbf{W}}) = \mathbb{E}_{\mathcal{D}_i}[\mathcal{L}({\mathbf{W}};\mathcal{D}_i)]$, it can be observed that $V_i=0$ arises from the i.i.d. data which suggests consistent statistical parameters and gradients, i.e., $\mathcal{S}_{\mathcal{D}_i} = \mathcal{S}_{\mathcal{D}}$ and $\Delta \mathcal{S}_{\mathcal{D}_i} = \Delta \mathcal{S}_{\mathcal{D}}$, so $V_i=0$ generally indicates $B_i=0$.
By contrast, $V_i \neq 0$ caused by the non-i.i.d. data would generally imply $B_i \neq 0$.
The main convergence result is presented below.
\begin{theorem}\label{theorem:1}
Suppose Assumptions \ref{assumption:BN disturbance} to \ref{assumption:data var} hold.
Let $R$ represent the number of iterations and $T = RE$ denote the total number of gradient descent updates per client.
With $\gamma = N^{\frac{1}{2}} / (4L{T}^{\frac{1}{2}}) $ and $E \leq T^{\frac{1}{4}}/N^{\frac{3}{4}}$ where $T \geq N^3$, we have

\vspace{-0.3cm}
\begin{small}
\begin{align}\label{eq:theorem_1}
\frac{1}{R} \sum_{r = 1}^{R}
\big\| \nabla_{\mathbf{w}} F({\bar {\mathbf{w}}}_{r-1}; \mathcal{S}_{\mathcal{D}}^{r,1}, \Delta \mathcal{S}_{\mathcal{D}}^{r,1}) \big\|^2
& \leq \frac{24L}{( T N)^{\frac{1}{2}}}
\left( F({\bar {\mathbf{w}}}_{0};\mathcal{S}_{\mathcal{D}}^{1,1}, \Delta \mathcal{S}_{\mathcal{D}}^{1,1}) - \underline{F} \right) \notag \\
&
+
\underbrace{\frac{2}{( T N)^{\frac{1}{2}}} \sum_{i = 1}^N p_i V_i^2}_{\text{\small (a) caused by} \atop \text{\small non-i.i.d. data}}
+
\!\!\!\!\!
\underbrace{\frac{2}{( T N)^{\frac{1}{2}}} \sum_{i = 1}^N p_i B_i^2
+
6 \sum_{i = 1}^N p_i B_i^2}_{\text{\small (b) caused by mismatches in statistical} \atop \text{\small parameters and in their gradients}}
\!\!\!\!\!\!\!\!
,
\end{align}
\end{small}
\vspace{-0.2cm}

\noindent
where ${\bar {\mathbf{w}}}_{r-1}$ refers to the gradient parameters in global model defined in \eqref{eq:global model}\footnote{
The selected learning rate $\gamma = N^{\frac{1}{2}} / (4L{T}^{\frac{1}{2}}) $ enables a linear speedup in both $T$ and $N$ for FL but is generally conservative compared to practical values.
Besides, accurately estimating the Lipschitz continuous constant $L$ for training complex DNN models is challenging.
Thus, in the experiments of Section \ref{sec:experiment results}, we primarily rely on empirical values to select suitable $\gamma$.
}.
\end{theorem}

\emph{Proof}: see Appendix \ref{sec:thm 1 proof}.

It can be found from the RHS of \eqref{eq:theorem_1}, when training a DNN model containing BN layers, the convergence of FL with BN is affected by various parameters, including the number of clients $N$, the non-i.i.d. level of local datasets $\{V_i\}$, and the gradient deviation under BN $\{B_i\}$.
It can be seen from the terms (a) and (b) in \eqref{eq:theorem_1} that the FL convergence is hampered by both the non-i.i.d. level of local datasets $\{V_i\}$ and the gradient deviation under BN $\{B_i\}$.
Furthermore, we discover several important insights as follows.
\begin{itemize}
\item
First of all, the two terms in (b) are related to the gradient deviation under BN.
As a result, when training a model without using BN, these two terms can be omitted.
However, in practice, BN is known to facilitate training and enhance the generalization capability of DNN models.

\item
Secondly, due to mismatches in both the statistical parameters and their corresponding gradients, the second component in term (b) does not reduce as $T$ increases, meaning that BN would lead FL to a biased solution.
This result indicates that BN has a profound effect on FL convergence.

\item
Last but not the least, according to Assumption \ref{assumption:BN disturbance}, if $\mathcal{S}_{\mathcal{D}_i}^{r,1} = \mathcal{S}_{\mathcal{D}}^{r,1}$ and $\Delta \mathcal{S}_{\mathcal{D}_i}^{r,1} = \Delta \mathcal{S}_{\mathcal{D}}^{r,1}$ hold in each client $i$, then $B_i = 0$.
Consequently, the term (b) vanishes, enabling \texttt{FedAvg} to converge to a proper stationary solution.
In the next Section \ref{sec:relation S delta S}, we analyze the relation between local and global statistical parameters, as well as between their gradients, under different data distributions.
\end{itemize}

\begin{remark}
To reduce the computational and communication overload of FL in the practical implementation, some strategies such as the mini-batch SGD for local model updating and the partial client participation in global model aggregation can be employed \cite{wang2022quantized}.
With these strategies, the convergence analysis in Theorem \ref{theorem:1} can be extended, and does not change the conclusion.
\end{remark}

\section{Relation between local and global statistical parameters and gradients}
\label{sec:relation S delta S}

In supervised learning, each sample consists of data and a label. Thus, the loss function of each sample depends on both the output of the DNN model and the corresponding true label.
We consider the DNN with BN as in Fig. \ref{fig:plain DNN}.
Then, in FL, the local cost function in each client $i \in [N]$ is computed by
\begin{equation}\label{eq:Fi}
F_i
=
\mathbb{E}_{\mathcal{D}_{i}} [ \underbrace{\mathcal{L} (\mathbf{Y}_{i}^{L+1}, \operatorname{label}(\mathbf{X}_{i}^{0}))}_{\triangleq \mathcal{L}_i} ]
,
\end{equation}
where $\mathcal{L}_i$ is the loss function of each sample $\{\mathbf{X}_{i}^{0},$ $\operatorname{label}(\mathbf{X}_{i}^{0})\} \in \mathcal{D}_i$, $\mathbf{X}_i^{0}$ is the sample data (i.e., the input to the DNN model), $\operatorname{label}(\mathbf{X}_{i}^{0})$ is the corresponding label, and $\mathbf{Y}_{i}^{L+1}$ is the output of the DNN model.
Similarly, in centralized learning, with the global dataset $\mathcal{D}$, the global cost function is computed by
\begin{equation}\label{eq:F}
F
=
\mathbb{E}_{\mathcal{D}} [ \underbrace{\mathcal{L} (\mathbf{Y}_{g}^{L+1}, \operatorname{label}(\mathbf{X}_{g}^{0}))}_{\triangleq \mathcal{L}_g} ]
.
\end{equation}
where $\mathcal{L}_{g}$ is the loss function of each sample $\{\mathbf{X}_{g}^{0},$ $\operatorname{label}(\mathbf{X}_{i}^{0})\} \in \mathcal{D}$, $\mathbf{X}_{g}^{0}$ and $\mathbf{Y}_{g}^{L+1}$ are the input and the output of the DNN model, respectively.

\subsection{Relation in i.i.d. Data Case}

Before analyzing the relations between the local and the global statistical parameters, we first denote the probability of the samples with label $c$ in local dataset $\mathcal{D}_i$ as $\Pr (c | \mathcal{D}_i)$, and the one in global dataset $\mathcal{D}$ as $\Pr (c | \mathcal{D})$, where $\sum_{c} \Pr (c | \mathcal{D}_i) = \sum_{c} \Pr (c | \mathcal{D}) = 1$.
Meanwhile, we adopt the probability density function (PDF) $f_{{\mathbf{X}_i^0}}(\mathbf{x}^0 | \mathbf{x}^0 \in c)$ to describe the distribution of sample data with label $c$ in $\mathcal{D}_i$, and $f_{{\mathbf{X}_g^0}}(\mathbf{x}^0 | \mathbf{x}^0 \in c)$ to describe that in $\mathcal{D}$.
\begin{definition}\label{definition:iid noniid}
If each local dataset $\mathcal{D}_i$ and the global dataset $\mathcal{D}$ follow the conditions: (i) the same label distribution, i.e,
\begin{equation}\label{eq:condition1_a}
\Pr (c | \mathcal{D}_i) = \Pr (c | \mathcal{D})
, \,
\forall i \in [N];
\end{equation}
(ii) the same sample data distribution under a certain label, i.e,
\begin{equation}\label{eq:condition1_b}
f_{{\mathbf{X}_i^0}}(\mathbf{x}^0 | \mathbf{x}^0 \in c) = f_{{\mathbf{X}_g^0}}(\mathbf{x}^0 | \mathbf{x}^0 \in c)
, \forall c,
\end{equation}
then we call it \textbf{i.i.d. data} case.
If one of the above conditions does not hold, then we call it \textbf{non-i.i.d. data} case.
\end{definition}
Then, we have the following property for the i.i.d. data case.
\begin{property}\label{property:sD delta sD}
With the same initial gradient parameters ${\bar {\mathbf{w}}}_{r-1}$ for both FL and centralized learning, in the i.i.d. data case, the statistical parameters and their gradients obtained by each client in FL and by centralized learning are the same, i.e.,
\begin{equation}
\mathcal{S}_{\mathcal{D}_i}^{r,1} = \mathcal{S}_{\mathcal{D}}^{r,1}
, \,
\Delta \mathcal{S}_{\mathcal{D}_i}^{r,1} = \Delta \mathcal{S}_{\mathcal{D}}^{r,1}
.
\end{equation}
\end{property}

\subsubsection{Proof of $\mathcal{S}_{\mathcal{D}_i}^{r,1} = \mathcal{S}_{\mathcal{D}}^{r,1}$}

Since the forward propagation procedure of a DNN model is performed from the input to the output layer in order, we first analyze the relations between the local and the global statistical parameters (including batch mean and batch variance) for the first BN layer, and then extend them to other layers by induction.

\underline{\textbf{Batch mean and variance in the 1st BN layer}}:
According to \eqref{eq:batch input}, we have $\mathbf{Y}_{i}^{1} = G_0 (\mathbf{X}_{i}^{0}, \mathbf{\tilde w}^{0})$, $i\in[N]$, and $\mathbf{Y}_{g}^{1} = G_0 (\mathbf{X}_{g}^{0}, \mathbf{\tilde w}^{0})$, where $\mathbf{\tilde w}^{0}$ is the model coefficient of the 0-th feature extraction layer in ${\bar {\mathbf{w}}}_{r-1}$.
Then, with the same initial gradient parameters ${\bar {\mathbf{w}}}_{r-1}$ and \eqref{eq:condition1_b}, the input to the first BN layer (or equivalently the output of the 0-th feature extraction layer) obtained by each client in FL has the same distribution as that obtained by centralized learning, i.e.,
\begin{equation}\label{eq:pdf y1 ig}
f_{{\mathbf{Y}_i^1} |{\mathbf{X}_i^0}} (\mathbf{y}^1 |\mathbf{x}^0 \in c)
=
f_{{\mathbf{Y}_g^1} |{\mathbf{X}_g^0}} (\mathbf{y}^1 |\mathbf{x}^0 \in c)
, \,
i\in[N]
.
\end{equation}
Meanwhile, based on \eqref{eq:mu ellj}, the local batch mean of the first BN layer obtained by each client $i \in [N]$ in FL and the global one obtained by centralized learning are \eqref{eq:mu i 1} and \eqref{eq:mu g 1}, respectively.

\vspace{-0.3cm}
\begin{small}
\begin{subequations}\label{eq:mu ig 1}
\begin{equation}\label{eq:mu i 1}
\bm{\mu}_{i}^1
=
\mathbb{E}_{\mathcal{D}_{i}}[\mathbf{Y}_{i}^{1}]
=
\sum_{c}
\left( \int
\mathbf{y}^1
\cdot
f_{{\mathbf{Y}_i^1} |{\mathbf{X}_i^0}} (\mathbf{y}^1 |\mathbf{x}^0 \in c)
\cdot
d \mathbf{y}^1
\right)
\Pr (c | \mathcal{D}_i)
,
\end{equation}
\begin{equation}\label{eq:mu g 1}
\bm{\mu}_{g}^1
=
\mathbb{E}_{\mathcal{D}}[\mathbf{Y}_{g}^{1}]
=
\sum_{c}
\left( \int
\mathbf{y}^1
\cdot
f_{{\mathbf{Y}_g^1} |{\mathbf{X}_g^0}} (\mathbf{y}^1 |\mathbf{x}^0 \in c)
\cdot
d \mathbf{y}^1
\right)
\Pr (c | \mathcal{D})
.
\end{equation}
\end{subequations}
\end{small}
\vspace{-0.3cm}

\noindent
Then, combining \eqref{eq:condition1_a}, \eqref{eq:pdf y1 ig} with \eqref{eq:mu ig 1}, we have
\begin{equation}\label{eq:mu i g 1}
\bm{\mu}_{i}^1 = \bm{\mu}_{g}^1
\, , \,
i \in [N]
.
\end{equation}

Similarly, based on \eqref{eq:sigma ellj}, the local batch variance of the first BN layer in each client $i \in [N]$ and the global one in centralized learning, $(\bm{\sigma}_{i}^{1})^2$ and $(\bm{\sigma}_{g}^{1})^2$, can be obtained by \eqref{eq:mu i 1} and \eqref{eq:mu g 1} via replacing $\mathbf{y}^1$ with $( \mathbf{y}^1 - \bm{\mu}_{i}^1 )^2$ and $( \mathbf{y}^1 - \bm{\mu}_{g}^1 )^2$, respectively.
Then, with \eqref{eq:condition1_a}, \eqref{eq:pdf y1 ig} and \eqref{eq:mu i g 1}, we have
\begin{equation}\label{eq:sigma i g 1}
(\bm{\sigma}_{i}^1)^2 = (\bm{\sigma}_{g}^1)^2
\, , \,
i \in [N]
.
\end{equation}

\underline{\textbf{Induction}}:
Combining \eqref{eq:pdf y1 ig}, \eqref{eq:mu i g 1} and \eqref{eq:sigma i g 1} with \eqref{eq:batch output}, we have
\begin{equation}\label{eq:pdf x1 ig}
f_{{\mathbf{X}_i^1} |{\mathbf{X}_i^0}}(\mathbf{x}^1 |\mathbf{x}^0 \in c)
=
f_{{\mathbf{X}_g^1} |{\mathbf{X}_g^0}}(\mathbf{x}^1 |\mathbf{x}^0 \in c)
.
\end{equation}
Then, repeating the above process, we can prove that both the input and the output of each feature extraction layer obtained by each client in FL have the same distributions as those obtained by centralized learning, i.e., for $\ell \in [L+1]$,
\begin{subequations}\label{eq:pdf xy ell ig}
\begin{align}
f_{{\mathbf{X}_i^{\ell-1}} |{\mathbf{X}_i^0}}({\mathbf{x}^{\ell-1}} |{\mathbf{x}^0} \in c)
& =
f_{{\mathbf{X}_g^{\ell-1}} |{\mathbf{X}_g^0}}({\mathbf{x}^{\ell-1}} |{\mathbf{x}^0} \in c)
,
\label{eq:pdf x ell ig} \\
f_{{\mathbf{Y}_i^{\ell}} |{\mathbf{X}_i^0}}({\mathbf{y}^{\ell}} |{\mathbf{x}^0} \in c)
& =
f_{{\mathbf{Y}_g^{\ell}} |{\mathbf{X}_g^0}}({\mathbf{y}^{\ell}} |{\mathbf{x}^0} \in c)
,
\label{eq:pdf y ell ig}
\end{align}
\end{subequations}
which further results in the same global and local statistical parameters for each BN layer as
\begin{equation}\label{eq:mu sigma i g ell}
\bm{\mu}_{i}^{\ell} = \bm{\mu}_{g}^{\ell}
\, , \,
(\bm{\sigma}_{i}^{\ell})^2 = (\bm{\sigma}_{g}^{\ell})^2
\, , \,
i \in [N]
\, , \,
\ell \in [L]
.
\end{equation}
Therefore, we have $\mathcal{S}_{\mathcal{D}_i}^{r,1} = \mathcal{S}_{\mathcal{D}}^{r,1}$.
\hfill $\blacksquare$

\subsubsection{Proof of $\Delta \mathcal{S}_{\mathcal{D}_i}^{r,1} = \Delta \mathcal{S}_{\mathcal{D}}^{r,1}$}

In the backward propagation procedure of a DNN model, the gradients w.r.t. model parameters are calculated from the output to the input layer by the chain rule.
Based on this, we first analyze the relation between the local gradients w.r.t. statistical parameters and the global ones for the last BN layer, and then induce the relation for each BN layer.
To simplify the formulation, in the following proof, ${F_i}({\bar {\mathbf{w}}}_{r-1}; \mathcal{S}_{\mathcal{D}_i}^{r,1}, \Delta \mathcal{S}_{\mathcal{D}_i}^{r,1})$ and ${F}({\bar {\mathbf{w}}}_{r-1}; \mathcal{S}_{\mathcal{D}}^{r,1}, \Delta \mathcal{S}_{\mathcal{D}}^{r,1})$ are abbreviated to $F_i$ and $F$, respectively.

\underline{\textbf{Gradients w.r.t. batch variance and batch mean in the}}
\underline{\textbf{last BN layer}}:
Based on \eqref{eq:batch output} and the chain rule, the local gradient w.r.t. the batch variance in the last BN layer obtained by each client $i \in [N]$ in FL is

\vspace{-0.3cm}
\begin{small}
\begin{align}\label{eq:grad sigma Fi}
{\nabla}_{(\bm{\sigma}_{i}^{L})^2}{F_i}
= &
\underbrace{{\nabla}_{\mathbf{Y}_i^{L+1}}{F_{i}}
\cdot
{\nabla}_{\mathbf{X}_i^L}{\mathbf{Y}_i^{L+1}}
\cdot
{\nabla}_{\mathbf{\hat Y}_i^L}{\mathbf{X}_i^{L}}}_{={\nabla}_{\mathbf{\hat Y}_i^L}{F_{i}}}
\cdot
{\nabla}_{(\bm{\sigma}_{i}^{L})^2}{\mathbf{\hat Y}_i^L}
\notag \\
\overset{(a)}{=} &
\mathbb{E}_{\mathcal{D}_{i}}
\Big[
{\nabla}_{\mathbf{Y}_i^{L+1}}{\mathcal{L}_{i}}
\cdot
{\nabla}_{\mathbf{X}_i^L}{\mathbf{Y}_i^{L+1}}
\cdot
{\nabla}_{\mathbf{\hat Y}_i^L}{\mathbf{X}_i^{L}}
\cdot
{\nabla}_{(\bm{\sigma}_{i}^{L})^2}{\mathbf{\hat Y}_i^L}\Big]
\notag \\
\overset{(b)}{=} &
\sum_{c}
\bigg(
\int
\underbrace{
{
{\nabla}_{\mathbf{y}^{L+1}}{\mathcal{L}_{i}}
\cdot
{\nabla}_{\mathbf{x}^L}{\mathbf{y}^{L+1}}
\cdot
{\nabla}_{\mathbf{\hat y}^L}{\mathbf{x}^L}
\cdot
{\nabla}_{(\bm{\sigma}_{i}^{L})^2}{\mathbf{\hat y}^L}
}
}_{\text{\footnotesize (\ref{eq:grad sigma Fi}c)}}
\cdot
f_{{\mathbf{Y}_i^L}|{\mathbf{X}_i^0}}(\mathbf{y}^L|{\mathbf{x}^0} \in c)
\cdot
d \mathbf{y}^L
\bigg)
\cdot
\Pr (c | \mathcal{D}_i)
,
\end{align}
\end{small}
\vspace{-0.3cm}

\noindent
where equality (a) comes from \eqref{eq:Fi} and $\mathbf{\hat Y}_i^L$ is defined in \eqref{eq:batch output}.
In equality (b), ${\nabla}_{\mathbf{y}^{L+1}}{\mathcal{L}_{i}}$ is the Jacobian matrix of the loss function ${\mathcal{L}_{i}}$ w.r.t. ${\mathbf{y}^{L+1}}$, and other gradients with notation ``$\nabla$'' have similar meanings.
Besides, with the considered DNN model in Fig. \ref{fig:plain DNN}, we have

\vspace{-0.3cm}
\begin{small}
\begin{subequations}
\begin{align}
& {\mathbf{X}_i^L} = \bm{\gamma}^{L} \frac{ {\mathbf{Y}_i^L} - \bm{\mu}_{i}^{L}}{\sqrt{(\bm{\sigma}_{i}^{L})^{2} + \epsilon}} + \bm{\beta}^{L}
, \label{eq:xiL} \\
& {\mathbf{Y}_i^{L+1}} = G_{L} ( {\mathbf{X}_i^L}, \mathbf{\tilde w}^{L} )
, \label{eq:xiL yiL1} \\
& \mathcal{L}_{i} = \mathcal{L} ({\mathbf{Y}_i^{L+1}}, c )
, \label{eq:lossi=LG2} \\
& {\nabla}_{\mathbf{\hat Y}_i^L}{\mathbf{X}_i^L} = \operatorname{diag}(\bm{\gamma}^{L})
, \label{eq:grad x = diag gamma} \\
& {\nabla}_{(\bm{\sigma}_{i}^{L})^2}{\mathbf{\hat Y}_i^L} = \operatorname{diag}
\bigg(
\frac{ {\mathbf{Y}_i^L} - \bm{\mu}_{i}^{L}}{((\bm{\sigma}_{i}^{L})^{2} + \epsilon)^{\frac{3}{2}}}
\bigg)
, \label{eq:nabla sigma yil}
\end{align}
\end{subequations}
\end{small}
\vspace{-0.3cm}

\noindent
where the arithmetic operations are element-wise, $\text{diag}(\mathbf{z})$ is a diagonal matrix with $\mathbf{z}$ being diagonal elements, and \eqref{eq:grad x = diag gamma} and \eqref{eq:nabla sigma yil} come from \eqref{eq:batch output}.
Thus, the term (\ref{eq:grad sigma Fi}c) depends on the value of {$\mathbf{y}^L$}, $\bm{\mu}_{i}^{L}$, $(\bm{\sigma}_{i}^{L})^{2}$, $\bm{\gamma}^{L}$, $\bm{\beta}^{L}$ and $\mathbf{\tilde w}^{L}$.
For notation simplicity, we denote the multiplication of gradients in term (\ref{eq:grad sigma Fi}c) as
\begin{equation}\label{eq:grad phi i}
\text{(\ref{eq:grad sigma Fi}c)}
\triangleq
\psi_{\bm{\sigma}}^L
\left({ \mathbf{y}^L}, \bm{\mu}_{i}^{L}, (\bm{\sigma}_{i}^{L})^{2}, \bm{\gamma}^{L}, \bm{\beta}^{L}, \mathbf{\tilde w}^{L} \right)
.
\end{equation}
Meanwhile, the local gradient w.r.t. the batch mean in the last BN layer obtained by client $i \in [N]$ in FL can be expressed as

\vspace{-0.3cm}
\begin{small}
\begin{align}\label{eq:grad mu Fi simplify}
{\nabla}_{\bm{\mu}_{i}^{L}}{F_i}
\overset{(a)}{=} &
\sum_{c}
\bigg( \int
\psi_{\bm{\mu}}^L
\left(
{\mathbf{y}^{L}}, \bm{\mu}_{i}^{L}, (\bm{\sigma}_{i}^{L})^{2}, \bm{\gamma}^{L}, \bm{\beta}^{L}, \mathbf{\tilde w}^{L}
\right)
\cdot
f_{{\mathbf{Y}_i^L}|{\mathbf{X}_i^0}}(\mathbf{y}^L|{\mathbf{x}^0} \in c)
\cdot
d \mathbf{y}^L
\bigg)
\cdot
\Pr (c | \mathcal{D}_i)
,
\end{align}
\end{small}
\vspace{-0.3cm}

\noindent
where the definition of $\psi_{\bm{\mu}}^L$ closely resembles that of $\psi_{\bm{\sigma}}^L$ in \eqref{eq:grad phi i}, achieved by replacing ${\nabla}_{(\bm{\sigma}_{i}^{L})^2}{\mathbf{\hat y}^L}$ in $\text{(\ref{eq:grad sigma Fi}c)}$ with ${\nabla}_{\bm{\mu}_{i}^{L}}{\mathbf{\hat y}^L}$.
For a more detailed explanation, refer to Section \ref{sec:grad mu equal proof} of the Supplementary Material.

Similarly, in centralized learning, the global gradients w.r.t. batch variance and batch mean in the last BN layer, ${\nabla}_{(\bm{\sigma}_{g}^{L})^2}{F}$ and ${\nabla}_{\bm{\mu}_{g}^{L}}{F}$, can be obtained by \eqref{eq:grad sigma Fi} and \eqref{eq:grad mu Fi simplify}, respectively, via replacing local dataset $\mathcal{D}_i$ with global dataset $\mathcal{D}$ and subscript $i$ with $g$.
Then, based on \eqref{eq:condition1_a}, \eqref{eq:pdf y ell ig} and \eqref{eq:mu sigma i g ell}, we have
\begin{equation}\label{eq:gradient F sigma mu i g j}
{\nabla}_{(\bm{\sigma}_{i}^{L})^2}{F_i}
=
{\nabla}_{(\bm{\sigma}_{g}^{L})^2}{F}
\, , \,
{\nabla}_{\bm{\mu}_{i}^{L}}{F_i}
=
{\nabla}_{\bm{\mu}_{g}^{L}}{F}
.
\end{equation}

\underline{\textbf{Induction}}:
Similar to \eqref{eq:grad sigma Fi}, \eqref{eq:grad phi i} and \eqref{eq:grad mu Fi simplify}, for each client $i \in [N]$ in FL, the local gradient w.r.t. the batch variance in each BN layer $\ell \in [L]$ is obtained by

\vspace{-0.3cm}
\begin{small}
\begin{align}\label{eq:grad sigma ell Fi simplify}
& {\nabla}_{(\bm{\sigma}_{i}^{\ell})^2}{F_i} \notag \\
\triangleq &
 \sum_{c}
 \bigg(
 \int
 \psi_{\bm{\sigma}}^{\ell}
\Big(
\underbrace{
\mathbf{y}^{\ell},
\big\{ \bm{\mu}_{i}^{\ell'}, (\bm{\sigma}_{i}^{\ell'})^{2}, \bm{\gamma}^{\ell'}, \bm{\beta}^{\ell'}, \mathbf{\tilde w}^{\ell'} \big\}_{\ell'=\ell}^{L}
}_{\text{\footnotesize (\ref{eq:grad sigma ell Fi simplify}a)}}
\, , \,
\underbrace{
\big\{
{\nabla}_{ (\bm{\sigma}_{i}^{\ell'})^2}{F_i},
{\nabla}_{ \bm{\mu}_{i}^{\ell'}}{ F_i} \big\}_{\ell'=\ell+1}^L}_{\text{\footnotesize (\ref{eq:grad sigma ell Fi simplify}b)}}
\Big)
\cdot
f_{{\mathbf{Y}_i^\ell}|{\mathbf{X}_i^0}}(\mathbf{y}^\ell|{\mathbf{x}^0} \in c)
\cdot
d \mathbf{y}^\ell
\bigg)
\cdot \Pr (c | \mathcal{D}_i)
,
\end{align}
\end{small}
\vspace{-0.3cm}

\noindent
while the local gradient w.r.t. the batch mean is computed by

\vspace{-0.3cm}
\begin{small}
\begin{align}\label{eq:grad mu ell Fi simplify}
{\nabla}_{\bm{\mu}_{i}^{\ell}}{F_i}
\triangleq
&
\sum_{c}
\bigg(
\int
\psi_{\bm{\mu}}^{\ell}
\Big(
{\text{(\ref{eq:grad sigma ell Fi simplify}a)}}
\, , \,
{\text{(\ref{eq:grad sigma ell Fi simplify}b)}}
\Big)
\cdot
f_{{\mathbf{Y}_i^\ell}|{\mathbf{X}_i^0}}(\mathbf{y}^\ell|{\mathbf{x}^0} \in c)
\cdot
d \mathbf{y}^\ell
\bigg)
\cdot \Pr (c | \mathcal{D}_i)
,
\end{align}
\end{small}
\vspace{-0.3cm}

\noindent
where the detailed derivation of \eqref{eq:grad sigma ell Fi simplify} and \eqref{eq:grad mu ell Fi simplify} are presented in Section \ref{sec:grad sigma equal ell proof} of the Supplementary Material.
Meanwhile, the global gradients w.r.t. the batch variance and batch mean in centralized learning, ${\nabla}_{(\bm{\sigma}_{g}^{\ell})^2}{F}$ and ${\nabla}_{\bm{\mu}_{g}^{\ell}}{F}$, can be calculated by \eqref{eq:grad sigma ell Fi simplify} and \eqref{eq:grad mu ell Fi simplify}, respectively, via replacing $\mathcal{D}_i$ with $\mathcal{D}$ and replacing subscript $i$ with $g$.

\begin{remark}\label{remark:stat grad}
As indicated by \eqref{eq:grad sigma ell Fi simplify} and \eqref{eq:grad mu ell Fi simplify}, in the backward propagation procedure of a DNN model, the gradients w.r.t. statistical parameters in the $\ell$-th BN layer (i.e, ${\nabla}_{(\bm{\sigma}_{i}^{\ell})^2}{F_i}$ and ${\nabla}_{\bm{\mu}_{i}^{\ell}}{F_i}$) depend on those in its subsequent BN layers (i.e., ${\nabla}_{(\bm{\sigma}_{i}^{\ell'})^2}{F_i}$ and ${\nabla}_{\bm{\mu}_{i}^{\ell'}}{F_i}$, $\ell'=\ell+1,\ldots,L$).
\end{remark}

Finally, based on \eqref{eq:condition1_a}, \eqref{eq:pdf y ell ig}, \eqref{eq:mu sigma i g ell}, \eqref{eq:grad sigma ell Fi simplify} and\eqref{eq:grad mu ell Fi simplify}, we can iteratively prove
\begin{equation}
{\nabla}_{(\bm{\sigma}_{i}^{\ell})^2}{F_i}
=
{\nabla}_{(\bm{\sigma}_{g}^{\ell})^2}{F}
\, , \,
{\nabla}_{\bm{\mu}_{i}^{\ell}}{F_i}
=
{\nabla}_{\bm{\mu}_{g}^{\ell}}{F}
\end{equation}
holds from the last BN layer ($\ell = L$) to the first BN layer ($\ell = 1$), which leads to $\Delta \mathcal{S}_{\mathcal{D}_i}^{r,1} = \Delta \mathcal{S}_{\mathcal{D}}^{r,1}$.
\hfill $\blacksquare$

\subsection{Relation in Non-i.i.d. Data Case}\label{sec:relation S delta S noniid}

With Definition \ref{definition:iid noniid}, we have the following property for the non-i.i.d. data case.
\begin{property}
With the same initial gradient parameters ${\bar {\mathbf{w}}}_{r-1}$ of a DNN model, in the non-i.i.d. case, the statistical parameters and their gradients obtained by each client in FL and by centralized learning are generally different, i.e.,
\begin{equation}\label{ineq:SDi SD Delta SDi SD}
\mathcal{S}_{\mathcal{D}_i}^{r,1} \neq \mathcal{S}_{\mathcal{D}}^{r,1}
, \,
\Delta \mathcal{S}_{\mathcal{D}_i}^{r,1} \neq \Delta \mathcal{S}_{\mathcal{D}}^{r,1}
.
\end{equation}
\end{property}

\emph{Proof:}
Based on \eqref{eq:mu sigma ellj}, \eqref{eq:mu ig 1} and Definition \ref{definition:iid noniid}, in the non-i.i.d. data case, we generally have
\begin{equation}\label{ineq:mu sigma i g 1}
\bm{\mu}_{i}^1 \neq \bm{\mu}_{g}^1
\, , \,
(\bm{\sigma}_{i}^1)^2 \neq (\bm{\sigma}_{g}^1)^2
\, , \,
\end{equation}
which results in $\mathcal{S}_{\mathcal{D}_i}^{r,1} \neq \mathcal{S}_{\mathcal{D}}^{r,1}$.
Next, based on \eqref{eq:batch input}, if the input to the DNN model in client $i$ has different distribution from the one in centralized learning, namely $f_{{\mathbf{X}_i^0}}(\mathbf{x}^0 | \mathbf{x}^0 \in c) \neq f_{{\mathbf{X}_g^0}}(\mathbf{x}^0 | \mathbf{x}^0 \in c)$, then the input to the first BN layer obtained by client $i$ also has different distribution from that by centralized learning, i.e.,
\begin{equation}\label{ineq:pdf y1 ig}
f_{{\mathbf{Y}_i^1} |{\mathbf{X}_i^0}} (\mathbf{y}^1 |\mathbf{x}^0 \in c)
\neq
f_{{\mathbf{Y}_g^1} |{\mathbf{X}_g^0}} (\mathbf{y}^1 |\mathbf{x}^0 \in c)
.
\end{equation}
Thus, in the non-i.i.d. data case, we will have \eqref{ineq:pdf y1 ig} or $ \Pr (c | \mathcal{D}_i) \neq \Pr (c | \mathcal{D})$ or both.
Combining this with \eqref{eq:grad sigma ell Fi simplify}, \eqref{eq:grad mu ell Fi simplify} and \eqref{ineq:mu sigma i g 1}, we have
\begin{equation}\label{ineq:gradient F sigma mu i g j}
{\nabla}_{(\bm{\sigma}_{i}^{1})^2}{F_i}
\neq
{\nabla}_{(\bm{\sigma}_{g}^{1})^2}{F}
\, , \,
{\nabla}_{\bm{\mu}_{i}^{1}}{F_i}
\neq
{\nabla}_{\bm{\mu}_{g}^{1}}{F}
\end{equation}
in general, which leads to $\Delta \mathcal{S}_{\mathcal{D}_i}^{r,1} \neq \Delta \mathcal{S}_{\mathcal{D}}^{r,1}$.
\hfill $\blacksquare$

\subsection{Necessity of $\Delta \mathcal{S}_{\mathcal{D}_i}^{r,1} = \Delta \mathcal{S}_{\mathcal{D}}^{r,1}$ for Convergence Guarantee}\label{sec:necessity equal stat grad}

As seen, the non-i.i.d. data brings heterogeneous statistical parameters among clients.
Based on this, in the training process of FL, \texttt{FedDNA} \cite{duan2021feddna} considers making all clients possess the same statistical parameters in the forward propagation procedure,
i.e., $\mathcal{S}_{\mathcal{D}_i}^{r,1} = \mathcal{S}_{\mathcal{D}}^{r,1}$, to improve FL performance.
However, we find that this idea cannot guarantee the convergence of FL as expounded in the following Property \ref{property:noniid necessity}.
\begin{property}\label{property:noniid necessity}
With the initial gradient parameters ${\bar {\mathbf{w}}}_{r-1}$ of a DNN model, if we only keep the local and the global statistical parameters consistent (i.e., $\mathcal{S}_{\mathcal{D}_i}^{r,1} = \mathcal{S}_{\mathcal{D}}^{r,1}$) but do not ensure their gradients w.r.t. statistical parameters the same, then for the non-i.i.d. data case, we generally have
\begin{equation}\label{ineq:nabla w Fi F}
\nabla_{\mathbf{w}} F_i({\bar {\mathbf{w}}}_{r-1}; \mathcal{S}_{\mathcal{D}_i}^{r,1}, \Delta \mathcal{S}_{\mathcal{D}_i}^{r,1})
\neq
\nabla_{\mathbf{w}} F_i ({\bar {\mathbf{w}}}_{r-1}; \mathcal{S}_{\mathcal{D}}^{r,1}, \Delta \mathcal{S}_{\mathcal{D}}^{r,1})
,
\end{equation}
which leads to $B_i > 0$ in Theorem \ref{theorem:1} and biases the convergence of FL.
\end{property}

The detailed proof is given in Section \ref{sec:noniid necessity proof} of the Supplementary Material.
Based on the above discussion, it can be found that in order to guarantee the convergence of FL with BN under different data distributions, $\mathcal{S}_{\mathcal{D}_i}^{r,1} = \mathcal{S}_{\mathcal{D}}^{r,1}$ alone is insufficient and $\Delta \mathcal{S}_{\mathcal{D}_i}^{r,1} = \Delta \mathcal{S}_{\mathcal{D}}^{r,1}$ is necessary.

\section{Proposed FedTAN}\label{Sec:FedTAN}

\subsection{Overall Procedure}

It has been suggested by Assumption \ref{assumption:BN disturbance} and Theorem \ref{theorem:1} that FL with BN can converge properly if each client $i \in [N]$ satisfies
\begin{equation}\label{eq:SDi=SD delta SDi=SD}
\mathcal{S}_{\mathcal{D}_i}^{r,1} = \mathcal{S}_{\mathcal{D}}^{r,1} \, , \,
\Delta \mathcal{S}_{\mathcal{D}_i}^{r,1} = \Delta \mathcal{S}_{\mathcal{D}}^{r,1}
.
\end{equation}
Inspired by this, we develop a FL algorithm named \texttt{FedTAN}.
Specifically, for each iteration of \texttt{FedAvg}, the first ($t=1$) local updating step for each client is modified, where the layer-wise aggregations of local statistical parameters and of their gradients are implemented after \eqref{eq:local SGD_a} to achieve \eqref{eq:SDi=SD delta SDi=SD}.
The detailed procedure of \texttt{FedTAN} is summarized in Algorithm \ref{algorithm:FedTAN procedure}.

\begin{algorithm}[t]
\caption{\texttt{FedTAN}: FL algorithm tailored for BN}
\begin{algorithmic}[1]
\State Initialize global model $\{ {\bar {\mathbf{w}}}_{0}, {\bar {\mathcal{S}}}_{0} \}$ by the server.
\For{$r=1,2,\ldots,R$}
\State Server sends global model $\{ {\bar {\mathbf{w}}}_{r-1}, {\bar {\mathcal{S}}}_{r-1} \}$ to clients;
\State Each client initializes local model $\{\mathbf{w}^{r,0}_{i},{\bar {\mathcal{S}}}^{r,0}_{\mathcal{D}_i}\}$ by \eqref{eq:local SGD_a};
\For {$t = 1$} \textbf{(Modified local updating step)}
\State
Clients update local statistical parameters ${\mathcal{S}}^{r,1}_{\mathcal{D}_i}$,
$\forall i \in [N]$ by \textbf{Algorithm \ref{algorithm:FedTAN modification forward}};
\State Clients update local gradients $\Delta {\mathcal{S}}^{r,1}_{\mathcal{D}_i}$ and obtain local gradients w.r.t. $\mathbf{w}^{r,0}_{i}$, $\nabla_{\mathbf{w}} F_{i}$, by
\Statex \qquad\quad \textbf{Algorithm \ref{algorithm:FedTAN modification backward}};
\State Each client updates local model $\{\mathbf{w}^{r,1}_{i},{\bar {\mathcal{S}}}^{r,1}_{\mathcal{D}_i}\}$ by
\eqref{eq:local SGD_b} and \eqref{eq:moving avg};
\EndFor
\For {client ${i} \in [N]$} (\textbf{in parallel})
\For {$t = 2, \ldots, E$}
\State Update local model $\{\mathbf{w}^{r,t}_{i}, {\bar {\mathcal{S}}}^{r,t}_{\mathcal{D}_i}\}$ by \eqref{eq:local SGD_b} and \eqref{eq:moving avg};
\EndFor
\State Send updated local model $\{ \mathbf{w}_i^{r,E}, {\bar {\mathcal{S}}}^{r,E}_{\mathcal{D}_i}\}$ to server;
\EndFor
\State Server updates the global model by \eqref{eq:global model};
\EndFor
\end{algorithmic}
\label{algorithm:FedTAN procedure}
\end{algorithm}

\subsection{Layer-wise Aggregation in Forward Propagation}

Given the gradient parameters ${\bar {\mathbf{w}}}_{r-1}$ and the input data $\mathbf{Y}_{i}^{\ell}$ to the BN layer $\ell$, each client $i \in [N]$ computes the local batch mean $\bm{\mu}_{i}^{\ell}$ by \eqref{eq:mu sigma ellj} and uploads it to the server.
Then, the server aggregates the local batch means from clients to obtain the global batch mean by
\begin{equation}\label{eq:bar mu}
{\bar{\bm{\mu}}}^{\ell} = \sum_{i=1}^N p_i {\bm{\mu}}_{i}^{\ell}
,
\end{equation}
and returns ${\bar{\bm{\mu}}}^{\ell}$ back to the clients.
Next, each client uses ${\bar{\bm{\mu}}}^{\ell}$ to calculate $(\bm{\sigma}_{i}^{\ell})^2$ and uploads to the server.
After receiving the global batch variance $({\bar{\bm{\sigma}}}^{\ell})^2$ from the server, where
\begin{equation}\label{eq:bar sigma}
({\bar{\bm{\sigma}}}^{\ell})^2 = \sum_{i = 1}^N p_i (\bm{\sigma}_{i}^{\ell})^2
,
\end{equation}
each client $i$ calculates the output of the BN layer, $\mathbf{X}_{i}^{\ell}$, by \eqref{eq:batch output}.
Each BN layer $\ell \in \{L\}$ is sequentially processed using the above steps, where $L$ is the number of BN layers.
A detailed description of the execution steps is presented in Algorithm \ref{algorithm:FedTAN modification forward}.
\begin{algorithm}[t]
\caption{Modified forward propagation}
\begin{algorithmic}[1]
\For {${\ell} =1,\ldots,L$}
\For {client ${i} \in [N]$} (\textbf{in parallel})
\State Calculate local batch mean $\bm{\mu}_{i}^{\ell} = \{ \mu_{i}^{\ell, j} \}$ by \eqref{eq:mu ellj};
\State Upload $\bm{\mu}_{i}^{\ell}$ to the server;
\EndFor
\State Server computes global batch mean ${\bar{\bm{\mu}}}^{\ell}$ by \eqref{eq:bar mu}, and
sends it to each client;
\For {client ${i} \in [N]$} (\textbf{in parallel})
\State $\bm{\mu}_{i}^{\ell} \leftarrow {\bar{\bm{\mu}}}^{\ell}$;
\State Obtain batch variance $(\bm{\sigma}_{i}^{\ell})^2 = \{ (\sigma_{i}^{\ell, j})^2 \}$ by \eqref{eq:sigma ellj};
\State Upload $(\bm{\sigma}_{i}^{\ell})^2$ to the server;
\EndFor
\State Server computes global batch variance by \eqref{eq:bar sigma}, and
sends $({\bar{\bm{\sigma}}}^{\ell})^2$ to each client;
\For {client ${i} \in [N]$} (\textbf{in parallel})
\State $(\bm{\sigma}_{i}^{\ell})^2 \leftarrow ({\bar{\bm{\sigma}}}^{\ell})^2$;
\State Calculate output of the $\ell$-th BN layer by \eqref{eq:batch output};
\EndFor
\EndFor
\renewcommand{\algorithmicensure}{\textbf{Output:}}
\Ensure Modified statistical parameters $\{\bm{\mu}_{i}^{\ell}, (\bm{\sigma}_{i}^{\ell})^2\}_{\ell = 1}^L$.
\end{algorithmic}
\label{algorithm:FedTAN modification forward}
\end{algorithm}

\begin{remark}
It can be easily proved that ${\bar{\bm{\mu}}}^{\ell}$ in \eqref{eq:bar mu} and $({\bar{\bm{\sigma}}}^{\ell})^2$ in \eqref{eq:bar sigma} equal to the global batch mean $\bm{\mu}_{g}^{\ell}$ and batch variance $(\bm{\sigma}_{g}^{\ell})^2$ obtained by centralized learning, respectively.
Thus, with resetting $\bm{\mu}_{i}^{\ell} = {\bar{\bm{\mu}}}^{\ell}$ and $(\bm{\sigma}_{i}^{\ell})^2 = ({\bar{\bm{\sigma}}}^{\ell})^2$ in each client, $\mathcal{S}_{\mathcal{D}_i}^{r,1} = \mathcal{S}_{\mathcal{D}}^{r,1}$ can hold.
However, according to Property \ref{property:noniid necessity}, the above modified forward propagation procedure still cannot guarantee $\Delta \mathcal{S}_{\mathcal{D}_i}^{r,1} = \Delta \mathcal{S}_{\mathcal{D}}^{r,1}$, thus the following modified backward propagation procedure is required.
\end{remark}

\subsection{Layer-wise Aggregation in Backward Propagation}

According to Remark \ref{remark:stat grad}, $\nabla_{\bm{\mu}_{i}^{\ell}} F_{i}$ and $\nabla_{(\bm{\sigma}_{i}^{\ell})^2} F_{i}$ in the $\ell$-th BN layer depend on those in its subsequent BN layers and they have to be computed in a backward manner.
Hence, after Algorithm \ref{algorithm:FedTAN modification forward}, the local gradients $\nabla_{\bm{\mu}_{i}^{\ell}} F_{i}$ and $\nabla_{(\bm{\sigma}_{i}^{\ell})^2} F_{i}$ in each client are uploaded to the server for average aggregation by
\begin{equation}\label{eq:sigma mu bar F}
\big\{ \nabla_{(\bm{\bar \sigma}^{\ell})^2} {\bar F}, \nabla_{\bm{\bar \mu}^{\ell}} {\bar F} \big\}
= \sum_{i=1}^{N} p_{i} \big\{ \nabla_{(\bm{\sigma}_{i}^{\ell})^2} F_{i}, \nabla_{\bm{\mu}_{i}^{\ell}} F_{i} \big\},
\end{equation}
which are then sent back to the clients in a layer-by-layer manner.
The detailed procedure is concluded in Algorithm \ref{algorithm:FedTAN modification backward}, which achieves $\Delta \mathcal{S}_{\mathcal{D}_i}^{r,1} = \Delta \mathcal{S}_{\mathcal{D}}^{r,1}$, $\forall i\in [N]$.
\begin{algorithm}[t]
\caption{Modified backward propagation}
\begin{algorithmic}[1]
\For {${\ell} =L,\ldots,1$}
\For {client ${i} \in [N]$} (\textbf{in parallel})
\State Calculate $\nabla_{\{ \mathbf{\tilde w}^{\ell},\bm{\gamma}^{\ell},\bm{\beta}^{\ell} \}} F_{i}$ by chain rule, $\nabla_{(\bm{\sigma}_{i}^{\ell})^2} F_{i}$
by \eqref{eq:grad sigma ell Fi simplify} and $\nabla_{\bm{\mu}_{i}^{\ell}} F_{i}$ by \eqref{eq:grad mu ell Fi simplify};
\State Upload $\{ \nabla_{(\bm{\sigma}_{i}^{\ell})^2} F_{i}, \nabla_{\bm{\mu}_{i}^{\ell}} F_{i} \}$ to the server;
\EndFor
\State Server computes the average $\{ \nabla_{(\bm{\bar \sigma}^{\ell})^2}{\bar F} , \nabla_{\bm{\bar \mu}^{\ell}} {\bar F} \}$ by
\eqref{eq:sigma mu bar F}, and sends it to clients;
\For {client ${i} \in [N]$} (\textbf{in parallel})
\State $\{ \nabla_{(\bm{\sigma}_{i}^{\ell})^2} F_{i}, \nabla_{\bm{\mu}_{i}^{\ell}} F_{i} \}
\leftarrow
\{ \nabla_{(\bm{\bar \sigma}^{\ell})^2} {\bar F}, \nabla_{\bm{\bar \mu}^{\ell}} {\bar F} \}$;
\If{$\ell = 1$}
\State Calculate ${\nabla}_{\mathbf{\tilde w}^{0}}{F_i}$ by the chain rule;
\EndIf
\EndFor
\EndFor
\renewcommand{\algorithmicensure}{\textbf{Output:}}
\Ensure Local gradients $\{\nabla_{\mathbf{\tilde w}^{\ell}} F_{i}\}_{\ell = 0}^{L}$, $\{\nabla_{\bm{\gamma}^{\ell}} F_{i}, \nabla_{\bm{\beta}^{\ell}} F_{i}\}_{\ell = 1}^L$.
\end{algorithmic}
\label{algorithm:FedTAN modification backward}
\end{algorithm}

\begin{remark}
We note that exchanging BN parameters between clients and the server would pose a privacy risk, as attackers can exploit BN parameter knowledge to reconstruct private data more accurately.
To mitigate this, adopting a larger batch size (typically $\geq$ 32) and implementing defensive mechanisms like secure multiparty computation, homomorphic encryption, and differential privacy are recommended \cite{huang2021evaluating}.
However, since this paper primarily focuses on investigating the impact of BN on FL, privacy protection will not be extensively discussed.
\end{remark}

It should be noted that regardless of the number of local updating steps $E$, the layer-wise aggregation procedure above is performed only once per iteration.
Additionally, during this aggregation, only the statistical parameters and their gradients are exchanged between the server and each client, with the statistical parameters in BN layers occupying a tiny portion of the total model size \cite{mills2021multi}.
Thus, although \texttt{FedTAN} requires a total of $(3L+1)$ communication rounds for each iteration, the extra data exchanged between the server and clients is negligible, especially when training a large-scale DNN.

\subsection{Enhanced \texttt{FedTAN} for Reducing Communication Rounds}\label{Sec:FedTAN-II}

To reduce communication rounds, we enhance \texttt{FedTAN} by reducing the execution times of layer-wise aggregations in Algorithms \ref{algorithm:FedTAN modification forward} and \ref{algorithm:FedTAN modification backward}.
Inspired by \cite{vsajina2023peer} and \cite{zhong2023making}, wherein a portion of model parameters is frozen after a certain number of FL iterations, allowing only the remaining parameters to be updated, we propose a two-step approach as follows:
\begin{enumerate}[(i)]
\item
Firstly, we run \texttt{FedTAN} in Algorithm \ref{algorithm:FedTAN procedure} for $M$ iterations, obtaining global statistical parameters ${\bar {\mathcal{S}}}_{M}$.
\item
Next, we fix the statistical parameters in BN layers using ${\bar {\mathcal{S}}}_{M}$ and only update the gradient parameters ${\mathbf{w}}$ with the classic \texttt{FedAvg} algorithm in subsequent iterations.
\end{enumerate}
This two-step strategy yields notable benefits.
By executing \texttt{FedTAN} initially, we acquire more precise statistical parameters for BN layers.
Subsequently, applying \texttt{FedAvg} in later iterations avoids the need for layer-wise aggregations, effectively reducing communication rounds.
We denote this communication-efficient scheme of \texttt{FedTAN} as \texttt{FedTAN-II}.

\section{Experimental results}\label{sec:experiment results}

\subsection{Parameter Setting}\label{sec:Parameter Setting}

\subsubsection{Datasets and DNN models}

In the experiments, we train DNN models under the following four datasets.
\begin{itemize}
\item
\textbf{CIFAR-10 dataset} \cite{krizhevsky2009learning}:
We assume that the server coordinates $N=5$ clients to train the ResNet-20 with BN \cite{he2016deep} for image classification.
The dataset is divided into two types: i.i.d. and non-i.i.d.
In the i.i.d. data case, the 50000 training samples in the CIFAR-10 dataset are shuffled and randomly distributed among the clients, while in the non-i.i.d. data case, each client is assigned only two classes of training samples.
For both data cases, each client receives an equal number of training samples, and we evaluate the testing accuracy of global model using the CIFAR-10 testing dataset containing 10000 samples with ten classes.

\item
\textbf{CIFAR-100 dataset} \cite{krizhevsky2009learning}:
The data partition for the i.i.d. data case aligns with the CIFAR-10 experiment.
In the non-i.i.d. data case, each client possesses twenty classes of training samples.
For this dataset, we use the ResNet-20 with BN.

\item
\textbf{MNIST dataset} \cite{lecun1998gradient}:
In both the i.i.d. and non-i.i.d. data cases, we adopt the data partition similar to the CIFAR-10 experiment, and employ a 3-layer DNN with dimensions $784 \times 30 \times 10$, where the hidden layer with 30 neurons is followed by a BN layer.

\item
\textbf{Office-Caltech10 dataset} \cite{gong2012geodesic}:
This dataset includes ten overlapping image classes from four different data sources: Amazon, Caltech, DSLR, and Webcam.
We assume $N=4$ clients, and each client possesses training samples from one data source.
For this dataset, we train the AlexNet \cite{li2021fedbn} model.
\end{itemize}

\subsubsection{Parameter values}

To update local models, mini-batch SGD is used with following parameter values (if not specified).

\begin{itemize}
\item {\textbf{CIFAR-10}:}
We adopt the parameter value from \cite{Idelbayev18a} with fine-tuning on the learning rate $\gamma$.
Specifically, the batch size is 128, and $\gamma$ starts at 0.5, decreasing to 0.05 after 6000 iterations.
If momentum ($=0.9$) and weight decay ($=10^{-4}$) are used in mini-batch SGD, $\gamma$ further decreases to 0.005 after 8000 iterations.

\item {\textbf{CIFAR-100}:}
Following \cite{weiaicunzai2022Practice}, we use the same batch size and momentum as the CIFAR-10, with weight decay set to $5\times10^{-4}$.
Meanwhile, $\gamma$ starts at 0.1 and is subsequently divided by 5 after 4000, 6000, and 8000 iterations.

\item {\textbf{MNIST}:}
Using empirical values, batch size is set to 128, and $\gamma$ is fixed at 0.5 without momentum or weight decay.

\item {\textbf{Office-Caltech10}:}
We use the identical parameter setting as described in \cite{li2021fedbn}, with a batch size of 32 and $\gamma$ set to 0.01, without applying momentum or weight decay.

\end{itemize}

Additionally, the momentum of the moving average, denoted as $\rho$ in equation \eqref{eq:moving avg}, is set to 0.1, and the number of local updating steps $E$ is 5.
The DNN parameters are stored and transmitted as 32-bit floating point numbers, and the reported results are an average of five independent experiments\footnote{The codes are available at \url{https://github.com/wangyanmeng/FedTAN}.}.

\subsubsection{Baselines}

\begin{figure}[t]
\centering
\subfigure[i.i.d.]{
\includegraphics[width= 2 in ]{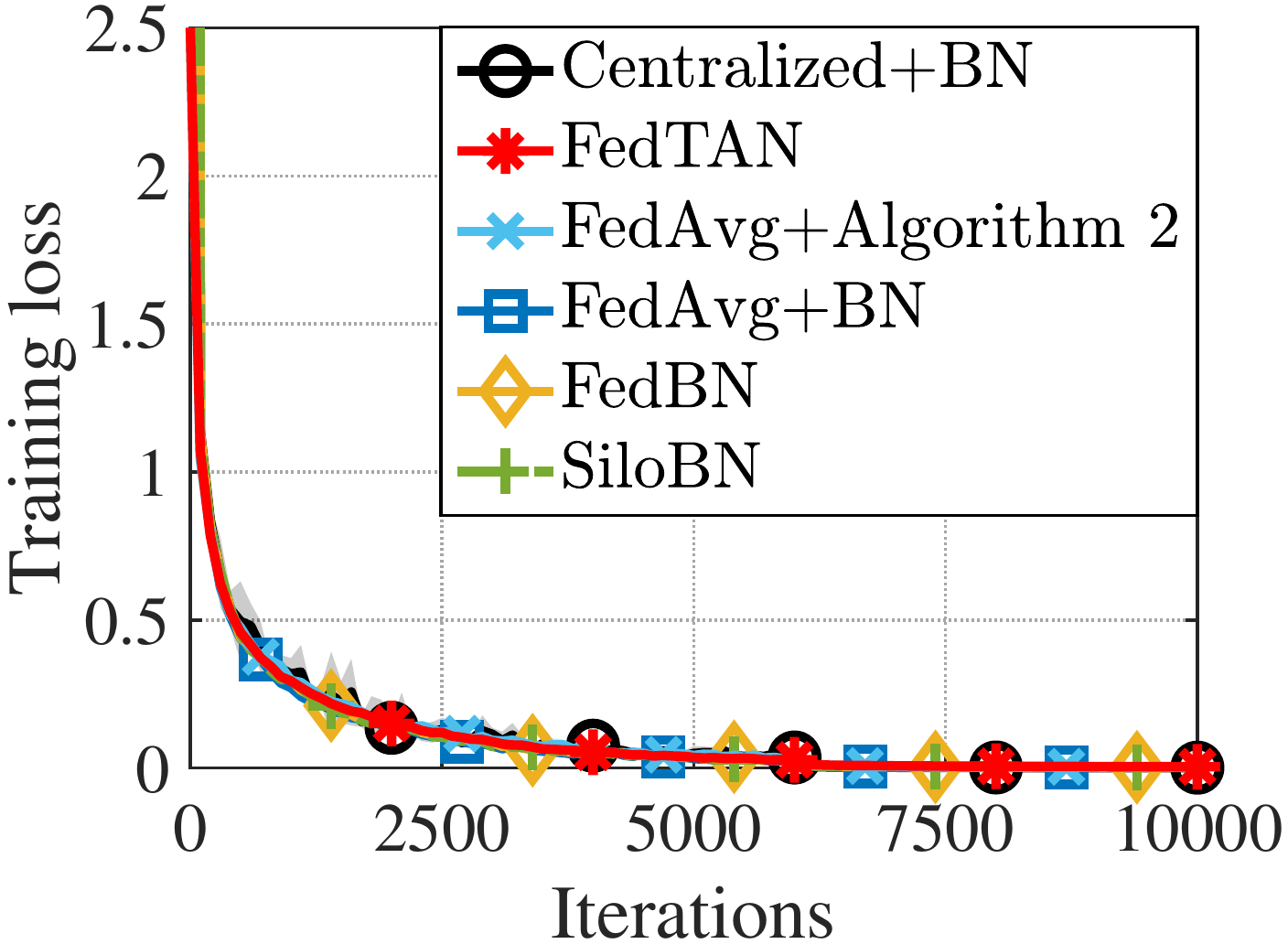}}
\subfigure[i.i.d.]{
\includegraphics[width= 2 in ]{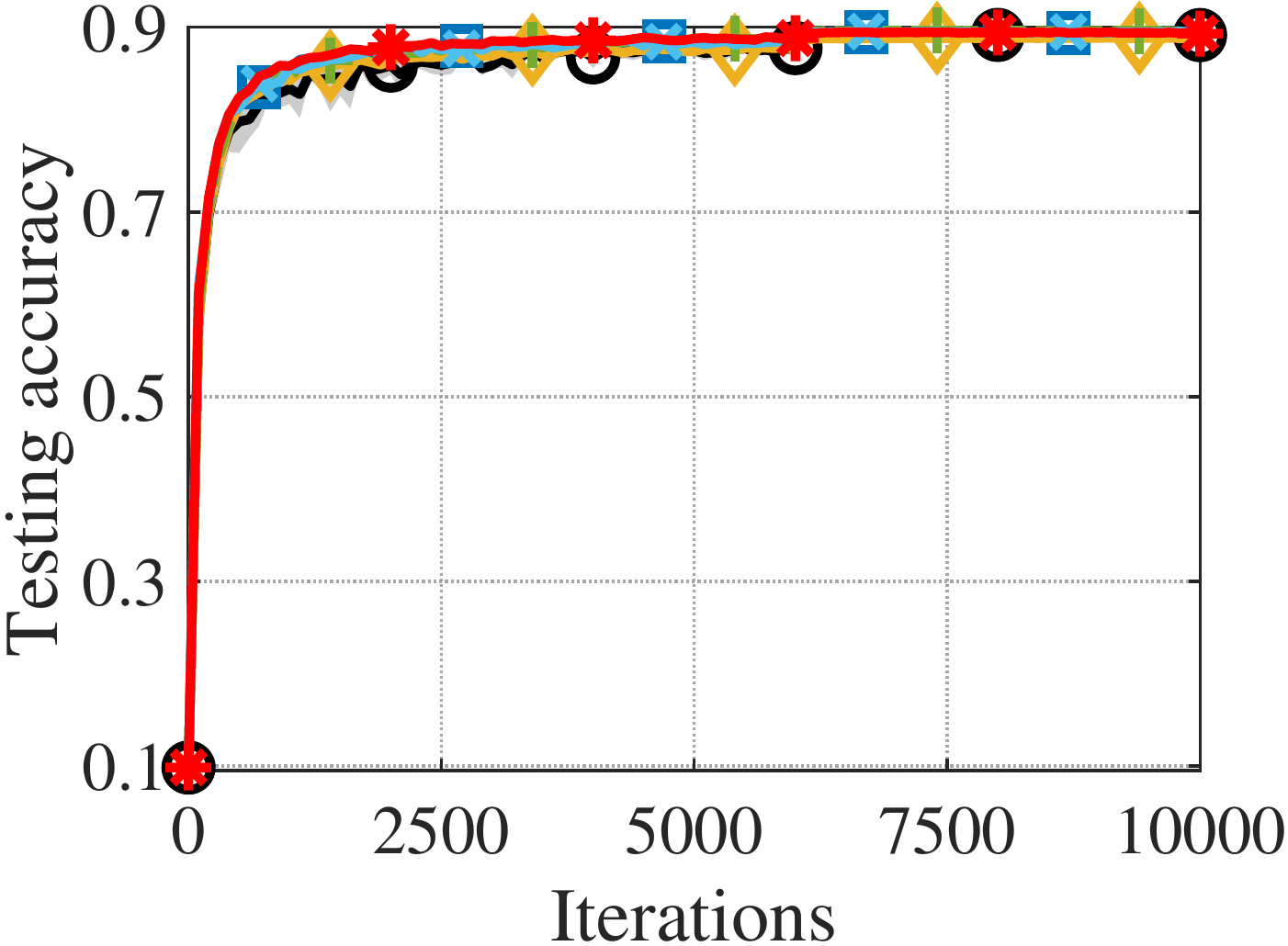}}\\
\subfigure[Non-i.i.d.]{
\includegraphics[width= 2 in ]{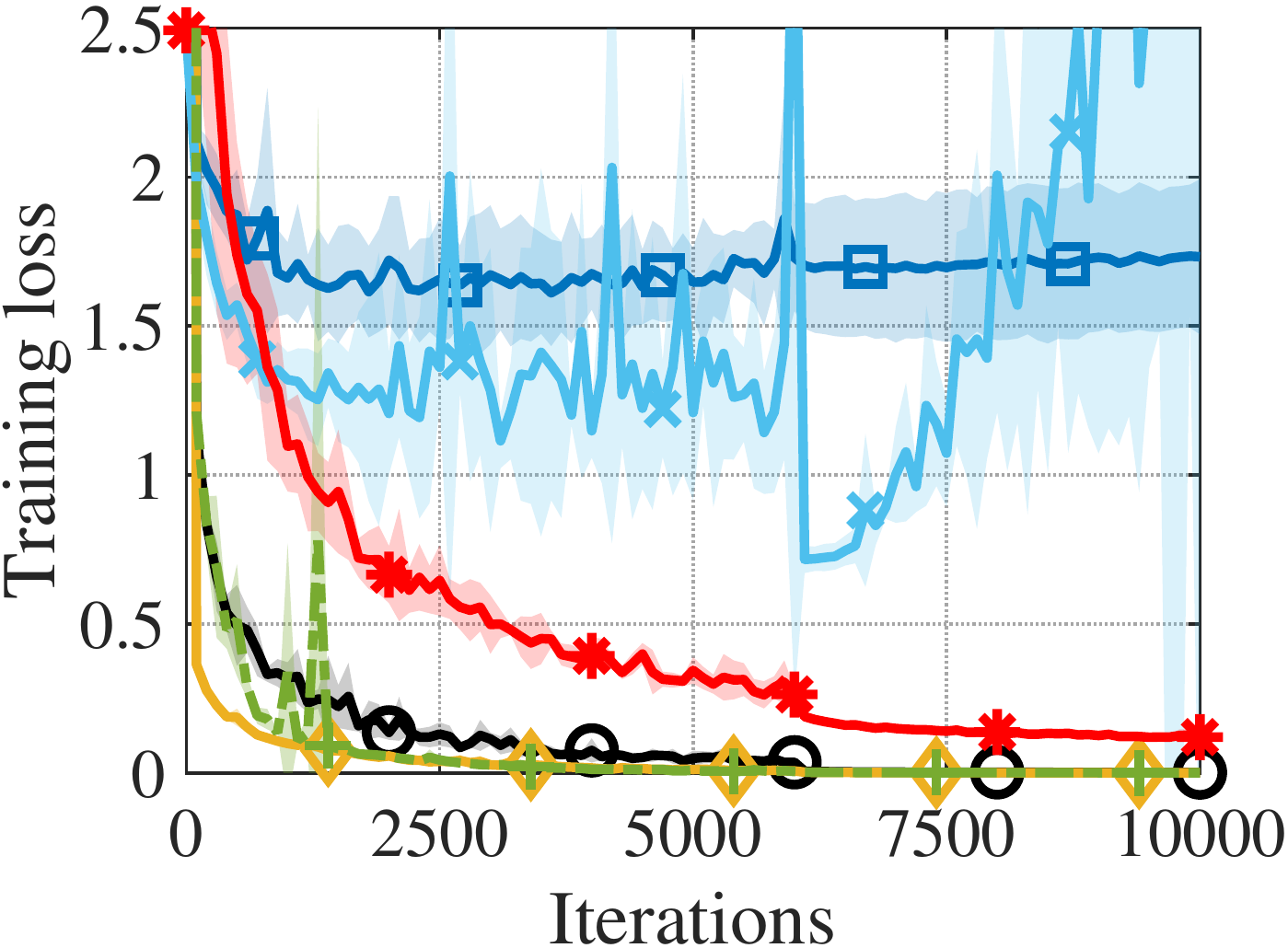}}
\subfigure[Non-i.i.d.]{
\includegraphics[width= 2 in ]{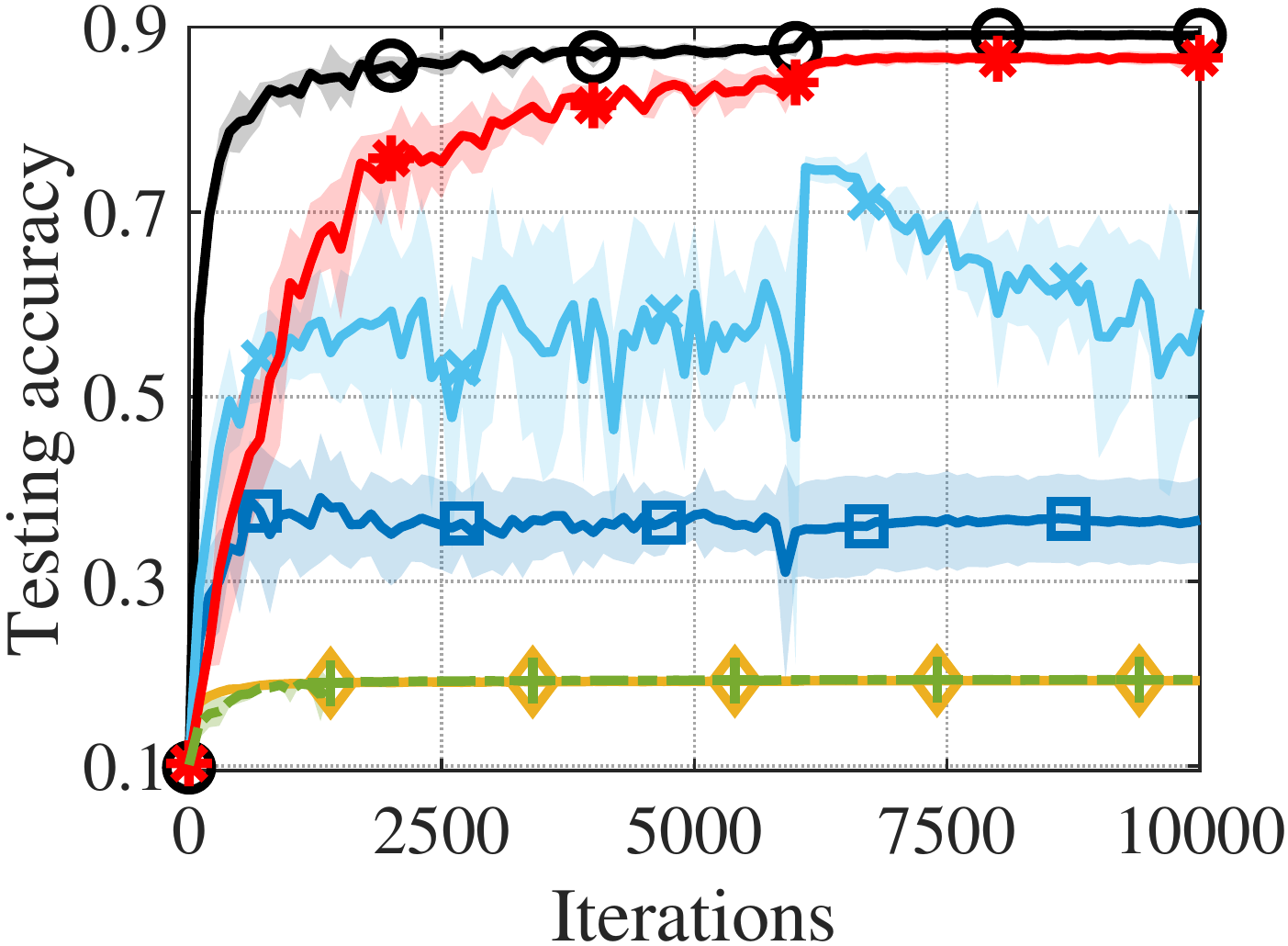}}
\caption{Performance of different FL schemes on CIFAR-10 dataset. The mini-batch SGD without momentum or weight decay is adopted.}
\label{fig:performance iid noniid}
\end{figure}

\begin{table}[t]
\small
\centering
\caption{Testing accuracy of different FL schemes on CIFAR-10 dataset. The mini-batch SGD with momentum and weight decay is adopted.}\label{table:testing accuracy momentum}
\begin{tabular}{c|cc}
\hline
\textbf{FL scheme} & \textbf{i.i.d. (\%)}
& \textbf{Non-i.i.d. (\%)} \\ \hline
\texttt{Centralized+BN} & 91.53 $\pm$ 0.50 & 91.53 $\pm$ 0.50 \\
\texttt{FedTAN} & 91.26 $\pm$ 0.21 & \textbf{87.66} $\pm$ 0.43 \\
\texttt{FedAvg+BN} & 91.35 $\pm$ 0.24 & 45.90 $\pm$ 6.39 \\
\texttt{FedAvg+Algorithm \ref{algorithm:FedTAN modification forward}} & 91.42 $\pm$ 0.15 & 76.01 $\pm$ 1.37 \\
\texttt{FedBN} & 90.74 $\pm$ 0.28 & 19.46 $\pm$ 0.05 \\
\texttt{SiloBN} & 91.31 $\pm$ 0.29 & 19.43 $\pm$ 0.09 \\ \hline
\end{tabular}
\end{table}

Six baselines and a centralized learning scheme are considered for comparison with \texttt{FedTAN}.
\begin{itemize}
\item
\textbf{\texttt{FedAvg+BN}}:
This scheme uses the vanilla \texttt{FedAvg} algorithm to train ResNet-20 with BN.
\item
\textbf{\texttt{FedAvg+Algorithm \ref{algorithm:FedTAN modification forward}}}:
We consider a simplified implementation of \texttt{FedDNA} in \cite{duan2021feddna} by assuming $\mathcal{S}_{\mathcal{D}_i}^{r,1} = \mathcal{S}_{\mathcal{D}}^{r,1}$, $\forall i\in [N]$.
However, \texttt{FedDNA} does not consider the deviation $\Delta \mathcal{S}_{\mathcal{D}_i}^{r,1} \neq \Delta \mathcal{S}_{\mathcal{D}}^{r,1}$.
\item
\textbf{\texttt{FedBN}} \cite{li2021fedbn}:
This scheme updates all parameters in BN layers locally while uploading model coefficients in other layers to the server for global aggregation.
\item
\textbf{\texttt{SiloBN}} \cite{andreux2020siloed}:
This scheme updates only the statistical parameters in BN locally while uploading all gradient parameters to the server for global aggregation.
\item
\textbf{\texttt{FedAvg+GN}} \cite{hsieh2020non}:
As an alternative to BN, GN is less sensitive to data distribution and is also widely used.
This scheme employs \texttt{FedAvg} to train ResNet-20 with GN.
\item
\textbf{\texttt{SingleNet}} \cite{li2021fedbn}:
This scheme updates all parameters locally without exchanging any parameter with the server.
\item
\textbf{\texttt{Centralized+BN}}:
In centralized learning, the global dataset is used to train ResNet-20 with BN, which serves as the performance upper bound in the simulations.
\end{itemize}

\subsection{Robustness under Different Data Distributions}\label{sec:Robustness Data Distributions}

\subsubsection{The i.i.d. data case}

Fig. \ref{fig:performance iid noniid} compares the training loss and testing accuracy of different FL schemes on CIFAR-10 dataset by adopting mini-batch SGD without momentum or weight decay in local model updating.
Each curve and filled area represent the mean and standard variance of five independent experimental results.
\texttt{FedBN} and \texttt{SiloBN}, unlike \texttt{FedAvg}, keep part or all of BN parameters updating locally, and would possess different local models after global aggregation.
Thus, we estimate the local training loss of each client in \texttt{FedBN} or \texttt{SiloBN} by its model and dataset, while assessing the generalization ability of each local model based on the 10000 samples in the CIFAR-10 testing dataset.
Afterward, the training loss and testing accuracy of these two FL schemes are calculated by averaging the local results of all clients.

According to Fig. \ref{fig:performance iid noniid}(a) and \ref{fig:performance iid noniid}(b), in the i.i.d. data case, all FL schemes are comparable to the centralized learning method.
Specifically, in the i.i.d. case, $\mathcal{S}_{\mathcal{D}_i}^{r,1} = \mathcal{S}_{\mathcal{D}}^{r,1}$ and $\Delta \mathcal{S}_{\mathcal{D}_i}^{r,1} = \Delta \mathcal{S}_{\mathcal{D}}^{r,1}$ are true as discussed in Property \ref{property:sD delta sD}.
As a result, the gradient deviation under BN $B_i=0$ in Theorem \ref{theorem:1}, and the learned model by FL with BN can converge in the right direction.

\subsubsection{The non-i.i.d. data case}

Comparing Fig. \ref{fig:performance iid noniid}(a) and 4(b) with Fig. \ref{fig:performance iid noniid}(c) and 4(d), we can find that the non-i.i.d. data damages the performance of all FL schemes, but the proposed \texttt{FedTAN} still performs closely to the centralized learning scheme and outperforms other FL benchmarks.
Specifically, Fig. \ref{fig:performance iid noniid}(c) and \ref{fig:performance iid noniid}(d) show that both \texttt{FedBN} and \texttt{SiloBN} exhibit good training performance on the local datasets with only two classes of samples, even converging faster than the centralized learning scheme, but have poor testing performance on the entire testing dataset with ten classes of samples.
This observation indicates that the DNN models learned by these two FL schemes are local optimal solutions and cannot be generalized.
The reason for this is that \texttt{FedBN} assumes the clients possess i.i.d. label distributions, while \texttt{SiloBN} considers each client has all classes of samples locally, which does not hold in our setting on non-i.i.d. data.
In addition, \texttt{FedAvg+Algorithm \ref{algorithm:FedTAN modification forward}} improves FL performance significantly over the vanilla \texttt{FedAvg+BN}, taking into account the mismatched statistical parameters.
However, \texttt{FedAvg+Algorithm \ref{algorithm:FedTAN modification forward}} ignores the deviation $\Delta \mathcal{S}_{\mathcal{D}_i}^{r,1} \neq \Delta \mathcal{S}_{\mathcal{D}}^{r,1}$, and thus it fluctuates widely and still fails to converge to a satisfactory result.
In contrast to the FL benchmarks above, through the layer-wise aggregations on statistical parameters in Algorithm \ref{algorithm:FedTAN modification forward} and on their gradients in Algorithm \ref{algorithm:FedTAN modification backward}, \texttt{FedTAN} is capable of meeting the condition \eqref{eq:SDi=SD delta SDi=SD} to achieve superior performance.

We further demonstrate the effectiveness of proposed \texttt{FedTAN} by adopting mini-batch SGD with momentum and weight decay.
Table \ref{table:testing accuracy momentum} compares the testing accuracy of different FL schemes, based on the mean and standard deviation of five independent experiments.
As can be observed, \texttt{FedTAN} performs well for both data cases, while other FL benchmarks significantly degrade in the non-i.i.d. data case.
In particular, under the non-i.i.d. data, \texttt{FedTAN} increases its average testing accuracy by 41.76\% compared to \texttt{FedAvg+BN}.

\begin{figure}[t]
\centering
\subfigure[Training loss.]{
\includegraphics[width= 2 in ]{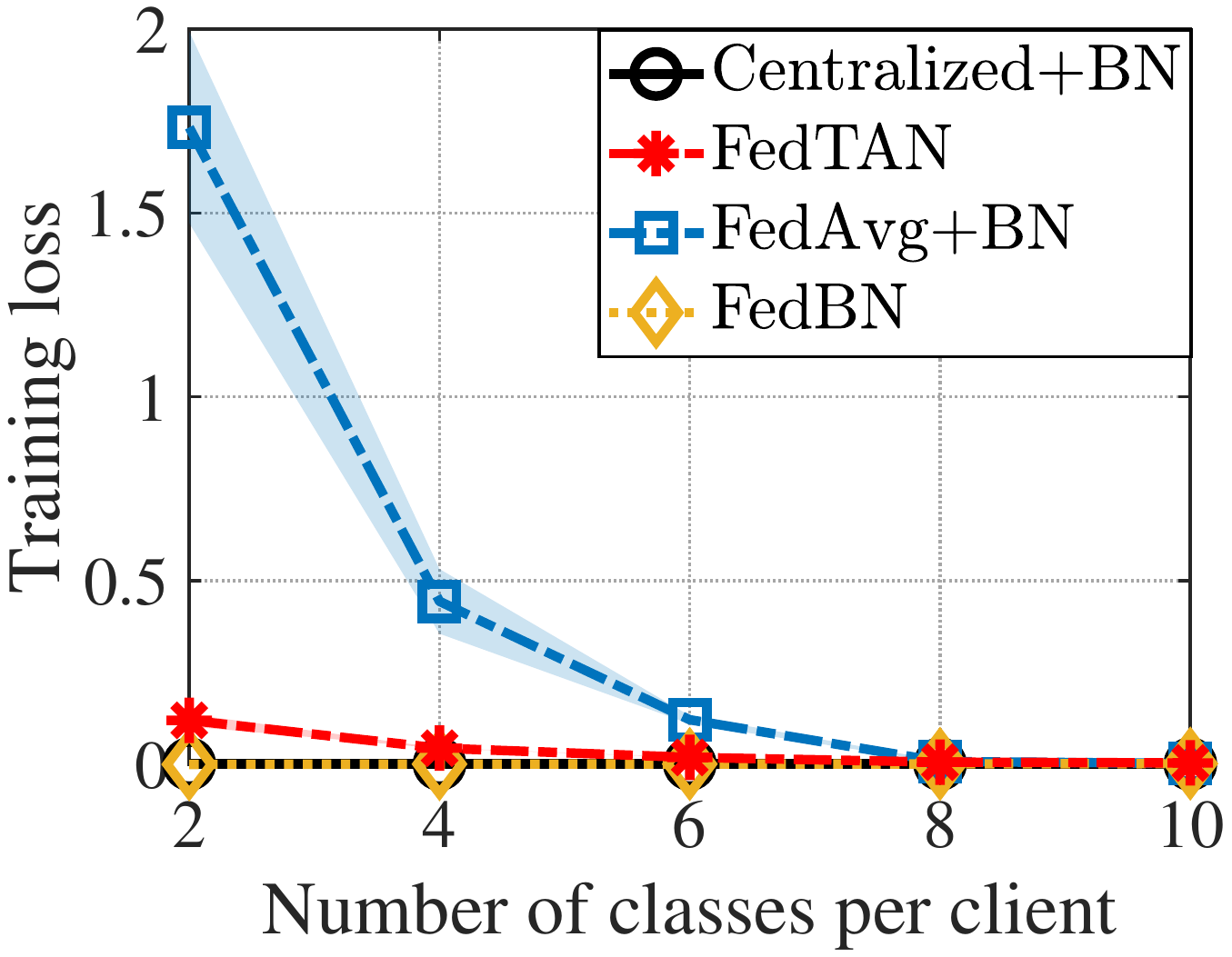}}
\subfigure[Testing accuracy.]{
\includegraphics[width= 2 in ]{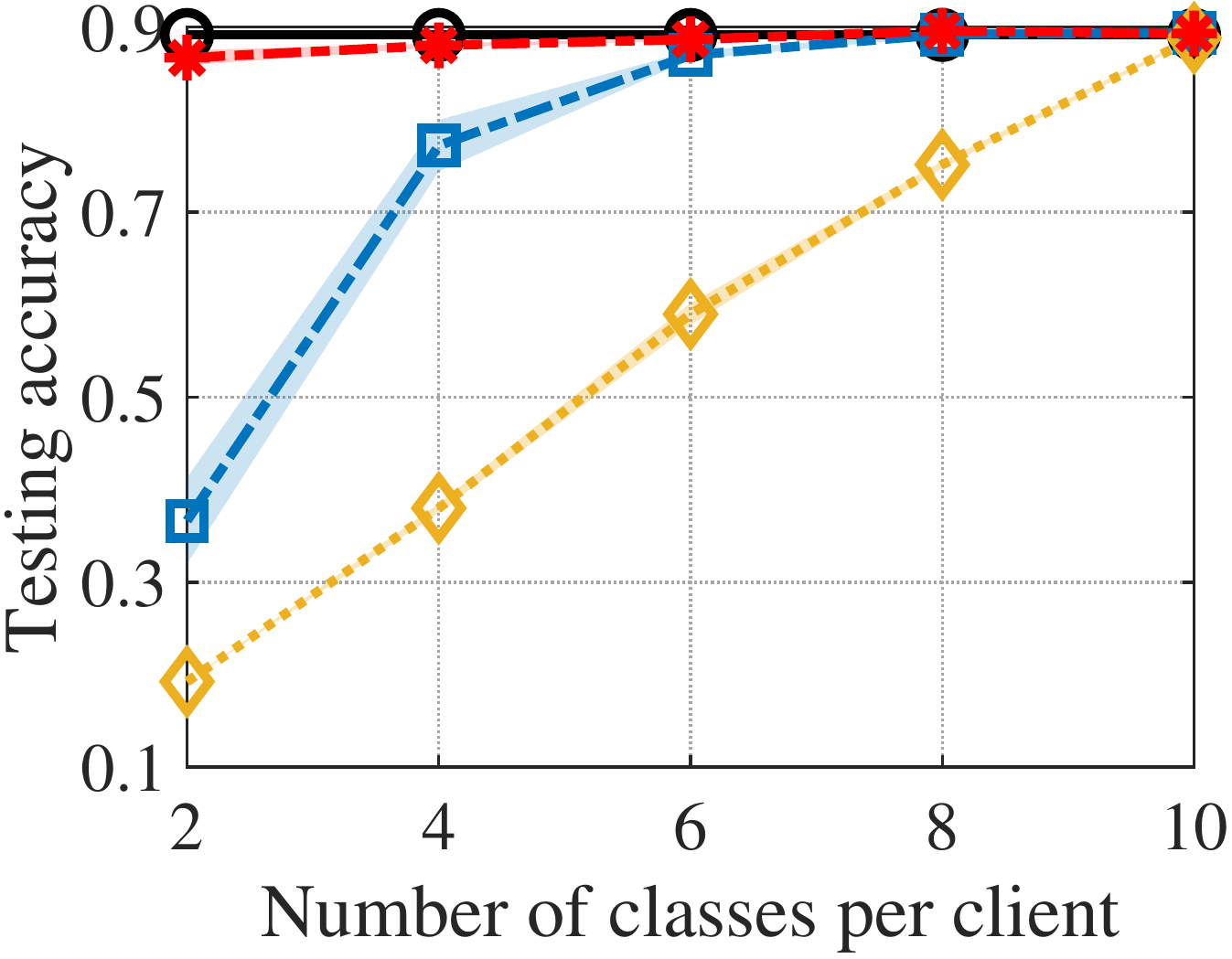}}
\caption{FL Performance on CIFAR-10 dataset with different distribution shifts. The mini-batch SGD without momentum or weight decay is adopted.}
\label{fig:performance shift}
\end{figure}

\subsubsection{Influence of different distribution shifts}

As mentioned in \cite{qu2022rethinking}, a higher number of classes per client reduces distribution shifts among clients.
Consequently, we adopt the data partition method from \cite{qu2022rethinking} and compare the FL performance under different numbers of classes per client, as illustrated in Fig. \ref{fig:performance shift}.
The case with two classes per client corresponds to the considered (extremely) non-i.i.d. data scenario in Section \ref{sec:Parameter Setting}, while ten classes per client represent the i.i.d. data case.
Detailed class distributions for each case are provided in Section \ref{appendix:Class distributions} of the Supplementary Material.
Notably, \texttt{FedTAN} closely matches centralized learning in all cases.
However, as the number of classes per client decreases, both \texttt{FedAvg+BN} and \texttt{FedBN} experience significant performance drops due to gradient deviation and poor generalization ability, respectively.

\begin{figure}[t]
\centering
\subfigure[i.i.d.]{
\includegraphics[width= 2 in ]{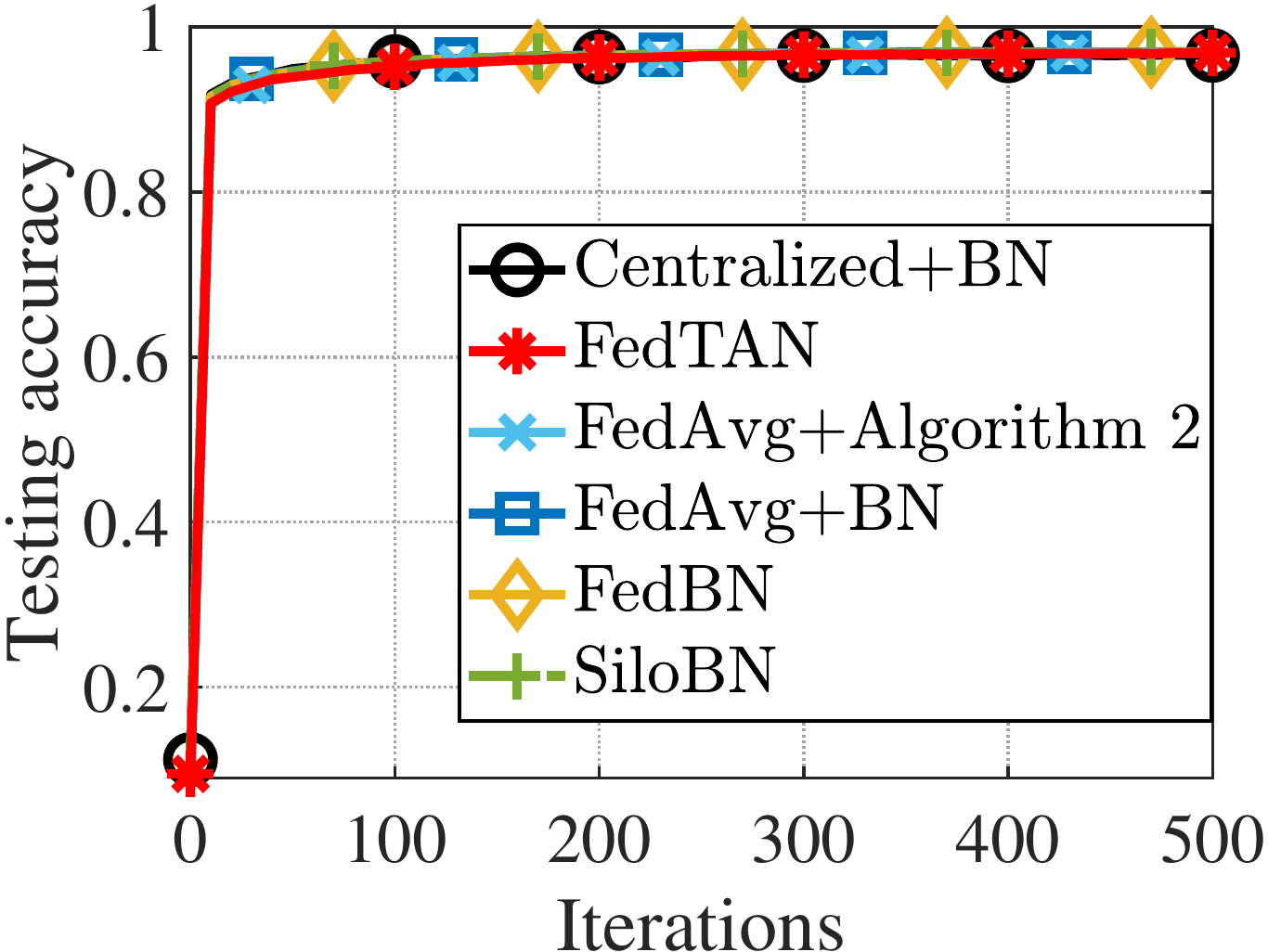}}
\subfigure[Non-i.i.d.]{
\includegraphics[width= 2 in ]{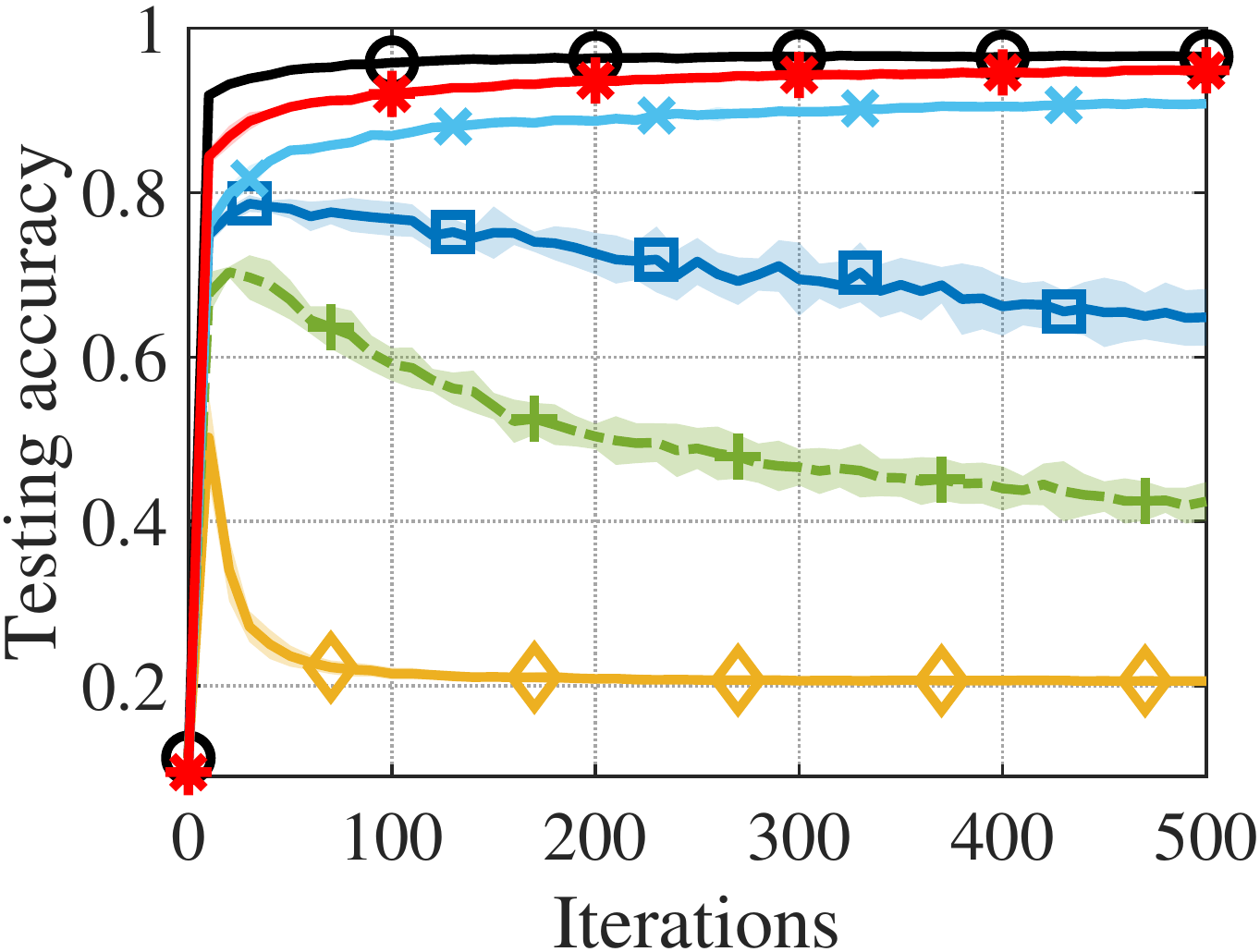}}
\caption{Testing accuracy of different FL schemes on MNIST dataset.}
\label{fig:performance MNIST}
\end{figure}

\begin{figure}[t]
\centering
\subfigure[i.i.d.]{
\includegraphics[width= 2 in ]{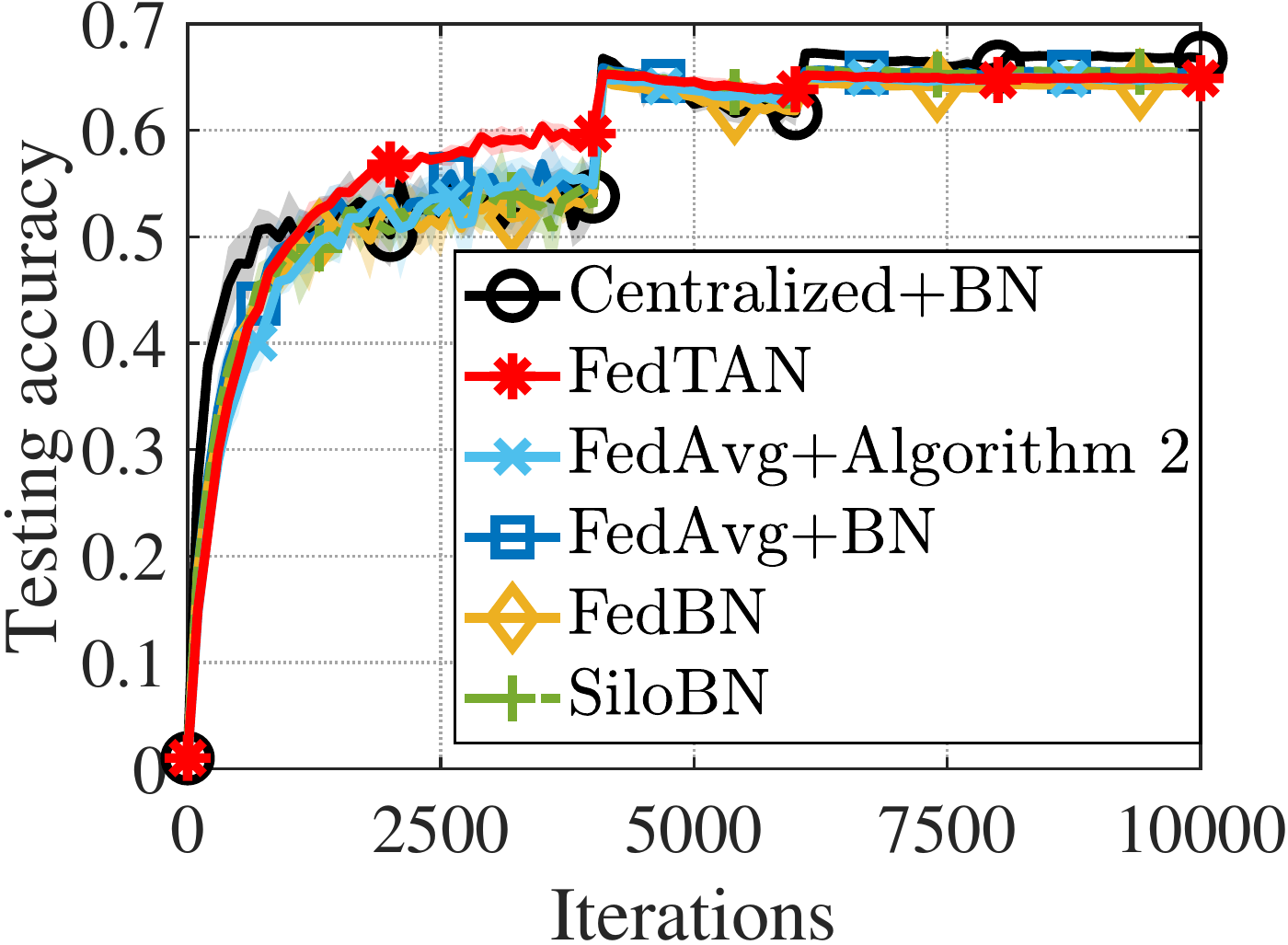}}
\subfigure[Non-i.i.d.]{
\includegraphics[width= 2 in ]{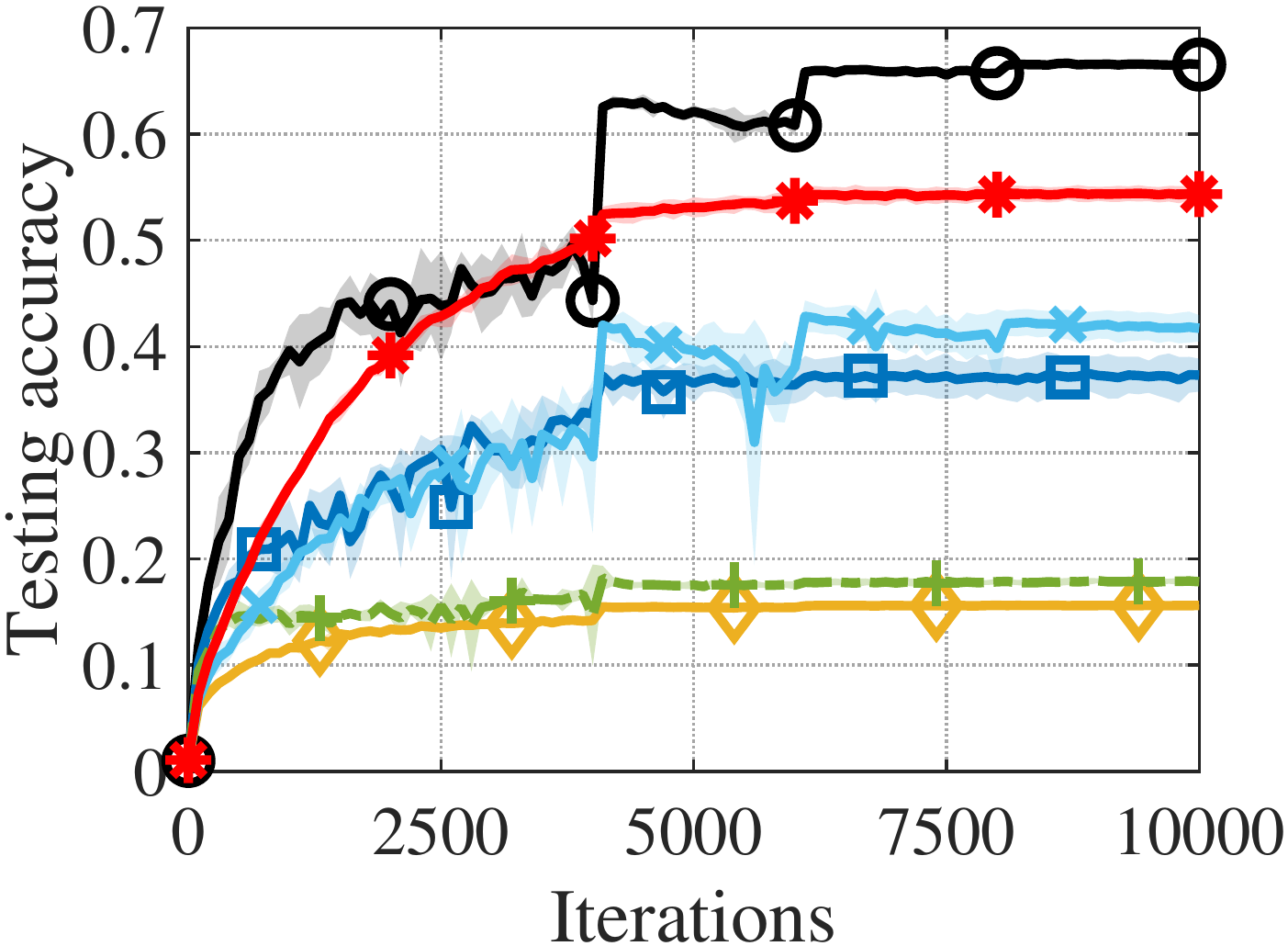}}
\caption{Performance of different FL schemes on CIFAR-100 dataset.}
\label{fig:performance CIFAR-100}
\end{figure}

\begin{figure}[t]
\begin{minipage}[h]{1\linewidth}
\centering
\includegraphics[width= 4.4 in ]{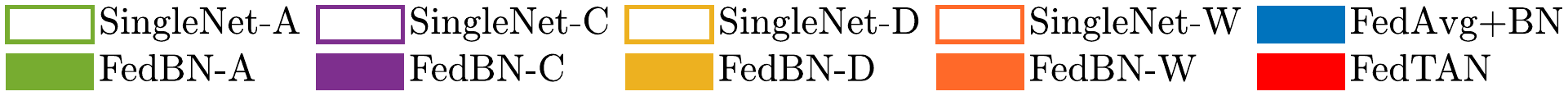}
\end{minipage}
\begin{minipage}[h]{1\linewidth}
\centering
\subfigure[Amazon.]{
\includegraphics[width= 1.4 in ]{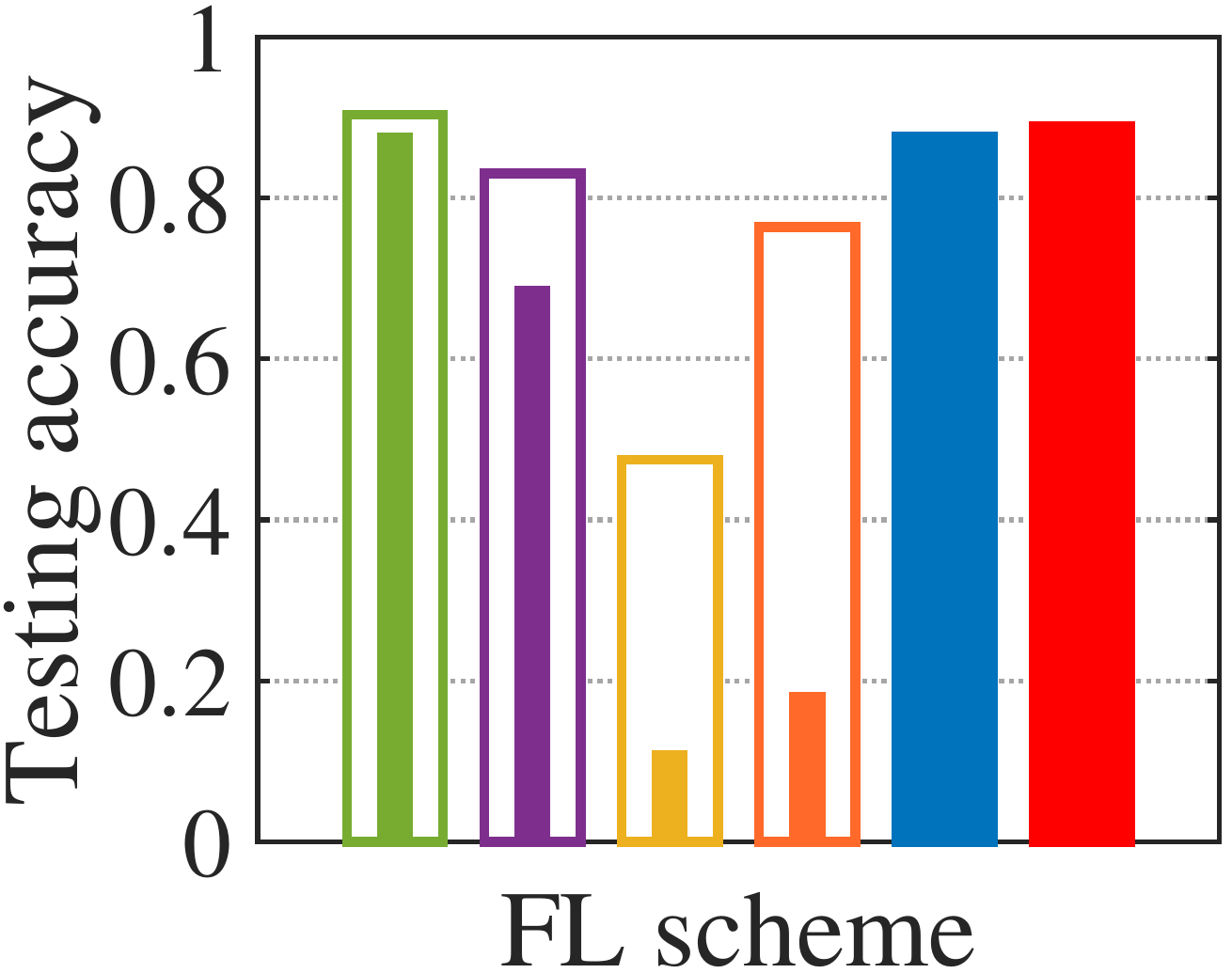}}
\subfigure[Caltech.]{
\includegraphics[width= 1.4 in ]{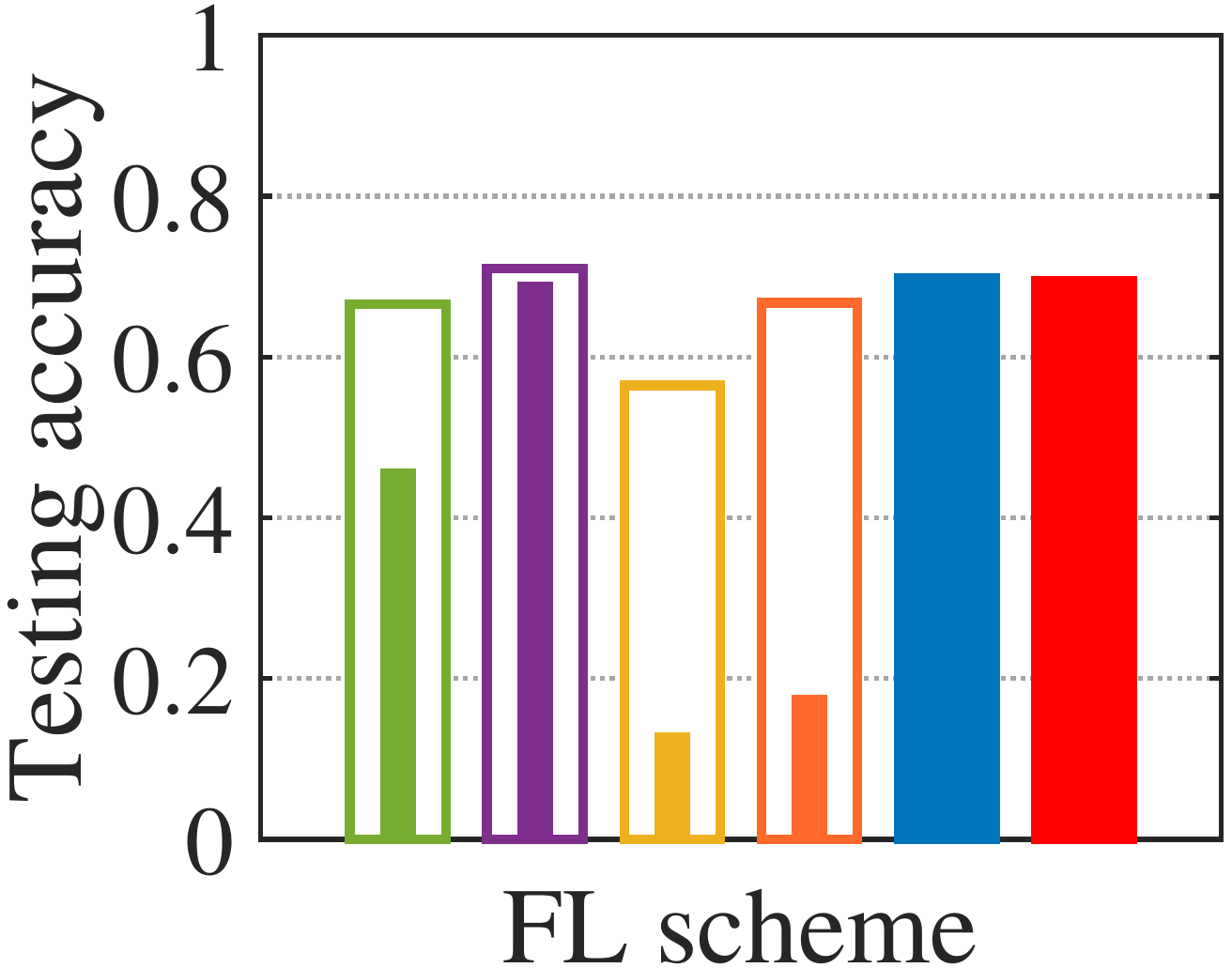}}
\subfigure[DSLR.]{
\includegraphics[width= 1.4 in ]{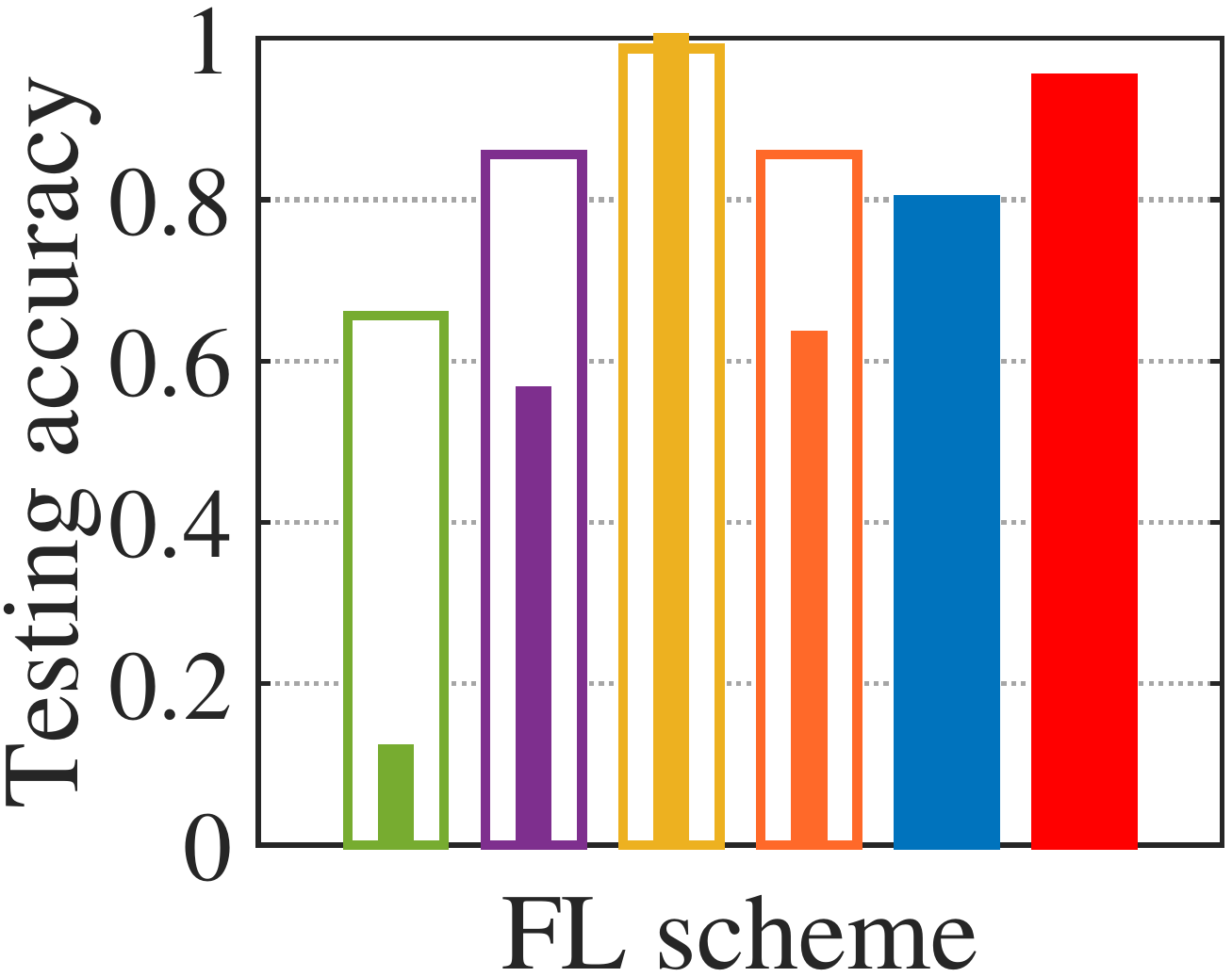}} \\
\subfigure[Webcam.]{
\includegraphics[width= 1.4 in ]{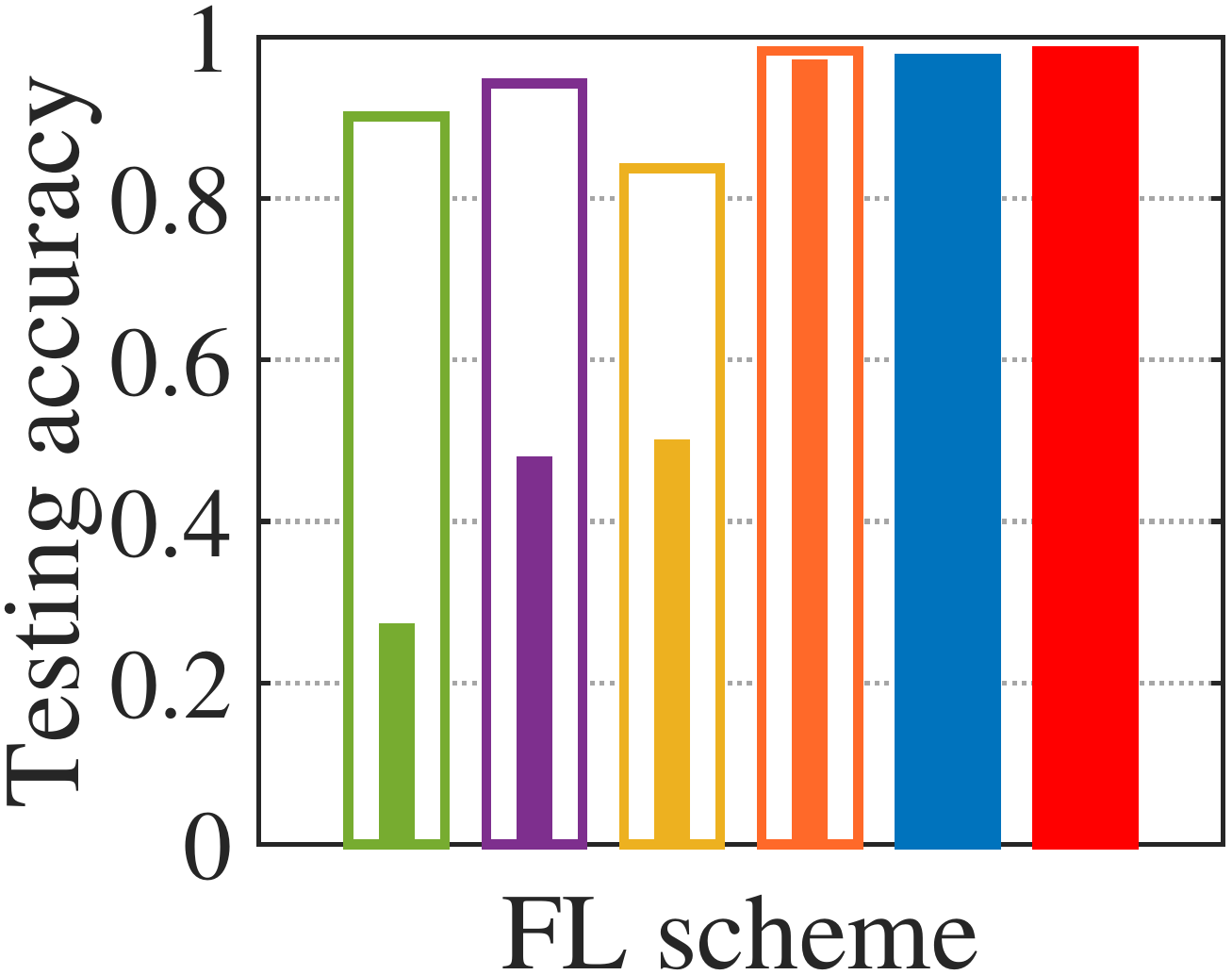}}
\subfigure[Global.]{
\includegraphics[width= 1.4 in ]{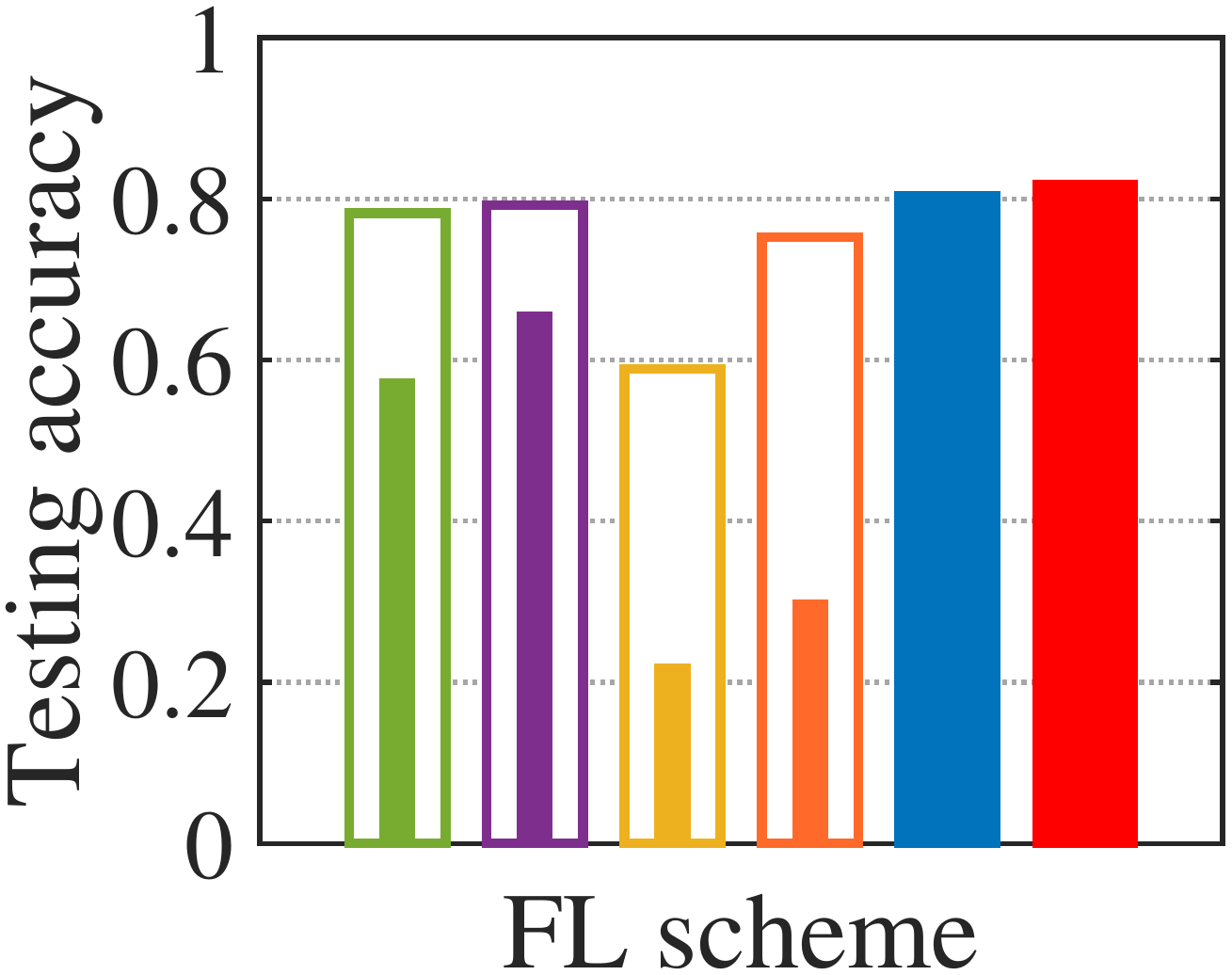}}
\caption{Performance of different FL schemes on Office-Caltech10 dataset.}
\label{fig:performance Office Caltech}
\end{minipage}
\end{figure}

\subsection{Adaptability under Different-Scale Datasets}
To assess the adaptability of proposed \texttt{FedTAN} on datasets of varying scales, we further compare its performance on two additional datasets: MNIST with ten classes of images (Fig. \ref{fig:performance MNIST}), and CIFAR-100 with one hundred classes of images (Fig. \ref{fig:performance CIFAR-100}).
As shown in Fig. \ref{fig:performance MNIST} and \ref{fig:performance CIFAR-100}, similar to the previous CIFAR-10 experiment depicted in Fig. \ref{fig:performance iid noniid}, \texttt{FedTAN} exhibits even closer performance to centralized learning in both the i.i.d. and non-i.i.d. data cases.
Conversely, other FL baselines experience a significant performance drop in the non-i.i.d. data case.

\subsection{Generalization Ability under Feature Shifts}

In the non-i.i.d. case of previous experiments, clients have different label distributions for their local training samples.
In this subsection, we use the Office-Caltech10 dataset to demonstrate how the proposed \texttt{FedTAN} effectively addresses another non-i.i.d. issue called feature shift \cite{li2021fedbn}.
Feature shift refers to a scenario where clients have the same label distributions but different feature distributions.

We denote \texttt{FedBN-A}, \texttt{FedBN-C}, \texttt{FedBN-D}, and \texttt{FedBN-W} as \texttt{FedBN} schemes updated by clients with local datasets from Amazon, Caltech, DSLR, and Webcam, respectively.
The same notation applies to \texttt{SingleNet} variants.
Fig. \ref{fig:performance Office Caltech} compares the performance of various FL schemes on testing datasets from different data sources and the whole global testing dataset.
It can be observed that \texttt{FedBN} outperforms \texttt{SingleNet} when the training and testing datasets come from different sources due to its improved generalization ability.
However, \texttt{FedBN} would still perform worse when datasets have different sources, as seen in \texttt{FedBN-D} in Fig. \ref{fig:performance Office Caltech}(a) and \texttt{FedBN-A} in Fig. \ref{fig:performance Office Caltech}(c), because it updates all BN parameters locally.
In contrast, \texttt{FedTAN}, with aggregated model parameters in the server, exhibits superior generalization ability and achieves satisfactory performance under each testing dataset.
It proves more robust than \texttt{FedAvg+BN}, which suffers from gradient deviation, especially evident in Fig. \ref{fig:performance Office Caltech}(c).
Additionally, we conduct experiments on training AlexNet using the DomainNet dataset \cite{peng2019moment}, obtaining similar observations.
See Section \ref{appendix:DomainNet} in the Supplementary Material for details.

\subsection{Comparison with Group Normalization}

\subsubsection{Learning performance}
\begin{figure}[t]
\begin{minipage}[h]{1\linewidth}
\centering
\includegraphics[width= 2.5 in ]{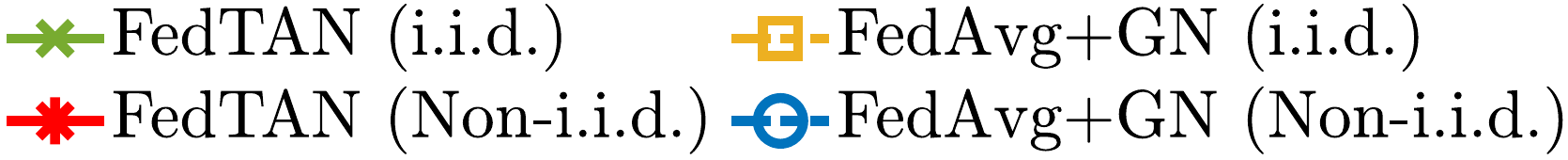}
\end{minipage}
\begin{minipage}[h]{1\linewidth}
\centering
\subfigure[Without momentum.]{
\includegraphics[width= 2 in ]{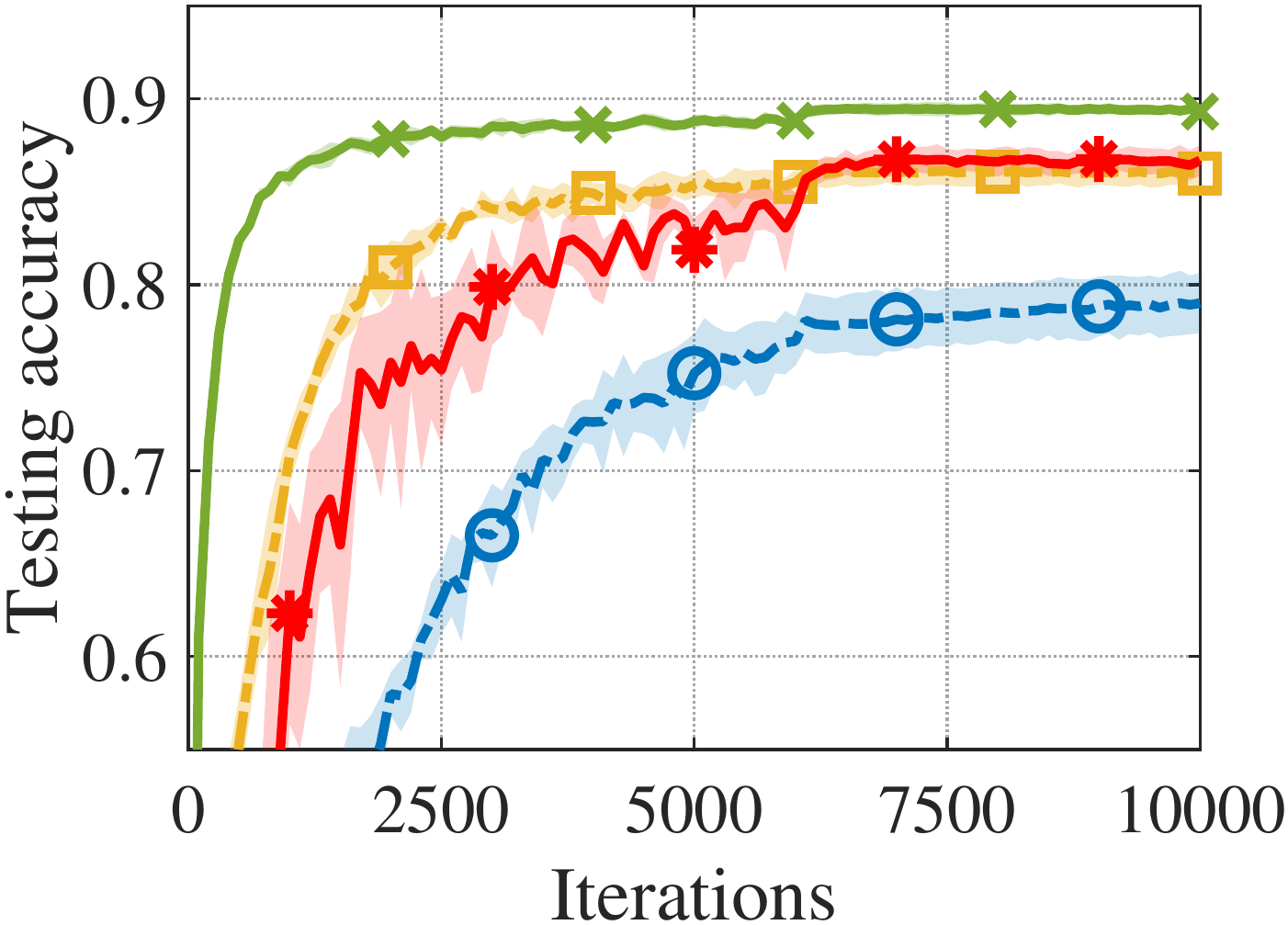}}
\subfigure[With momentum.]{
\includegraphics[width= 2 in ]{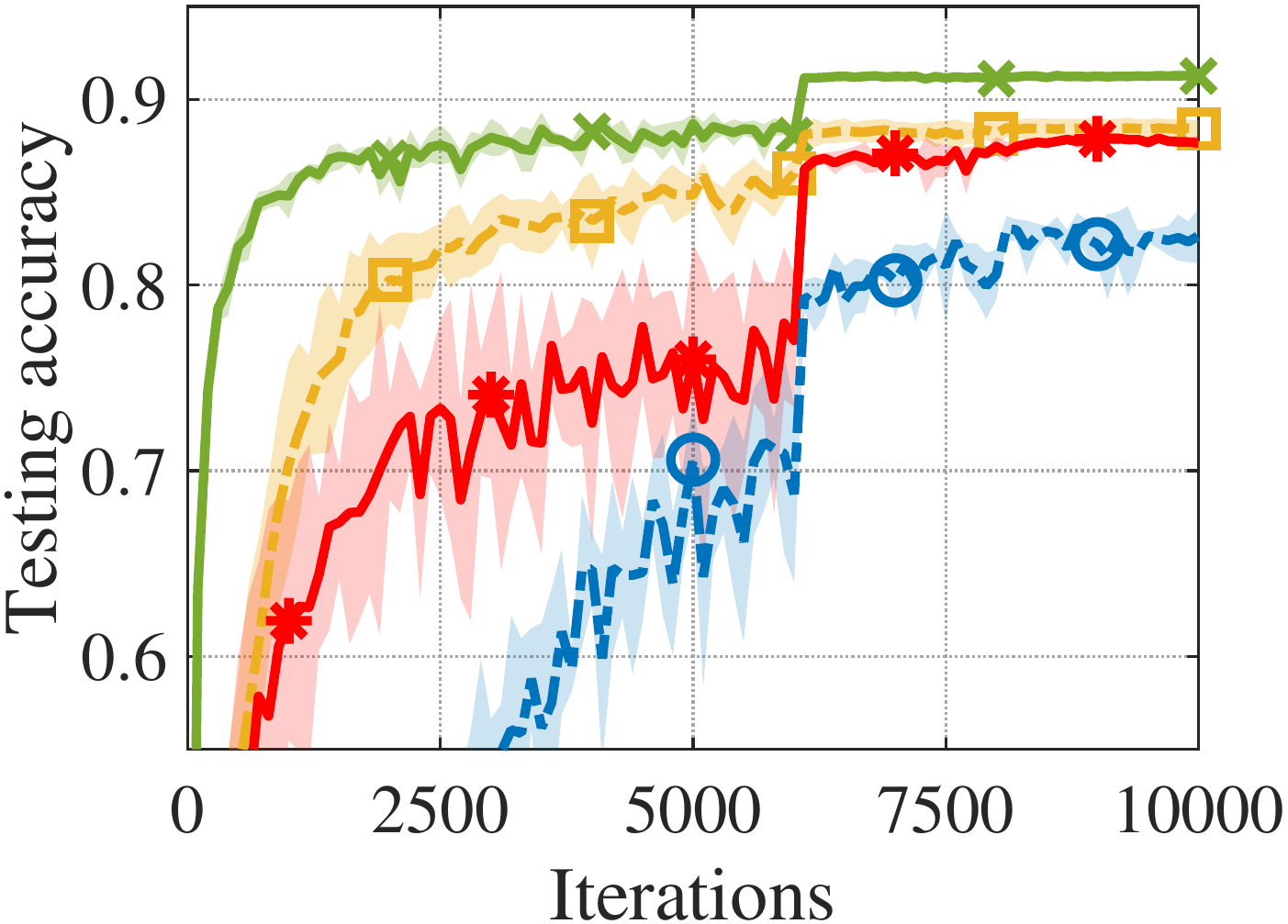}}
\caption{Performance comparison between \texttt{FedTAN} and \texttt{FedAvg+GN} on CIFAR-10 dataset.}
\label{fig:performance GN}
\end{minipage}
\end{figure}

In comparison to BN, GN \cite{wu2018group} is less sensitive to data distribution and also widely used.
Despite this, GN is still unable to match the performance of BN in many recognition tasks.
For instance, replacing BN with GN during ResNet training on the CIFAR dataset leads to performance degradation \cite{zhang2020passport}.
Based on CIFAR-10 dataset, Fig. \ref{fig:performance GN} compares the performance of the ResNet-20 with BN trained by \texttt{FedTAN} and the one with GN trained by \texttt{FedAvg}, in terms of iterations.
Additionally, Table \ref{table:testing accuracy FedTAN GN} compares the testing accuracy of the global models obtained through these two schemes after the entire training process.
It can be seen from Fig. \ref{fig:performance GN} and Table \ref{table:testing accuracy FedTAN GN} that, for both i.i.d. and non-i.i.d data cases, \texttt{FedTAN} maintains the superiority of BN and achieves higher testing accuracy with faster convergence rates.
Furthermore, employing mini-batch SGD with momentum enhances the learning performance of both \texttt{FedTAN} and \texttt{FedAvg+GN} compared to the case without momentum.

\begin{table}[t]
\small
\centering
\caption{Testing accuracy of \texttt{FedTAN} and \texttt{FedAvg+GN} on CIFAR-10 dataset.}\label{table:testing accuracy FedTAN GN}
\begin{tabular}{c|cc|cc}
\hline
\multirow{2}{*}{\textbf{FL scheme}} & \multicolumn{2}{c|}{\textbf{Without momentum (\%)}} & \multicolumn{2}{c}{\textbf{With momentum (\%)}} \\
& \textbf{i.i.d.} & \textbf{Non-i.i.d.} & \textbf{i.i.d.} & \textbf{Non-i.i.d.} \\ \hline
\texttt{FedTAN} & \textbf{89.32}$\pm$0.28 & \textbf{86.69}$\pm$0.82 & \textbf{91.26}$\pm$0.21 & \textbf{87.66}$\pm$0.43 \\
\texttt{FedAvg+GN}& 85.99$\pm$0.71 & 79.01$\pm$1.63 & 88.41$\pm$0.46 & 82.66$\pm$1.48 \\ \hline
\end{tabular}
\end{table}

\subsubsection{Communication overhead}\label{sec:communication overhead}

\begin{figure}[t]
\begin{minipage}[h]{1\linewidth}
\centering
\includegraphics[width= 4.2 in ]{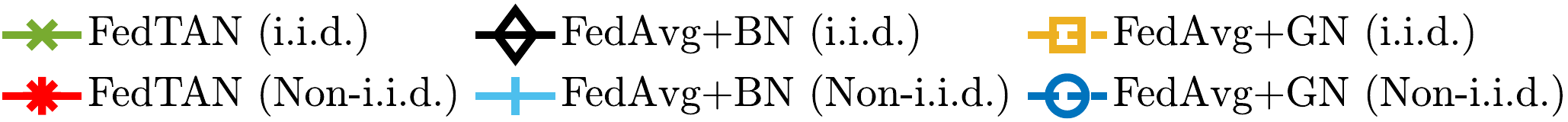}
\end{minipage}
\begin{minipage}[h]{1\linewidth}
\centering
\subfigure[Without momentum.]{
\includegraphics[width= 2 in ]{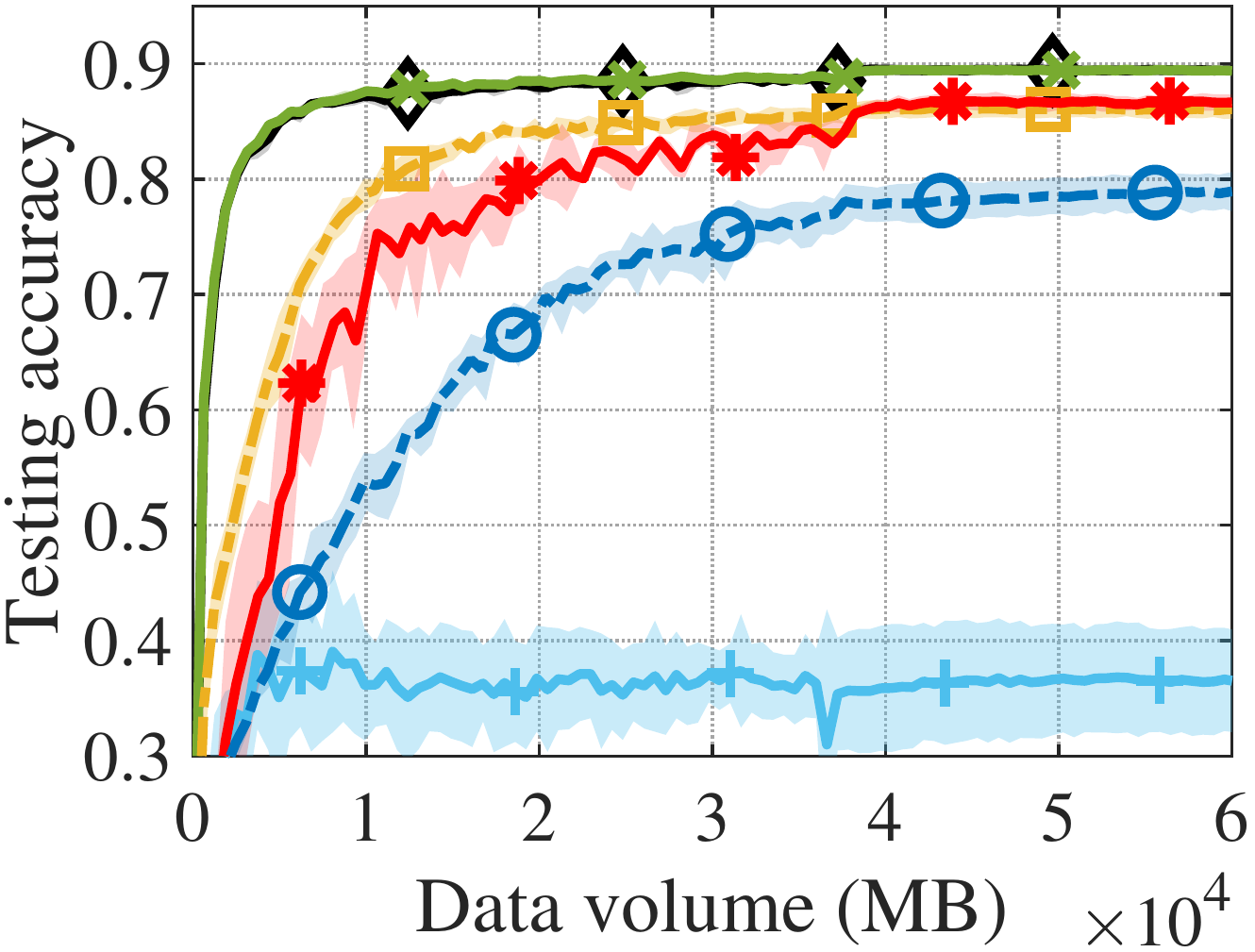}}
\subfigure[With momentum.]{
\includegraphics[width= 2 in ]{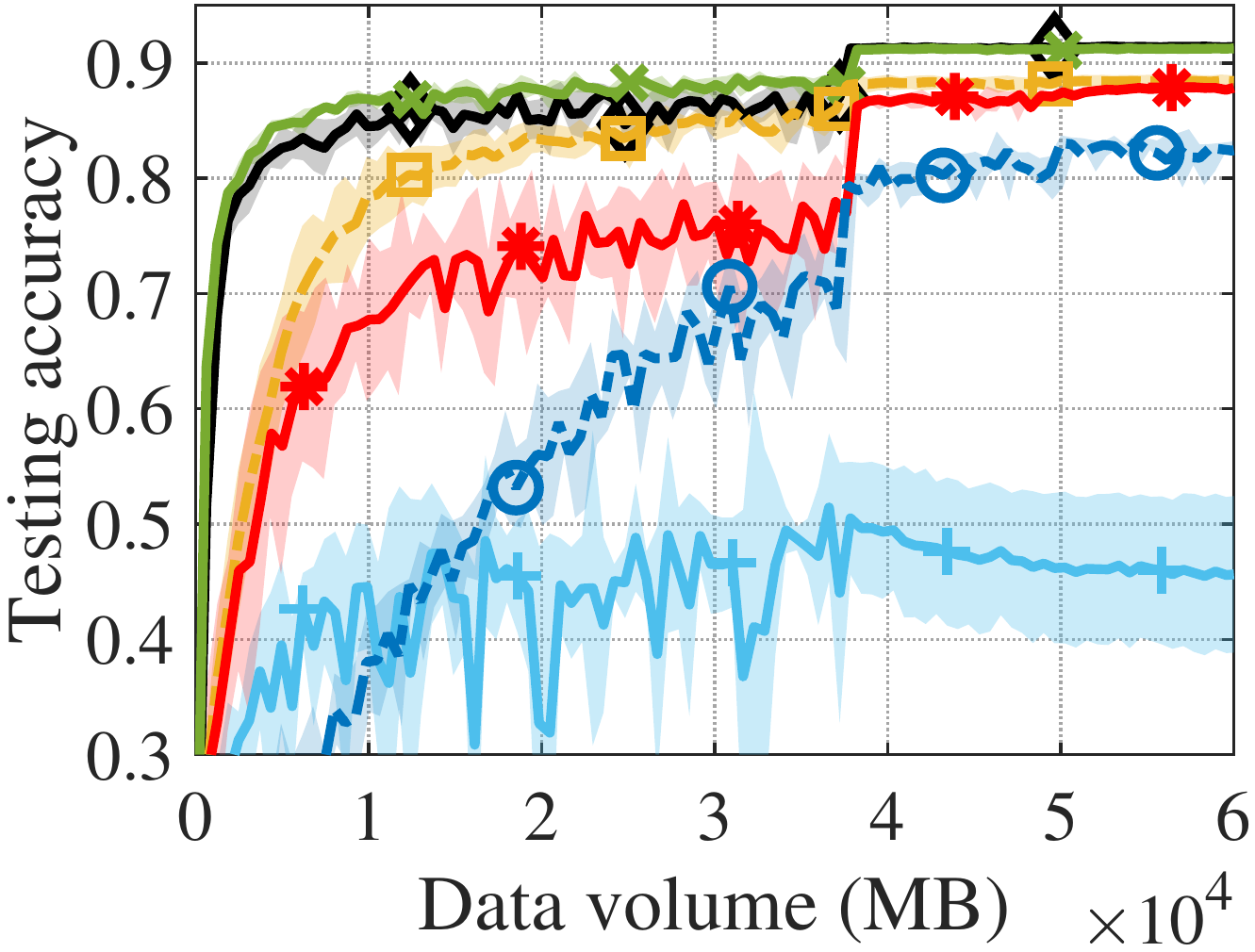}}
\caption{FL Performance w.r.t. exchanged bits on CIFAR-10 dataset.}
\label{fig:performance data volume}
\end{minipage}
\end{figure}

In \texttt{FedTAN}, as depicted in Algorithms \ref{algorithm:FedTAN procedure}, \ref{algorithm:FedTAN modification forward} and \ref{algorithm:FedTAN modification backward}, the modified first local updating step brings extra transmission of statistical parameters and their gradients between the server and the clients.
Table \ref{table:communication overhead} compares the number of bits exchanged between the server and five clients per iteration required by different FL schemes when training ResNet-20 with the CIFAR-10 dataset, where the detailed calculation process of each exchanged data volume is provided in Section \ref{appendix:transmitted data volume} of the Supplementary Material.
Table \ref{table:communication overhead} shows that, compared to \texttt{FedAvg+BN} and \texttt{FedAvg+GN}, the transmitted data volume between the server and clients in \texttt{FedTAN} only increases by 1.02\% and 1.52\%, respectively.
This slight increase is due to statistical parameters comprising a small portion of all parameters, especially in large-scale DNNs.
Next, Fig. \ref{fig:performance data volume} compares the testing accuracy of various FL schemes concerning the number of bits exchanged between the server and all clients.
The figure shows that with negligible additional exchanged bits, \texttt{FedTAN} is still capable of maintaining the faster convergence rate w.r.t. the transmission data volume.
Specifically, given an expected level of testing accuracy, \texttt{FedAvg+GN} requires significantly more bits to be exchanged between the server and clients due to its slower convergence rate compared to \texttt{FedTAN}.
Moreover, while \texttt{FedTAN} achieves higher learning accuracy, the extra communication rounds caused by the layer-wise aggregations in Algorithms \ref{algorithm:FedTAN modification forward} and \ref{algorithm:FedTAN modification backward} will increase the total communication rounds required for \texttt{FedTAN}.
Thus, \texttt{FedAvg+GN} is recommended if the number of communication rounds is critical.

\begin{table}[t]
\small
\centering
\caption{Communication overhead per iteration of FL schemes on CIFAR-10 dataset.}\label{table:communication overhead}
\begin{tabular}{c|c|cc}
\hline
\multirow{2}{*}{\textbf{FL scheme}} & \textbf{Exchanged bits} & \multicolumn{2}{c}{\textbf{Percentage higher than}}  \\
& \textbf{(MB)} & \textbf{\texttt{FedAvg+BN}} & \textbf{\texttt{FedAvg+GN}} \\ \hline
\texttt{FedTAN} & 6.2679 & 1.02\% & 1.52\% \\
\texttt{FedAvg+BN} & 6.2049 & 0 & 0.51\% \\
\texttt{FedAvg+GN} & 6.1734 & -0.51\% & 0 \\ \hline
\end{tabular}
\end{table}

\subsection{Performance of \texttt{FedTAN-II}}

\begin{figure}[t]
\centering
\subfigure[Without momentum.]{
\includegraphics[width= 2 in ]{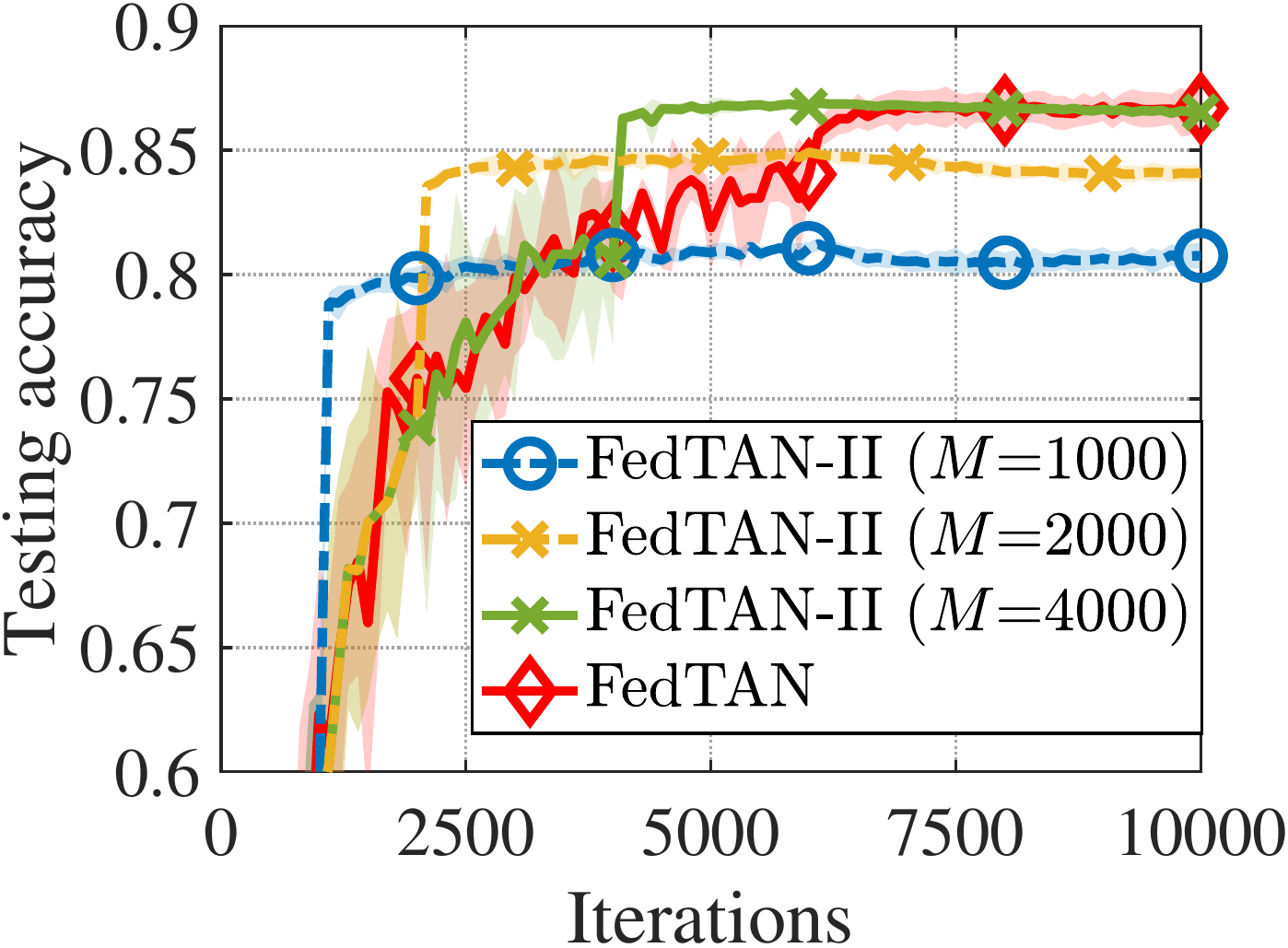}}
\subfigure[Without momentum.]{
\includegraphics[width= 1.95 in ]{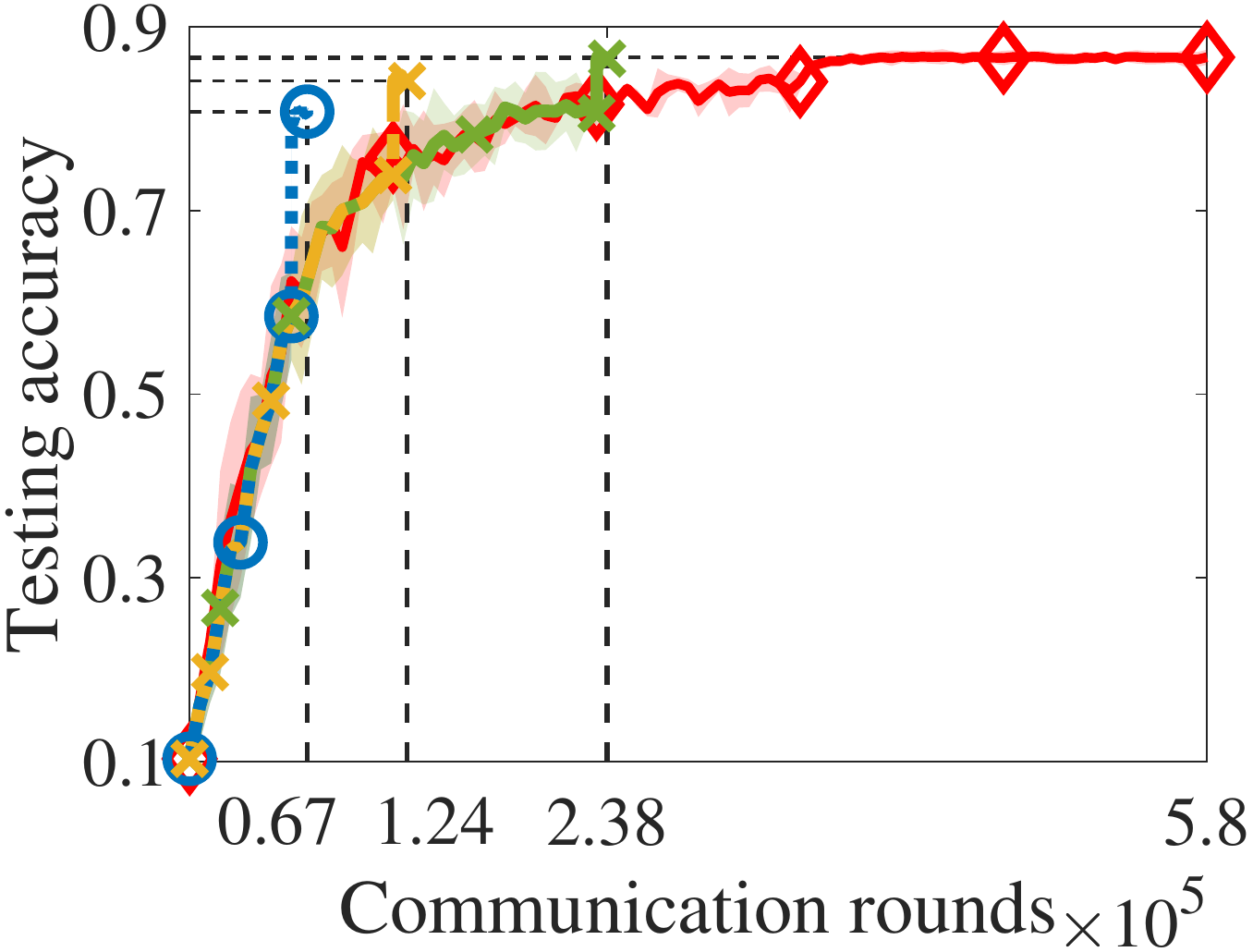}} \\
\subfigure[With momentum.]{
\includegraphics[width= 2 in ]{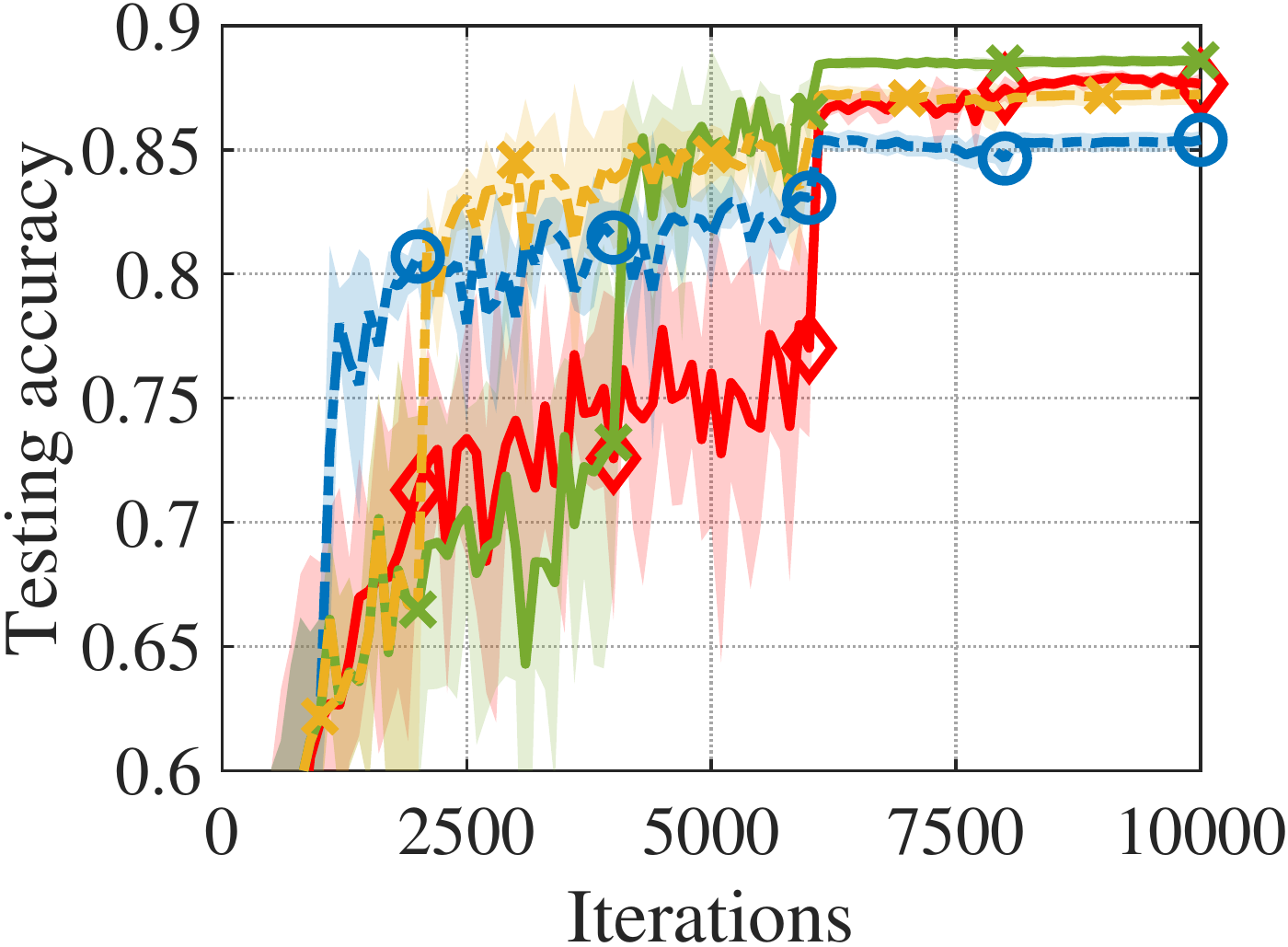}}
\subfigure[With momentum.]{
\includegraphics[width= 1.95 in ]{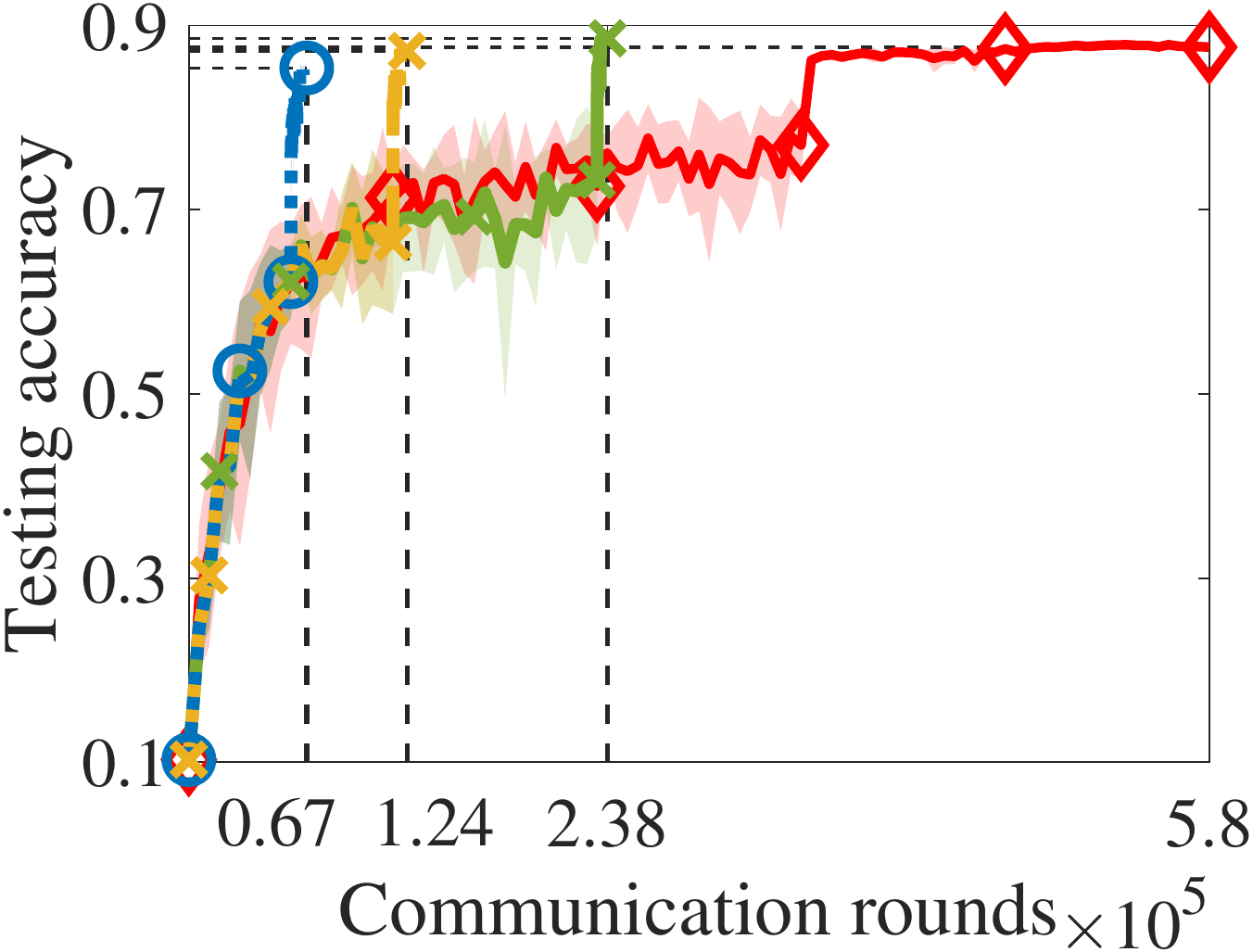}}
\caption{Testing accuracy of \texttt{FedTAN-II} on CIFAR-10 dataset.}
\label{fig:performance FedTAN-II}
\end{figure}

Finally, we demonstrate the effectiveness of \texttt{FedTAN-II}.
Fig. \ref{fig:performance FedTAN-II}(a) and (b) compare \texttt{FedTAN-II} with \texttt{FedTAN} by training ResNet-20 with BN on the CIFAR-10 dataset in terms of iterations, while Fig. \ref{fig:performance FedTAN-II}(c) and (d) compare them in terms of communication rounds.
In the case of \texttt{FedTAN-II}, we initially set the learning rate $\gamma$ to 0.5 during the first $M$ iterations, similar to \texttt{FedTAN}, as mentioned in Section \ref{sec:Parameter Setting}.
After $M$ iterations, we adjust the learning rate $\gamma$ since we only update the gradient parameters while keeping the statistical parameters in BN layers fixed.
Consequently, $\gamma$ is decreased to 0.01 and 0.001 after $M$ and 6000 iterations, respectively.
If momentum ($0.9$) and weight decay ($10^{-4}$) are used in mini-batch SGD, $\gamma$ is further reduced to 0.0001 after 8000 iterations.

From Fig. \ref{fig:performance FedTAN-II}, we observe that reducing $M$ leads to a slight accuracy loss if $M$ is smaller than a certain value (e.g., $M \leq 2000$ in Fig. \ref{fig:performance FedTAN-II}(a) and $M \leq 1000$ in Fig. \ref{fig:performance FedTAN-II}(c)), primarily due to less accurate statistical parameters.
However, a smaller $M$ significantly reduces the required communication rounds, as shown in Fig. \ref{fig:performance FedTAN-II}(b) and (d).
Furthermore, a comparison between Fig. \ref{fig:performance FedTAN-II}(a) and (c) reveals that adopting mini-batch SGD with momentum and weight decay greatly improves the testing accuracy of \texttt{FedTAN-II} under different $M$.

Next, Table \ref{table:performance FedTAN-II} compares the total exchanged bits, the percentages of extra communication rounds and exchanged bits caused by layer-aggregations in Algorithms \ref{algorithm:FedTAN modification forward} and \ref{algorithm:FedTAN modification backward} during the entire training process of \texttt{FedTAN-II} for varying values of $M$.
The table shows that the difference in total exchanged bits between \texttt{FedTAN} and \texttt{FedTAN-II} under different $M$ is negligible, as the statistical parameters in BN layers occupy only a small portion of the total model size.
Moreover, we find that although the extra communication rounds caused by layer-aggregations in Algorithms \ref{algorithm:FedTAN modification forward} and \ref{algorithm:FedTAN modification backward} account for a large percentage of the total exchanged rounds during the entire training process, the corresponding extra exchanged bits are negligible.
Thus, the layer-aggregations in Algorithms \ref{algorithm:FedTAN modification forward} and \ref{algorithm:FedTAN modification backward} constitute a high-frequent low-volume communication mode, which effectively mitigates divergence among local models, as highlighted in \cite{sattler2019robust}.

Additionally, we carry out experiments involving the training of a 3-layer DNN with BN on the MNIST dataset, leading to consistent observations.
See Section \ref{appendix:FedTAN-II MNIST} in the Supplementary Material for details.

\begin{table}[t]
\centering
\caption{Communication overhead of \texttt{FedTAN-II} on CIFAR-10 dataset.}\label{table:performance FedTAN-II}
\begin{tabular}{c|c|cc}
\hline
\multirow{2}{*}{\textbf{FL scheme}} & \textbf{Exchanged} & \multicolumn{2}{c}{\textbf{Percentage of extra}}  \\
& \textbf{bits (GB)} & \textbf{Rounds} & \textbf{Bits} \\ \hline
\texttt{FedTAN-II} ($M$$=$1000) & 60.6563 & 85.07\% & 0.1014\% \\
\texttt{FedTAN-II} ($M$$=$2000) & 60.7178 & 91.94\% & 0.2027\% \\
\texttt{FedTAN-II} ($M$$=$4000) & 60.8408 & 95.80\% & 0.4045\% \\
\texttt{FedTAN}         & 61.2100 & 98.28\% & 1.0051\% \\ \hline
\end{tabular}
\end{table}

\section{Conclusion}\label{sec:conclusion}

In this paper, we have investigated why BN damages the performance of FL in the non-i.i.d. data case.
Our novel convergence analysis has revealed that BN significantly hinders the FL convergence in the presence of non-i.i.d. data.
Specifically, we have shown that non-i.i.d. data would result in two types of mismatches: one between local and global statistical parameters and the other between the associated gradients w.r.t. the statistical parameters.
These mismatches lead to gradient deviation in local models, which subsequently biases the FL convergence (Theorem \ref{theorem:1}).
Fortunately, the gradient deviation can be eliminated if the local statistical parameters and related gradients, obtained by clients in the first local updating step, are the same as the global ones obtained by the centralized learning scheme.
Inspired by this insight, we have proposed \texttt{FedTAN}, which leverages layer-wise statistical parameter aggregation to address the aforementioned mismatches and mitigate gradient deviation.
The experimental results have showcased that, compared to existing schemes, \texttt{FedTAN} preserves the training advantage of BN while exhibiting superior robustness, particularly in non-i.i.d. data settings.
Moreover, although layer-wise aggregation in \texttt{FedTAN} leads to a negligible increase in transmitted data volume, it does introduce a noticeable rise in communication rounds.
To tackle this challenge, we have proposed \texttt{FedTAN-II}, which achieves satisfactory learning performance while significantly reducing the communication rounds.

\section*{Appendix}

\appendix

\section{Proof of Theorem \ref{theorem:1}}\label{sec:thm 1 proof}

\subsection{Proof of Convergence Rate}

Based on Assumption \ref{assumption:L continuous BN}, we have

\vspace{-0.3cm}
\begin{small}
\begin{align}\label{Proof_Thm1_1_rewrite}
F({\bar {\mathbf{w}}}_{r};\mathcal{S}_{\mathcal{D}}^{r+1,1}, \Delta \mathcal{S}_{\mathcal{D}}^{r+1,1})
\leq F({\bar {\mathbf{w}}}_{r-1};\mathcal{S}_{\mathcal{D}}^{r,1}, \Delta \mathcal{S}_{\mathcal{D}}^{r,1})
+ \langle\nabla_{\mathbf{w}} F({\bar {\mathbf{w}}}_{r-1};\mathcal{S}_{\mathcal{D}}^{r,1}, \Delta \mathcal{S}_{\mathcal{D}}^{r,1}), {\bar {\mathbf{w}}}_{r} - {\bar {\mathbf{w}}}_{r-1}\rangle
+  \frac{L}{2} \| {\bar {\mathbf{w}}}_{r} - {\bar {\mathbf{w}}}_{r-1}  \|^2
,
\end{align}
\end{small}
\vspace{-0.3cm}

\noindent
where $\mathcal{S}_{\mathcal{D}}^{r,1}$ and $\Delta \mathcal{S}_{\mathcal{D}}^{r,1}$ are obtained via \eqref{eq:mu sigma ellj} with ${\bar {\mathbf{w}}}_{r-1}$ and the global dataset $\mathcal{D}$.
Then, based on \eqref{Proof_Thm1_1_rewrite}, we use the following three key lemmas which are proved in subsequent subsections.

\begin{lemma}\label{lemma:1}
Under Assumptions \ref{assumption:BN disturbance} and \ref{assumption:L continuous BN}, it holds that

\vspace{-0.3cm}
\begin{small}
\begin{align}\label{eq:lemma 1}
\langle\nabla_{\mathbf{w}} F({\bar {\mathbf{w}}}_{r-1};\mathcal{S}_{\mathcal{D}}^{r,1}, \Delta \mathcal{S}_{\mathcal{D}}^{r,1}), {\bar {\mathbf{w}}}_{r} - {\bar {\mathbf{w}}}_{r-1}\rangle
\leq
&
-\frac{\gamma E}{2} \big\| \nabla_{\mathbf{w}} F({\bar {\mathbf{w}}}_{r-1};\mathcal{S}_{\mathcal{D}}^{r,1}, \Delta \mathcal{S}_{\mathcal{D}}^{r,1}) \big\|^2
\! - \! \frac{\gamma}{2}  \sum_{t = 1}^{E} \left\|  \sum_{i = 1}^N p_i \nabla_{\mathbf{w}} F_{i} (\mathbf{w}^{r,t-1}_{i}; \mathcal{S}_{\mathcal{D}_i}^{r,t}, \Delta \mathcal{S}_{\mathcal{D}_i}^{r,t}) \right\|^2 \notag \\
&
+ \gamma E \sum_{i = 1}^N p_i B_i^2
+ \gamma L^2 \sum_{i = 1}^N  p_i \sum_{t = 1}^{E}  \left\| \mathbf{w}^{r,t-1}_{i} - \mathbf{\bar w}_{r-1}  \right\|^2
,
\end{align}
\end{small}
\vspace{-0.3cm}

\noindent
where $\mathcal{S}_{\mathcal{D}_i}^{r,t}$ and $\Delta \mathcal{S}_{\mathcal{D}_i}^{r,t}$ are obtained via \eqref{eq:mu sigma ellj} with $\mathbf{w}^{r,t-1}_{i}$ and the local dataset $\mathcal{D}_{i}$.
\end{lemma}

\begin{lemma}\label{lemma:2}
With \eqref{eq:local SGD} and \eqref{eq:global model}, we have

\vspace{-0.3cm}
\begin{small}
\begin{equation}\label{bar wr - wr-1}
 {\bar {\mathbf{w}}}_{r} - {\bar {\mathbf{w}}}_{r-1} =  - \gamma \sum_{t = 1}^{E}  \sum_{i = 1}^N p_i \nabla_{ \mathbf{w}} F_{i} (\mathbf{w}^{r,t-1}_{i} ; \mathcal{S}_{\mathcal{D}_i}^{r,t}, \Delta \mathcal{S}_{\mathcal{D}_i}^{r,t})
,
\end{equation}
\end{small}
\vspace{-0.3cm}

\noindent
which results in

\vspace{-0.3cm}
\begin{small}
\begin{align}\label{eq:lemma2}
\left\| {\bar {\mathbf{w}}}_{r} - {\bar {\mathbf{w}}}_{r-1}  \right\|^2
\leq
\gamma^2 E
\sum_{t = 1}^{E}
\bigg\|
\sum_{i = 1}^N p_i \nabla_{\mathbf{w}}F_{i} (\mathbf{w}^{r,t-1}_{i}; \mathcal{S}_{\mathcal{D}_i}^{r,t}, \Delta \mathcal{S}_{\mathcal{D}_i}^{r,t}) \bigg\|^2
.
\end{align}
\end{small}
\end{lemma}

\begin{lemma}\label{lemma:3}
Under Assumptions \ref{assumption:L continuous BN} and \ref{assumption:data var}, the difference between the local model at each round $r$ and the global model at the previous round is bounded by

\vspace{-0.3cm}
\begin{small}
\begin{align}\label{eq:lemma3}
\sum_{t = 1}^{E} \left\| \mathbf{w}^{r,t-1}_{i} - {\bar {\mathbf{w}}}_{r-1} \right\|^2
\leq
\frac{4 \gamma^2 E^3 B_i^2 + 4 \gamma^2 E^3 V_i^2 }{ 1 - 4 \gamma^2 E^2 L^2 }
+ \frac{ 4 \gamma^2 E^3 \big\| \nabla_{\mathbf{w}} F(\mathbf{\bar w}_{r-1}; \mathcal{S}_{\mathcal{D}}^{r,1}, \Delta \mathcal{S}_{\mathcal{D}}^{r,1}) \big\|^2}{ 1 - 4 \gamma^2 E^2 L^2 }
.
\end{align}
\end{small}
\end{lemma}

Then, by substituting \eqref{eq:lemma 1} into the third term in the RHS of \eqref{Proof_Thm1_1_rewrite} and substituting \eqref{eq:lemma2} into the second term, we have

\vspace{-0.3cm}
\begin{small}
\begin{align}\label{eq:E F wr upper bound}
& F({\bar {\mathbf{w}}}_{r}; \mathcal{S}_{\mathcal{D}}^{r+1,1}, \Delta \mathcal{S}_{\mathcal{D}}^{r+1,1}) \notag \\
\leq
& F({\bar {\mathbf{w}}}_{r-1}; \mathcal{S}_{\mathcal{D}}^{r,1}, \Delta \mathcal{S}_{\mathcal{D}}^{r,1})
-\frac{\gamma E}{2} \big\| \nabla_{\mathbf{w}} F({\bar {\mathbf{w}}}_{r-1}; \mathcal{S}_{\mathcal{D}}^{r,1}, \Delta \mathcal{S}_{\mathcal{D}}^{r,1}) \big\|^2
- \frac{\gamma(1  -  \gamma E L)}{2}  \sum_{t = 1}^{E} \left\| \sum_{i = 1}^N p_i \nabla_{\mathbf{w}} F_{i} (\mathbf{w}^{r,t-1}_{i}; \mathcal{S}_{\mathcal{D}_i}^{r,t}, \Delta \mathcal{S}_{\mathcal{D}_i}^{r,t}) \right\|^2
\notag \\
&
+ \gamma E \sum_{i = 1}^N p_i B_i^2
+ \gamma L^2 \sum_{i = 1}^N  p_i \sum_{t = 1}^{E}
\left\| \mathbf{w}^{r,t-1}_{i} - \mathbf{\bar w}_{r-1}  \right\|^2
\notag \\
\overset{(a)}{\leq}
& F({\bar {\mathbf{w}}}_{r-1}; \mathcal{S}_{\mathcal{D}}^{r,1}, \Delta \mathcal{S}_{\mathcal{D}}^{r,1})
-\frac{\gamma E}{2} \big\| \nabla_{\mathbf{w}} F({\bar {\mathbf{w}}}_{r-1}; \mathcal{S}_{\mathcal{D}}^{r,1}, \Delta \mathcal{S}_{\mathcal{D}}^{r,1}) \big\|^2
+ \gamma E \sum_{i = 1}^N p_i B_i^2
+  \gamma L^2 \sum_{i = 1}^N  p_i \sum_{t = 1}^{E} \left\| \mathbf{w}^{r,t-1}_{i}
 -  \mathbf{\bar w}_{r-1}  \right\|^2
,
\end{align}
\end{small}
\vspace{-0.3cm}

\noindent
where (a) is obtained by setting $0 < \gamma \leq \frac{1}{EL}$.

Next, by substituting \eqref{eq:lemma3} into the last term of \eqref{eq:E F wr upper bound}, we have

\vspace{-0.3cm}
\begin{small}
\begin{align}
F({\bar {\mathbf{w}}}_{r}; \mathcal{S}_{\mathcal{D}}^{r+1,1}, \Delta \mathcal{S}_{\mathcal{D}}^{r+1,1})
\leq &
F({\bar {\mathbf{w}}}_{r-1}; \mathcal{S}_{\mathcal{D}}^{r,1}, \Delta \mathcal{S}_{\mathcal{D}}^{r,1})
- \left( \frac{\gamma E}{2} - \frac{4 \gamma^3 E^3 L^2}{ 1 - 4 \gamma^2 E^2 L^2 } \right) \big\| \nabla_{\mathbf{w}} F({\bar {\mathbf{w}}}_{r-1}; \mathcal{S}_{\mathcal{D}}^{r,1}, \Delta \mathcal{S}_{\mathcal{D}}^{r,1}) \big\|^2 \notag \\
& + \frac{4 \gamma^3 E^3 L^2}{1 - 4 \gamma^2 E^2 L^2} \sum_{i = 1}^N  p_i V_i^2
+ \frac{4 \gamma^3 E^3 L^2}{1 - 4 \gamma^2 E^2 L^2} \sum_{i = 1}^N  p_i B_i^2
+ \gamma E \sum_{i = 1}^N p_i B_i^2
\text{.}
\end{align}
\end{small}
\vspace{-0.3cm}

After that, summing the above items from ${r} = 1$ to $R$ and dividing both sides by the multiplication of the learning rate and the total number of local mini-batch SGD steps, $\gamma T ( = \gamma RE)$, yields

\vspace{-0.3cm}
\begin{small}
\begin{align}\label{eq:coefficient mean F bar w r-1 2}
& \underbrace{\left( \frac{1}{2} - \frac{4 \gamma^2 E^2 L^2}{ 1 - 4 \gamma^2 E^2 L^2 } \right)}_{\triangleq H_1}
\frac{\sum_{r = 1}^{R} \| \nabla_{\mathbf{w}} F({\bar {\mathbf{w}}}_{r - 1}; \mathcal{S}_{\mathcal{D}}^{r,1}, \Delta \mathcal{S}_{\mathcal{D}}^{r,1}) \|^2 }{R} \notag \\
\leq & \underbrace{ \frac{1}{\gamma T} }_{\triangleq H_2}
\Big( F({\bar {\mathbf{w}}}_{0}; \mathcal{S}_{\mathcal{D}}^{1,1}, \Delta \mathcal{S}_{\mathcal{D}}^{1,1})
- F({\bar {\mathbf{w}}}_{R}; \mathcal{S}_{\mathcal{D}}^{R+1,1}, \Delta \mathcal{S}_{\mathcal{D}}^{R+1,1}) \Big)
+ \underbrace{ \frac{4 \gamma^2 E^2 L^2}{1 - 4 \gamma^2 E^2 L^2} }_{\triangleq H_3} \sum_{i = 1}^N  p_i V_i^2
+ \underbrace{ \frac{4 \gamma^2 E^2 L^2}{1 - 4 \gamma^2 E^2 L^2} }_{\triangleq H_3} \sum_{i = 1}^N  p_i B_i^2
+ \sum_{i = 1}^N p_i B_i^2
\text{.}
\end{align}
\end{small}
\vspace{-0.3cm}

\noindent
Let the learning rate $\gamma = N^{\frac{1}{2}} / (4L{T}^{\frac{1}{2}}) $ and the number of local updating steps $E \leq T^{\frac{1}{4}}/N^{\frac{3}{4}}$, where $T \geq N^3$ in order to ensure $E \geq 1$.
By this, the condition $0 < \gamma \leq \frac{1}{EL}$ in \eqref{eq:E F wr upper bound} can be satisfied, and $H_2 = 4L(TN)^{-\frac{1}{2}}$.
Next, since $\gamma E L \leq ( T N)^{-\frac{1}{4}}/4$, we have

\vspace{-0.3cm}
\begin{small}
\begin{equation}
H_3
\leq
\frac{ \frac{4}{ 4^2 }(TN)^{-\frac{1}{2}} }{1 - \frac{4}{ 4^2  } ( TN)^{-\frac{1}{2}} }
\overset{(a)}{\leq} \frac{ \frac{4}{ 4^2 }( T N)^{-\frac{1}{2}} }{1 - \frac{4}{ 4^2  } }
= \frac{1 }{ 3 ( T N)^{\frac{1}{2}} }
,
\end{equation}
\end{small}
\vspace{-0.3cm}

\noindent
where inequality (a) is due to $T \geq 1$ and $N \geq 1$.
Then,

\vspace{-0.1cm}
\begin{small}
\begin{equation}
H_1
= \frac{1}{2} - H_3
\geq \frac{1}{2} -  \frac{1 }{ 3 ( T N)^{\frac{1}{2}} }
\geq \frac{1}{2} - \frac{1}{3}
= \frac{1}{6}
\, .
\end{equation}
\end{small}
\vspace{-0.3cm}

Finally, by substituting above coefficients and $F({\bar {\mathbf{w}}}_{R};$ $\mathcal{S}_{\mathcal{D}}^{R+1,1}, \Delta \mathcal{S}_{\mathcal{D}}^{R+1,1}) \geq \underline{F}$ in Assumption \ref{assumption:lowered bounded} into \eqref{eq:coefficient mean F bar w r-1 2}, Theorem \ref{theorem:1} is proved.

{\hfill $\blacksquare$}

\subsection{Proof of Lemma \ref{lemma:1}}

Based on \eqref{bar wr - wr-1}, we have

\vspace{-0.3cm}
\begin{small}
\begin{align}\label{eq:nabla F w r-1 wr wr-1 2}
& \langle \nabla_{\mathbf{w}} F({\bar {\mathbf{w}}}_{r-1};\mathcal{S}_{\mathcal{D}}^{r,1}, \Delta \mathcal{S}_{\mathcal{D}}^{r,1}), {\bar {\mathbf{w}}}_{r} - {\bar {\mathbf{w}}}_{r-1} \rangle \notag \\
= & \Big\langle \nabla_{\mathbf{w}} F({\bar {\mathbf{w}}}_{r-1};\mathcal{S}_{\mathcal{D}}^{r,1}, \Delta \mathcal{S}_{\mathcal{D}}^{r,1}),
- \gamma \sum_{t = 1}^{E} \sum_{i = 1}^N p_i \nabla_{\mathbf{w}} F_{i} (\mathbf{w}^{r,t-1}_{i}; \mathcal{S}_{\mathcal{D}_i}^{r,t}, \Delta \mathcal{S}_{\mathcal{D}_i}^{r,t}) \Big \rangle \notag \\
= & - \gamma \sum_{t = 1}^{E} \Big\langle \nabla_{\mathbf{w}} F({\bar {\mathbf{w}}}_{r-1};\mathcal{S}_{\mathcal{D}}^{r,1}, \Delta \mathcal{S}_{\mathcal{D}}^{r,1}),
\sum_{i = 1}^N p_i \nabla_{\mathbf{w}} F_{i} (\mathbf{w}^{r,t-1}_{i}; \mathcal{S}_{\mathcal{D}_i}^{r,t}, \Delta \mathcal{S}_{\mathcal{D}_i}^{r,t}) \Big \rangle \notag \\
\overset{(a)}{=} & - \frac{\gamma E}{2} \big\| \nabla_{\mathbf{w}} F({\bar {\mathbf{w}}}_{r-1};\mathcal{S}_{\mathcal{D}}^{r,1}, \Delta \mathcal{S}_{\mathcal{D}}^{r,1}) \big\|^2
-  \frac{\gamma}{2}  \sum_{t = 1}^{E} \bigg\|  \sum_{i = 1}^N p_i \nabla_{\mathbf{w}} F_{i} (\mathbf{w}^{r,t-1}_{i}; \mathcal{S}_{\mathcal{D}_i}^{r,t}, \Delta \mathcal{S}_{\mathcal{D}_i}^{r,t}) \bigg\|^2 \notag \\
& + \frac{\gamma}{2} \sum_{t = 1}^{E} \bigg\| \nabla_{\mathbf{w}} F({\bar {\mathbf{w}}}_{r-1}; \mathcal{S}_{\mathcal{D}}^{r,1}, \Delta \mathcal{S}_{\mathcal{D}}^{r,1})
- \sum_{i = 1}^N p_i \nabla_{\mathbf{w}}F_i({\bar {\mathbf{w}}}_{r-1}; \mathcal{S}_{\mathcal{D}_i}^{r,1}, \Delta \mathcal{S}_{\mathcal{D}_i}^{r,1}) \notag \\
& \qquad\qquad\;\; + \sum_{i = 1}^N p_i \nabla_{\mathbf{w}}F_i({\bar {\mathbf{w}}}_{r-1}; \mathcal{S}_{\mathcal{D}_i}^{r,1}, \Delta \mathcal{S}_{\mathcal{D}_i}^{r,1})
- \sum_{i = 1}^N p_i \nabla_{\mathbf{w}}F_{i} (\mathbf{w}^{r,t-1}_{i}; \mathcal{S}_{\mathcal{D}_i}^{r,t}, \Delta \mathcal{S}_{\mathcal{D}_i}^{r,t}) \bigg\|^2 \notag \\
\overset{(b)}{\leq} & - \frac{\gamma}{2}  \sum_{t = 1}^{E} \left\| \nabla_{\mathbf{w}} F({\bar {\mathbf{w}}}_{r-1};\mathcal{S}_{\mathcal{D}}^{r,1}, \Delta \mathcal{S}_{\mathcal{D}}^{r,1}) \right\|^2
-  \frac{\gamma}{2}  \sum_{t = 1}^{E} \bigg\|  \sum_{i = 1}^N p_i \nabla_{\mathbf{w}} F_{i} (\mathbf{w}^{r,t-1}_{i}; \mathcal{S}_{\mathcal{D}_i}^{r,t}, \Delta \mathcal{S}_{\mathcal{D}_i}^{r,t}) \bigg\|^2 \notag \\
& + \gamma E \underbrace{\bigg\| \nabla_{\mathbf{w}} F({\bar {\mathbf{w}}}_{r-1};\mathcal{S}_{\mathcal{D}}^{r,1}, \Delta \mathcal{S}_{\mathcal{D}}^{r,1})
- \sum_{i = 1}^N p_i \nabla_{\mathbf{w}}F_i({\bar {\mathbf{w}}}_{r-1}; \mathcal{S}_{\mathcal{D}_i}^{r,1}, \Delta \mathcal{S}_{\mathcal{D}_i}^{r,1}) \bigg\|^2 }_{\triangleq A_1} \notag \\
& + \gamma \sum_{t = 1}^{E} \underbrace{ \bigg\| \sum_{i = 1}^N p_i \Big( \nabla_{\mathbf{w}} F_i({\bar {\mathbf{w}}}_{r-1}; \mathcal{S}_{\mathcal{D}_i}^{r,1}, \Delta \mathcal{S}_{\mathcal{D}_i}^{r,1})
- \nabla_{\mathbf{w}}F_i(\mathbf{w}^{r,t-1}_{i}; \mathcal{S}_{\mathcal{D}_i}^{r,t}, \Delta \mathcal{S}_{\mathcal{D}_i}^{r,t})  \Big) \bigg\|^2 }_{\triangleq A_2}
\text{,}
\end{align}
\end{small}
\vspace{-0.3cm}

\noindent
where equality (a) follows the basic identity $\langle {\mathbf{x}}_1,{\mathbf{x}}_2 \rangle = \frac{1}{2}( \| {\mathbf{x}}_1 \|^2 + \| {\mathbf{x}}_2 \|^2 - \| {\mathbf{x}}_1 - {\mathbf{x}}_2 \|^2 )$,
inequality (b) is due to $\|x_1+x_2\|^2 \leq 2\|x_1\|^2 + 2\|x_2\|^2$, and $A_1$ in equality (b) is bounded by

\vspace{-0.3cm}
\begin{small}
\begin{align}\label{eq:A1 bound}
A_1
\overset{(a)}{=} & \Big\| \sum_{i = 1}^N p_i \nabla_{\mathbf{w}}F_i({\bar {\mathbf{w}}}_{r-1};\mathcal{S}_{\mathcal{D}}^{r}, \Delta \mathcal{S}_{\mathcal{D}}^{r})
- \sum_{i = 1}^N p_i \nabla_{\mathbf{w}}F_i({\bar {\mathbf{w}}}_{r-1}; \mathcal{S}_{\mathcal{D}_i}^{r}, \Delta \mathcal{S}_{\mathcal{D}_i}^{r}) \Big\|^2 \notag \\
\overset{(b)}{\leq} & \sum_{i = 1}^N p_i \big\| \nabla_{\mathbf{w}} F_i({\bar {\mathbf{w}}}_{r-  1};\mathcal{S}_{\mathcal{D}}^{r},  \Delta \mathcal{S}_{\mathcal{D}}^{r})
 -   \nabla_{ \mathbf{w}} F_i({\bar {\mathbf{w}}}_{r- 1}; \mathcal{S}_{\mathcal{D}_i}^{r} ,  \Delta \mathcal{S}_{\mathcal{D}_i}^{r}) \big\|^2
\overset{(c)}{\leq} \sum_{i = 1}^N p_i B_i^2
\, \text{,}
\end{align}
\end{small}
\vspace{-0.3cm}

\noindent
where equality (a) comes from \eqref{ineq:nabla F Fi SD}, inequality (b) is by Jensen's Inequality, and inequality (c) is obtained from Assumption \ref{assumption:BN disturbance}.
Meanwhile, we can bound $A_2$ as

\vspace{-0.3cm}
\begin{small}
\begin{align}\label{eq:A2 bound}
A_2
\overset{(a)}{\leq} & \sum_{i = 1}^N  p_i
\big\| \nabla_{\mathbf{w}}F_i({\bar {\mathbf{w}}}_{r-1}; \mathcal{S}_{\mathcal{D}_i}^{r,1}, \Delta \mathcal{S}_{\mathcal{D}_i}^{r,1})
\! - \! \nabla_{\mathbf{w}}F_i(\mathbf{w}^{r,t-1}_{i}; \mathcal{S}_{\mathcal{D}_i}^{r,t}, \Delta \mathcal{S}_{\mathcal{D}_i}^{r,t})
\big\|^2
\overset{(b)}{\leq} L^2 \sum_{i = 1}^N  p_i \big\| \mathbf{w}^{r,t-1}_{i}
\! - \! \mathbf{\bar w}_{r-1} \big\|^2
,
\end{align}
\end{small}
\vspace{-0.3cm}

\noindent
where inequality (a) is due to Jensen's Inequality and inequality (b) is obtained from Assumption \ref{assumption:L continuous BN}.
Finally, by substituting \eqref{eq:A1 bound} and \eqref{eq:A2 bound} into \eqref{eq:nabla F w r-1 wr wr-1 2}, we can obtain Lemma \ref{lemma:1} directly.
\hfill $\blacksquare$

\subsection{Proof of Lemma \ref{lemma:3}}

Based on \eqref{eq:local SGD}, at the $r$-th iteration, local gradient parameters obtained by client $i \in [N]$ in the $t$-th local updating step are

\vspace{-0.3cm}
\begin{small}
\begin{align}
\mathbf{w}_{i}^{r,t-1}
= \mathbf{\bar w}_{r-1} - \gamma \sum_{e = 1}^{t-1} \nabla_{\mathbf{w}}F_i (\mathbf{w}_{i}^{r,e-1}; \mathcal{S}_{\mathcal{D}_i}^{r,e}, \Delta \mathcal{S}_{\mathcal{D}_i}^{r,e}) \, \text{.}
\end{align}
\end{small}
\vspace{-0.3cm}

\noindent
Therefore,

\vspace{-0.3cm}
\begin{small}
\begin{align}\label{w_i-w_bar}
& \big\| \mathbf{w}^{r,t-1}_{i} - {\bar {\mathbf{w}}}_{r-1} \big\|^2 \notag \\
= & \Big \| \gamma \sum_{e = 1}^{t-1} \nabla_{\mathbf{w}}F_i (\mathbf{w}_{i}^{r,e-1}; \mathcal{S}_{\mathcal{D}_i}^{e,t}, \Delta \mathcal{S}_{\mathcal{D}_i}^{e,t}) \Big\|^2 \notag \\
\leq & \gamma^2 (t - 1) \sum_{e = 1}^{t-1} \big\| \nabla_{\mathbf{w}}F_i (\mathbf{w}_{i}^{r,e-1}; \mathcal{S}_{\mathcal{D}_i}^{e,t}, \Delta \mathcal{S}_{\mathcal{D}_i}^{e,t}) \big\|^2
\leq \gamma^2 E \sum_{e = 1}^{t-1} \big\| \nabla_{\mathbf{w}}F_i (\mathbf{w}_{i}^{r,e-1}; \mathcal{S}_{\mathcal{D}_i}^{e,t}, \Delta \mathcal{S}_{\mathcal{D}_i}^{e,t}) \big\|^2 \notag \\
\leq &  4 \gamma^2 E \sum_{e = 1}^{t-1} \bigg(
\big\| \nabla_{\mathbf{w}}F_i (\mathbf{w}_{i}^{r,e-1}; \mathcal{S}_{\mathcal{D}_i}^{e,t}, \Delta \mathcal{S}_{\mathcal{D}_i}^{e,t})
- \nabla_{\mathbf{w}}F_i (\mathbf{\bar w}_{r-1}; \mathcal{S}_{\mathcal{D}_i}^{r,1}, \Delta \mathcal{S}_{\mathcal{D}_i}^{r,1}) \big\|^2 \notag \\
&
\qquad\qquad\quad
+ \big\| \nabla_{\mathbf{w}}F_i (\mathbf{\bar w}_{r-1}; \mathcal{S}_{\mathcal{D}_i}^{r,1}, \Delta \mathcal{S}_{\mathcal{D}_i}^{r,1})
 -  \nabla_{\mathbf{w}}F_i (\mathbf{\bar w}_{r-1}; \mathcal{S}_{\mathcal{D}}^{r,1}, \Delta \mathcal{S}_{\mathcal{D}}^{r,1}) \big\|^2 \notag \\
&
\qquad\qquad\quad
+ \big\| \nabla_{\mathbf{w}}F_i (\mathbf{\bar w}_{r-1}; \mathcal{S}_{\mathcal{D}}^{r,1}, \Delta \mathcal{S}_{\mathcal{D}}^{r,1})
 -  \nabla_{\mathbf{w}}F(\mathbf{\bar w}_{r-1}; \mathcal{S}_{\mathcal{D}}^{r,1}, \Delta \mathcal{S}_{\mathcal{D}}^{r,1}) \big\|^2
+ \big\| \nabla_{\mathbf{w}}F(\mathbf{\bar w}_{r-1}; \mathcal{S}_{\mathcal{D}}^{r,1}, \Delta \mathcal{S}_{\mathcal{D}}^{r,1}) \big\|^2 \bigg) \notag \\
\overset{(a)}{\leq} & 4 \gamma^2 E L^2 \sum_{e = 1}^{t-1} \big\| \mathbf{w}_{i}^{r,e-1} - \mathbf{\bar w}_{r-1} \big\|^2
+ 4 \gamma^2 E^2 B_i^2
+ 4 \gamma^2 E^2 V_i^2
+ 4 \gamma^2 E^2 \big\| \nabla_{\mathbf{w}}F(\mathbf{\bar w}_{r-1}; \mathcal{S}_{\mathcal{D}}^{r,1}, \Delta \mathcal{S}_{\mathcal{D}}^{r,1}) \big\|^2
,
\end{align}
\end{small}
\vspace{-0.3cm}

\noindent
where inequality (a) is obtained from Assumptions \ref{assumption:BN disturbance}, \ref{assumption:L continuous BN} and \ref{assumption:data var}.
Then, summing both sides of \eqref{w_i-w_bar} from $t = 1$ to $E$ yields

\vspace{-0.3cm}
\begin{small}
\begin{align}\label{Proof_Thm1_lemma_3_step1}
& \sum_{t = 1}^{E} \big\| \mathbf{w}^{r,t-1}_{i} - {\bar {\mathbf{w}}}_{r-1} \big\|^2 \notag \\
\leq & 4 \gamma^2 E L^2
\underbrace{\sum_{t = 1}^{E} \sum_{e = 1}^{t-1} \big\| \mathbf{w}_{i}^{r,e-1} - \mathbf{\bar w}_{r-1} \big\|^2 }_{\triangleq A_3}
+ 4 \gamma^2 E^3 B_i^2
+ 4 \gamma^2 E^3 V_i^2
+ 4 \gamma^2 E^3 \big\| \nabla_{\mathbf{w}}F(\mathbf{\bar w}_{r-1}; \mathcal{S}_{\mathcal{D}}^{r,1}, \Delta \mathcal{S}_{\mathcal{D}}^{r,1})) \big\|^2 \notag \\
\overset{(a)}{\leq} & 4 \gamma^2 E^2 L^2 \sum_{t = 1}^{E} \big\| \mathbf{w}_{i}^{r,t-1} - \mathbf{\bar w}_{r-1} \big\|^2
+ 4 \gamma^2 E^3 B_i^2
+ 4 \gamma^2 E^3 V_i^2
+ 4 \gamma^2 E^3 \big\| \nabla_{\mathbf{w}}F(\mathbf{\bar w}_{r-1}; \mathcal{S}_{\mathcal{D}}^{r,1}, \Delta \mathcal{S}_{\mathcal{D}}^{r,1})) \big\|^2
\text{,}
\end{align}
\end{small}
\vspace{-0.3cm}

\noindent
where inequality (a) is because the occurrence number of $\| \mathbf{w}^{r,t-1}_{i} - {\bar {\mathbf{w}}}_{r-1} \|^2  $ for each $t \in  [1, E]$ in term $A_3$ is less than the number of local updating steps $E$, and then
$A_3 \leq  E  \sum_{t = 1}^{E} \| \mathbf{w}_{i}^{r,t-1} - \mathbf{\bar w}_{r-1} \|^2  $.

Finally, rearranging the terms in (\ref{Proof_Thm1_lemma_3_step1}) yields Lemma \ref{lemma:3}.
\hfill $\blacksquare$

\section*{Supplementary Material}

\renewcommand\thesection{\Alph{section}}

\setcounter{section}{0}

\section{Derivation of (28)}\label{sec:grad mu equal proof}

Based on \eqref{eq:batch output} and the chain rule, the local gradient w.r.t. the batch mean in the last BN layer obtained by client $i \in [N]$ in FL is
\begin{equation}\label{eq:nabla muiL Fi}
{\nabla}_{\bm{\mu}_{i}^{L}}{F_i}
=
{\nabla}_{\mathbf{\hat Y}_i^L}{F_{i}} \cdot
{\nabla}_{\bm{\mu}_{i}^{L}}{\mathbf{\hat Y}_i^L}
+ {\nabla}_{(\bm{\sigma}_{i}^{L})^2}{F_i} \cdot {\nabla}_{\bm{\mu}_{i}^{L}}{(\bm{\sigma}_{i}^{L})^2}
\end{equation}
where

\vspace{-0.3cm}
\begin{small}
\begin{align}\label{eq:nabla muiL sigmaiL}
{\nabla}_{\bm{\mu}_{i}^{L}}{(\bm{\sigma}_{i}^{L})^2}
\overset{(a)}{=}
\operatorname{diag}
\bigg(
\sum_{c}
\bigg(
\int
2( \bm{\mu}_{i}^{L}  -  \mathbf{y}^{L} )
\cdot
f_{{\mathbf{Y}_i^L}|{\mathbf{X}_i^0}}(\mathbf{y}^L|{\mathbf{x}^0}  \in  c)
 \cdot
d \mathbf{y}^L
 \bigg)
 \cdot
\Pr (c | \mathcal{D}_i)
 \bigg)
\overset{(b)}{=}
\operatorname{diag} ( \mathbf{0} )
,
\end{align}
\end{small}
\vspace{-0.3cm}

\noindent
where equalities (a) and (b) are based on \eqref{eq:mu sigma ellj} and \eqref{eq:mu i 1}.
Thus, by substituting \eqref{eq:nabla muiL sigmaiL} into \eqref{eq:nabla muiL Fi}, we have

\vspace{-0.3cm}
\begin{small}
\begin{align}\label{eq:grad mu Fi}
{\nabla}_{\bm{\mu}_{i}^{L}}{F_i}
= &
\underbrace{{\nabla}_{\mathbf{Y}_i^{L+1}}{F_{i}}
\cdot
{\nabla}_{\mathbf{X}_i^L}{\mathbf{Y}_i^{L+1}}
\cdot
{\nabla}_{\mathbf{\hat Y}_i^L}{\mathbf{X}_i^{L}}}_{={\nabla}_{\mathbf{\hat Y}_i^L}{F_{i}}}
\cdot
{\nabla}_{(\bm{\mu}_{i}^{L})^2}{\mathbf{\hat Y}_i^L}
=
\mathbb{E}_{\mathcal{D}_{i}}
\Big[
{\nabla}_{\mathbf{Y}_i^{L+1}}{\mathcal{L}_{i}}
\cdot
{\nabla}_{\mathbf{X}_i^L}{\mathbf{Y}_i^{L+1}}
\cdot
{\nabla}_{\mathbf{\hat Y}_i^L}{\mathbf{X}_i^{L}}
\cdot
{\nabla}_{(\bm{\mu}_{i}^{L})^2}{\mathbf{\hat Y}_i^L}
\Big]
\notag \\
= &
\sum_{c}
\bigg( \int
\underbrace{ {\nabla}_{ \mathbf{y}^{L+1}}{\mathcal{L}_{i}}
\cdot
{\nabla}_{\mathbf{x}^L}{\mathbf{y}^{L+1}}
\cdot
{\nabla}_{\mathbf{\hat y}^L}{\mathbf{x}^L}
\cdot
{\nabla}_{\bm{\mu}_{i}^{L}}{\mathbf{\hat y}^L}}_{\text{\footnotesize (\ref{eq:grad mu Fi}a)}}
\cdot
f_{{\mathbf{Y}_i^L}|{\mathbf{X}_i^0}}(\mathbf{y}^L|{\mathbf{x}^0} \in c)
\cdot
d \mathbf{y}^L
\bigg)
\cdot
\Pr (c | \mathcal{D}_i)
,
\end{align}
\end{small}
\vspace{-0.3cm}

\noindent
Based on \eqref{eq:xiL}-\eqref{eq:grad x = diag gamma} and
${\nabla}_{\bm{\mu}_{i}^{L}}{\mathbf{\hat Y}_i^{L}} = \operatorname{diag}
(-1/
\sqrt{(\bm{\sigma}_{i}^{L})^{2} + \epsilon}
)$ which comes from \eqref{eq:batch output}, we can denote
\begin{equation}\label{eq:grad psi i}
{\text{(\ref{eq:grad mu Fi}a)}}
\triangleq
\psi_{\bm{\mu}}^L
\left(
\mathbf{y}^{L}, \bm{\mu}_{i}^{L}, (\bm{\sigma}_{i}^{L})^{2}, \bm{\gamma}^{L}, \bm{\beta}^{L}, \mathbf{\tilde w}^{L}
\right)
.
\end{equation}
Finally, by substituting \eqref{eq:grad psi i} into \eqref{eq:grad mu Fi}, we obtain \eqref{eq:grad mu Fi simplify}.
\hfill $\blacksquare$

\section{Derivation of (30) and (31)}\label{sec:grad sigma equal ell proof}


We first derive the gradient w.r.t. the input to the BN layer, which is the important intermediate variable in the backward propagation procedure.
In FL, the local gradient w.r.t. the input to the $L$-th BN layer obtained by client $i \in [N]$ is

\vspace{-0.3cm}
\begin{small}
\begin{align}\label{eq:nabla YiL Fi}
{\nabla}_{\mathbf{Y}_{i}^{L}}{F_i}
= &
{\nabla}_{\mathbf{\hat Y}_i^{L}}{F_{i}}
\cdot
{\nabla}_{ \mathbf{Y}_{i}^{L}}{\mathbf{\hat Y}_i^{L}}
+ {\nabla}_{ (\bm{\sigma}_{i}^{L})^2}{F_i} \cdot {\nabla}_{ \mathbf{Y}_{i}^{L}}{(\bm{\sigma}_{i}^{L})^2}
+
{\nabla}_{ \bm{\mu}_{i}^{L}}{F_i} \cdot {\nabla}_{ \mathbf{Y}_{i}^{L}}{\bm{\mu}_{i}^{L}} \notag \\
= &
\mathbb{E}_{\mathcal{D}_{i}}
\big[
{\nabla}_{\mathbf{Y}_i^{L+1}}{\mathcal{L}_{i}}
\cdot
{\nabla}_{\mathbf{X}_i^L}{\mathbf{Y}_i^{L+1}}
\cdot
{\nabla}_{\mathbf{\hat Y}_i^L}{\mathbf{X}_i^{L}}
\cdot
{\nabla}_{\mathbf{Y}_{i}^{L}}{\mathbf{\hat Y}_i^{L}}\big]
+ {\nabla}_{(\bm{\sigma}_{i}^{L})^2}{F_i} \cdot {\nabla}_{\mathbf{Y}_{i}^{L}}{(\bm{\sigma}_{i}^{L})^2}
+
{\nabla}_{\bm{\mu}_{i}^{L}}{F_i} \cdot {\nabla}_{\mathbf{Y}_{i}^{L}}{\bm{\mu}_{i}^{L}} \notag \\
\overset{(a)}{\triangleq}
&
\sum_{c}
\bigg( \int
\psi_{\mathbf{Y}}^L
\Big(
\mathbf{y}^{L}, \bm{\mu}_{i}^{L}, (\bm{\sigma}_{i}^{L})^{2}, \bm{\gamma}^{L}, \bm{\beta}^{L}, \mathbf{\tilde w}^{L},
{\nabla}_{(\bm{\sigma}_{i}^{L})^2}{F_i},
{\nabla}_{\bm{\mu}_{i}^{L}}{F_i}
\Big)
\cdot
f_{{\mathbf{Y}_i^L}|{\mathbf{X}_i^0}}(\mathbf{y}^L|{\mathbf{x}^0} \in c)
\cdot
d \mathbf{y}^L
\bigg)
\cdot
\Pr (c | \mathcal{D}_i)
,
\end{align}
\end{small}
\vspace{-0.3cm}

\noindent
where the notation simplicity in equality (a) comes from \eqref{eq:xiL}-\eqref{eq:grad x = diag gamma}, ${\nabla}_{\mathbf{Y}_{i}^{L}}{\mathbf{\hat Y}_i^{L}} = \operatorname{diag}
( 1 / \sqrt{(\bm{\sigma}_{i}^{L})^{2} + \epsilon} )$ due to \eqref{eq:batch output}
as well as ${\nabla}_{\mathbf{Y}_{i}^{L}}{(\bm{\sigma}_{i}^{L})^2} = 2({\mathbf{Y}_i^L} - \bm{\mu}_{i}^{L})$ according to \eqref{eq:sigma ellj}.
Then, similar to \eqref{eq:nabla YiL Fi}, the local gradient w.r.t. the input to the ($\ell+1$)-th ($\ell+1 \in [L]$) BN layer obtained by each client is

\vspace{-0.3cm}
\begin{small}
\begin{subequations}\label{eq:nabla Yiell Fi}
\begin{align}
& {\nabla}_{\mathbf{Y}_{i}^{\ell+1}}{F_i} \notag \\
= &
{\nabla}_{ \mathbf{\hat Y}_i^{\ell+1}}{F_{i}}
\cdot
{\nabla}_{ \mathbf{Y}_{i}^{\ell+1}}{\mathbf{\hat Y}_i^{\ell+1}}
+ {\nabla}_{(\bm{\sigma}_{i}^{\ell+1})^2}{F_i} \cdot {\nabla}_{\mathbf{Y}_{i}^{\ell+1}}{(\bm{\sigma}_{i}^{\ell+1})^2}
+
{\nabla}_{\bm{\mu}_{i}^{\ell+1}}{F_i} \cdot {\nabla}_{\mathbf{Y}_{i}^{\ell+1}}{\bm{\mu}_{i}^{\ell+1}} \notag \\
= &
\mathbb{E}_{\mathcal{D}_{i}}
\Big[
{\nabla}_{\mathbf{Y}_i^{L+1}}{\mathcal{L}_{i}}
\cdot
{\prod}_{\ell'=\ell+1}^{L}
\underbrace{{\nabla}_{\mathbf{X}_i^{\ell'}}{\mathbf{Y}_i^{\ell'+1}}
\cdot
{\nabla}_{\mathbf{\hat Y}_i^{\ell'}}{\mathbf{X}_i^{\ell'}}
\cdot
{\nabla}_{\mathbf{Y}_{i}^{\ell'}}{\mathbf{\hat Y}_i^{\ell'}}}_{\text{\footnotesize (\ref{eq:nabla Yiell Fi}a)}}
\Big]
\notag \\
&
+ {\sum}_{\ell'' = \ell+1}^{L}  {\nabla}_{ (\bm{\sigma}_{i}^{\ell''})^2}{ F_i}
\cdot
{\nabla}_{ \mathbf{Y}_{i}^{\ell''}}{ (\bm{\sigma}_{i}^{\ell''})^2}
\cdot
\Big(
{\prod}_{\ell'=\ell+1}^{\ell''-1}
\text{(\ref{eq:nabla Yiell Fi}a)}
\Big)
+
{\sum}_{\ell'' = \ell+1}^{L} {\nabla}_{ \bm{\mu}_{i}^{\ell''}}{F_i}
\cdot {\nabla}_{ \mathbf{Y}_{i}^{\ell''}}{\bm{\mu}_{i}^{\ell''}}
\cdot
\Big(
{\prod}_{\ell'=\ell+1}^{\ell''-1}
\text{(\ref{eq:nabla Yiell Fi}a)}
\Big)
\tag{\ref{eq:nabla Yiell Fi}b}
\label{eq:nabla Yiell Fi a} \\
\triangleq &
\sum_{c}
\bigg(  \int
\psi_{\mathbf{Y}}^{\ell+1}
\Big(
\mathbf{y}^{\ell+1},
\big\{ \bm{\mu}_{i}^{\ell'}  , (\bm{\sigma}_{i}^{\ell'})^{2}  , \bm{\gamma}^{\ell'}  , \bm{\beta}^{\ell'}
\big\}_{\ell'=\ell+1}^L ,
\big\{ \mathbf{\tilde w}^{\ell'} \big\}_{\ell'=\ell+1}^{L}
,
\big\{
{\nabla}_{ (\bm{\sigma}_{i}^{\ell'})^2}{F_i},
{\nabla}_{ \bm{\mu}_{i}^{\ell'}}{ F_i} \big\}_{\ell'=\ell+1}^L
\Big)
\notag \\
&\qquad\qquad
\cdot
f_{{\mathbf{Y}_i^{\ell+1}}|{\mathbf{X}_i^0}}(\mathbf{y}^{\ell+1}|{\mathbf{x}^0} \in c)
\cdot
d \mathbf{y}^{\ell+1}
\bigg)
\cdot \Pr (c | \mathcal{D}_i)
,
\tag{\ref{eq:nabla Yiell Fi}c}
\label{eq:nabla Yiell Fi b}
\end{align}
\end{subequations}
\end{small}
\vspace{-0.3cm}

\noindent
where the local gradient w.r.t. the output of DNN, ${\nabla}_{\mathbf{Y}_{i}^{L+1}}{F_i}$, also satisfies \eqref{eq:nabla Yiell Fi}.
After that, for each BN layer $\ell \in [L]$, the local gradient w.r.t. the batch variance obtained by client $i \in [N]$ is

\vspace{-0.3cm}
\begin{small}
\begin{align}\label{eq:grad sigma ell Fi}
{\nabla}_{(\bm{\sigma}_{i}^{\ell})^2}{F_i}
=
{\nabla}_{\mathbf{Y}_i^{\ell+1}}{F_{i}}
\cdot
{\nabla}_{\mathbf{X}_i^\ell}{\mathbf{Y}_i^{\ell+1}}
\cdot
{\nabla}_{\mathbf{\hat Y}_i^\ell}{\mathbf{X}_i^{\ell}}
\cdot
{\nabla}_{(\bm{\sigma}_{i}^{\ell})^2}{\mathbf{\hat Y}_i^\ell}
\overset{(a)}{=}
\eqref{eq:nabla Yiell Fi a}
\cdot
\underbrace{
{\nabla}_{\mathbf{X}_i^\ell}{\mathbf{Y}_i^{\ell+1}}
\cdot
{\nabla}_{\mathbf{\hat Y}_i^\ell}{\mathbf{X}_i^{\ell}}
\cdot
{\nabla}_{(\bm{\sigma}_{i}^{\ell})^2}{\mathbf{\hat Y}_i^\ell}
}_{\text{\footnotesize (\ref{eq:grad sigma ell Fi}b)}}
,
\end{align}
\end{small}
\vspace{-0.3cm}

\noindent
where equality (a) is calculated by multiplying every addition term in \eqref{eq:nabla Yiell Fi a} with (\ref{eq:grad sigma ell Fi}b).
Meanwhile, the local gradient w.r.t. the batch mean obtained by each client is

\vspace{-0.3cm}
\begin{small}
\begin{align}\label{eq:grad mu ell Fi}
{\nabla}_{\bm{\mu}_{i}^{\ell}}{F_i}
=
{\nabla}_{\mathbf{Y}_i^{\ell+1}}{F_{i}}
\cdot
{\nabla}_{\mathbf{X}_i^\ell}{\mathbf{Y}_i^{\ell+1}}
\cdot
{\nabla}_{\mathbf{\hat Y}_i^\ell}{\mathbf{X}_i^{\ell}}
\cdot
{\nabla}_{\bm{\mu}_{i}^{\ell}}{\mathbf{\hat Y}_i^\ell}
=
\eqref{eq:nabla Yiell Fi a}
\cdot
{\nabla}_{\mathbf{X}_i^\ell}{\mathbf{Y}_i^{\ell+1}}
\cdot
{\nabla}_{\mathbf{\hat Y}_i^\ell}{\mathbf{X}_i^{\ell}}
\cdot
{\nabla}_{\bm{\mu}_{i}^{\ell}}{\mathbf{\hat Y}_i^\ell}
.
\end{align}
\end{small}
\vspace{-0.3cm}

\noindent
Finally, with similar notation simplicity as \eqref{eq:nabla Yiell Fi b}, we obtain \eqref{eq:grad sigma ell Fi simplify} and \eqref{eq:grad mu ell Fi simplify}.
\hfill $\blacksquare$

\section{Proof of Property \ref{property:noniid necessity}}\label{sec:noniid necessity proof}

We consider the DNN model setting as in Fig. \ref{fig:plain DNN} and Definition \ref{definition:iid noniid}.
Then, in FL, based on \eqref{eq:batch input} and the chain rule, the local gradient w.r.t. the network coefficients $\mathbf{\tilde w}^{\ell}$ ($\ell = 0,\ldots, L$) obtained by each client $i \in [N]$ is

\vspace{-0.3cm}
\begin{small}
\begin{align}\label{eq:grad w ell Fi}
& {\nabla}_{\mathbf{\tilde w}^{\ell}}{F_i({\bar {\mathbf{w}}}_{r-1}; \mathcal{S}_{\mathcal{D}_i}^{r,1}, \Delta \mathcal{S}_{\mathcal{D}_i}^{r,1})}
=
{\nabla}_{\mathbf{Y}_i^{\ell+1}}{F_{i}}
\cdot
{\nabla}_{\mathbf{\tilde w}^{\ell}}{\mathbf{Y}_i^{\ell+1}}
\overset{(a)}{=}
\eqref{eq:nabla Yiell Fi a}
\cdot
{\nabla}_{\mathbf{\tilde w}^{\ell}}{\mathbf{Y}_i^{\ell+1}}
\notag \\
\overset{(b)}{\triangleq}
&
\sum_{c}
\bigg(  \int
\psi_{\mathbf{\tilde w}}^{\ell}
\Big(
\mathbf{x}^{\ell}
,
\big\{
\bm{\mu}_{i}^{\ell'}  , (\bm{\sigma}_{i}^{\ell'})^{2}  , \bm{\gamma}^{\ell'}  , \bm{\beta}^{\ell'}
\big\}_{\ell'=\ell+1}^L
,
\big\{\mathbf{\tilde w}^{\ell'} \big\}_{\ell'=\ell}^{L}
,
\big\{
{\nabla}_{ (\bm{\sigma}_{i}^{\ell'})^2}{F_i},
{\nabla}_{ \bm{\mu}_{i}^{\ell'}}{ F_i}
\big\}_{\ell'=\ell+1}^L
\Big)
\notag \\
& \qquad\qquad
\cdot
f_{{\mathbf{X}_i^\ell}|{\mathbf{X}_i^0}}(\mathbf{x}^\ell|{\mathbf{x}^0} \in c)
\cdot
d \mathbf{x}^\ell
\bigg)
\cdot \Pr (c | \mathcal{D}_i)
,
\end{align}
\end{small}
\vspace{-0.3cm}

\noindent
where in equality (a), term ${\nabla}_{\mathbf{\tilde w}^{\ell}}{\mathbf{Y}_i^{\ell+1}}$ depends on the value of both $\mathbf{X}_{i}^{\ell}$ and $\mathbf{\tilde w}^{\ell}$ due to \eqref{eq:batch input}.
Meanwhile, similar to \eqref{eq:nabla Yiell Fi b}, we make notation simplicity in equality (b).

Differently, based on \eqref{ineq:nabla F Fi SD}, if we adopt centralized learning, then ${\nabla}_{\mathbf{\tilde w}^{\ell}}  F_i ({\bar {\mathbf{w}}}_{r-1}; \mathcal{S}_{\mathcal{D}}^{r,1}, \Delta \mathcal{S}_{\mathcal{D}}^{r,1})$ is computed by \eqref{eq:grad w ell Fi} via replacing the local statistical parameters $\{\bm{\mu}_{i}^{\ell'}, (\bm{\sigma}_{i}^{\ell'})^{2}\}_{\ell'=\ell+1}^L$ with the global ones, $\{\bm{\mu}_{g}^{\ell'}, (\bm{\sigma}_{g}^{\ell'})^{2}\}_{\ell'=\ell+1}^L$, and replacing the related local gradients $\{
{\nabla}_{ (\bm{\sigma}_{i}^{\ell'})^2}{F_i},
{\nabla}_{ \bm{\mu}_{i}^{\ell'}}{ F_i}
\}_{\ell'=\ell+1}^L$ with the global ones, $\{
{\nabla}_{ (\bm{\sigma}_{g}^{\ell'})^2}{F},
{\nabla}_{ \bm{\mu}_{g}^{\ell'}}{ F}
\}_{\ell'=\ell+1}^L$.
However, it can be observed from \eqref{eq:grad sigma ell Fi simplify} and \eqref{eq:grad mu ell Fi simplify} that, even if we make the local and the global statistical parameters consistent, namely $\bm{\mu}_{i}^{\ell} = \bm{\mu}_{g}^{\ell}$ and $(\bm{\sigma}_{i}^{\ell})^2 = (\bm{\sigma}_{g}^{\ell})^2$, $\forall \ell \in [L]$, we still generally have \eqref{ineq:gradient F sigma mu i g j} in the non-i.i.d. data case due to \eqref{ineq:pdf y1 ig} or $ \Pr (c | \mathcal{D}_i) \neq \Pr (c | \mathcal{D})$ or both.
Finally, \eqref{ineq:gradient F sigma mu i g j} results in \eqref{ineq:nabla w Fi F}.
\hfill $\blacksquare$

\section{Class distributions under different distribution shifts}\label{appendix:Class distributions}

The specific class distributions of the local training samples depicted in Fig. \ref{fig:performance shift} are presented below.

\begin{enumerate}
\item
Each client has training samples of 2 classes:
\begin{table}[!h]
\centering
\begin{tabular}{c|cccccccccc}
\hline
\multicolumn{1}{c|}{\multirow{2}{*}{Client}} & \multicolumn{10}{c}{Label} \\ \cline{2-11}
\multicolumn{1}{c|}{}                        & 1 & 2 & 3 & 4 & 5 & 6 & 7 & 8 & 9 & 10 \\ \hline
1                                            & 1/2 & 1/2 & 0 & 0 & 0 & 0 & 0 & 0 & 0 & 0 \\
2                                            & 0 & 0 & 1/2 & 1/2 & 0 & 0 & 0 & 0 & 0 & 0 \\
3                                            & 0 & 0 & 0 & 0 & 1/2 & 1/2 & 0 & 0 & 0 & 0 \\
4                                            & 0 & 0 & 0 & 0 & 0 & 0 & 1/2 & 1/2 & 0 & 0 \\
5                                            & 0 & 0 & 0 & 0 & 0 & 0 & 0 & 0 & 1/2 & 1/2 \\ \hline
\end{tabular}
\end{table}

\item
Each client has training samples of 4 classes:
\begin{table}[!h]
\centering
\begin{tabular}{c|cccccccccc}
\hline
\multicolumn{1}{c|}{\multirow{2}{*}{Client}} & \multicolumn{10}{c}{Label} \\ \cline{2-11}
\multicolumn{1}{c|}{}                        & 1 & 2 & 3 & 4 & 5 & 6 & 7 & 8 & 9 & 10 \\ \hline
1                                            & 1/4 & 1/4 & 1/4 & 1/4 & 0 & 0 & 0 & 0 & 0 & 0 \\
2                                            & 0 & 0 & 1/4 & 1/4 & 1/4 & 1/4 & 0 & 0 & 0 & 0 \\
3                                            & 0 & 0 & 0 & 0 & 1/4 & 1/4 & 1/4 & 1/4 & 0 & 0 \\
4                                            & 0 & 0 & 0 & 0 & 0 & 0 & 1/4 & 1/4 & 1/4 & 1/4 \\
5                                            & 1/4 & 1/4 & 0 & 0 & 0 & 0 & 0 & 0 & 1/4 & 1/4 \\ \hline
\end{tabular}
\vspace{-0.1cm}
\end{table}

\item
Each client has training samples of 6 classes:
\begin{table}[!h]
\centering
\begin{tabular}{c|cccccccccc}
\hline
\multicolumn{1}{c|}{\multirow{2}{*}{Client}} & \multicolumn{10}{c}{Label} \\ \cline{2-11}
\multicolumn{1}{c|}{}                        & 1 & 2 & 3 & 4 & 5 & 6 & 7 & 8 & 9 & 10 \\ \hline
1                                            & 1/6 & 1/6 & 1/6 & 1/6 & 1/6 & 1/6 & 0 & 0 & 0 & 0 \\
2                                            & 0 & 0 & 1/6 & 1/6 & 1/6 & 1/6 & 1/6 & 1/6 & 0 & 0 \\
3                                            & 0 & 0 & 0 & 0 & 1/6 & 1/6 & 1/6 & 1/6 & 1/6 & 1/6 \\
4                                            & 1/6 & 1/6 & 0 & 0 & 0 & 0 & 1/6 & 1/6 & 1/6 & 1/6 \\
5                                            & 1/6 & 1/6 & 1/6 & 1/6 & 0 & 0 & 0 & 0 & 1/6 & 1/6 \\ \hline
\end{tabular}
\end{table}

\item
Each client has training samples of 8 classes:
\begin{table}[!h]
\centering
\begin{tabular}{c|cccccccccc}
\hline
\multicolumn{1}{c|}{\multirow{2}{*}{Client}} & \multicolumn{10}{c}{Label} \\ \cline{2-11}
\multicolumn{1}{c|}{}                        & 1 & 2 & 3 & 4 & 5 & 6 & 7 & 8 & 9 & 10 \\ \hline
1                                            & 1/8 & 1/8 & 1/8 & 1/8 & 1/8 & 1/8 & 1/8 & 1/8 & 0 & 0 \\
2                                            & 0 & 0 & 1/8 & 1/8 & 1/8 & 1/8 & 1/8 & 1/8 & 1/8 & 1/8 \\
3                                            & 1/8 & 1/8 & 0 & 0 & 1/8 & 1/8 & 1/8 & 1/8 & 1/8 & 1/8 \\
4                                            & 1/8 & 1/8 & 1/8 & 1/8 & 0 & 0 & 1/8 & 1/8 & 1/8 & 1/8 \\
5                                            & 1/8 & 1/8 & 1/8 & 1/8 & 1/8 & 1/8 & 0 & 0 & 1/8 & 1/8 \\ \hline
\end{tabular}
\end{table}

\item
Each client has training samples of 10 classes:
\begin{table}[!h]
\centering
\begin{tabular}{c|cccccccccc}
\hline
\multicolumn{1}{c|}{\multirow{2}{*}{Client}} & \multicolumn{10}{c}{Label} \\ \cline{2-11}
\multicolumn{1}{c|}{}                        & 1 & 2 & 3 & 4 & 5 & 6 & 7 & 8 & 9 & 10 \\ \hline
1                                            & 1/10 & 1/10 & 1/10 & 1/10 & 1/10 & 1/10 & 1/10 & 1/10 & 1/10 & 1/10 \\
2                                            & 1/10 & 1/10 & 1/10 & 1/10 & 1/10 & 1/10 & 1/10 & 1/10 & 1/10 & 1/10 \\
3                                            & 1/10 & 1/10 & 1/10 & 1/10 & 1/10 & 1/10 & 1/10 & 1/10 & 1/10 & 1/10 \\
4                                            & 1/10 & 1/10 & 1/10 & 1/10 & 1/10 & 1/10 & 1/10 & 1/10 & 1/10 & 1/10 \\
5                                            & 1/10 & 1/10 & 1/10 & 1/10 & 1/10 & 1/10 & 1/10 & 1/10 & 1/10 & 1/10 \\ \hline
\end{tabular}
\end{table}

\end{enumerate}

\section{Influence of Different Batch Sizes}

\begin{figure}[!h]
\centering
\subfigure[i.i.d.]{
\includegraphics[width= 2 in ]{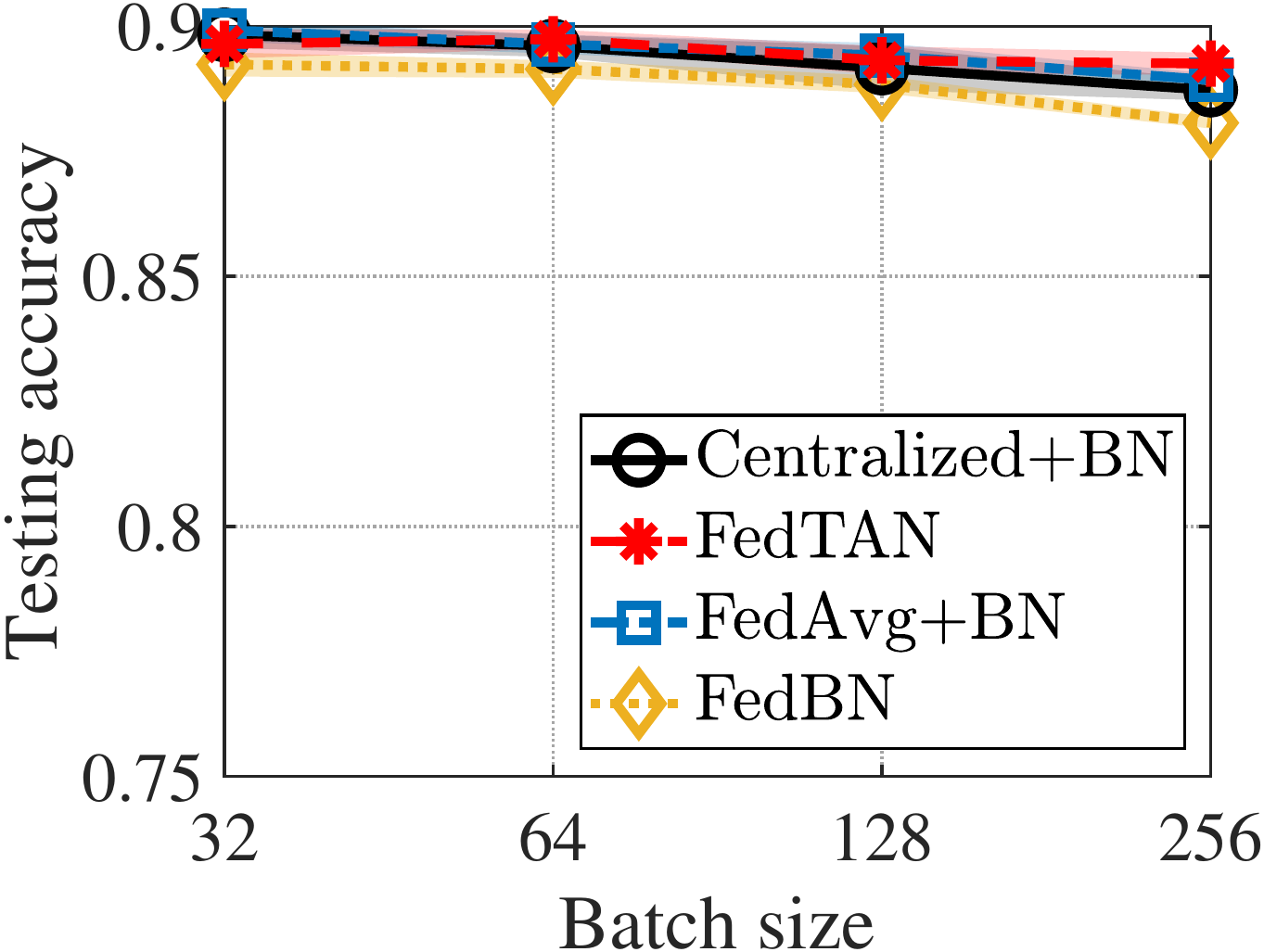}}
\subfigure[Non-i.i.d.]{
\includegraphics[width= 2 in ]{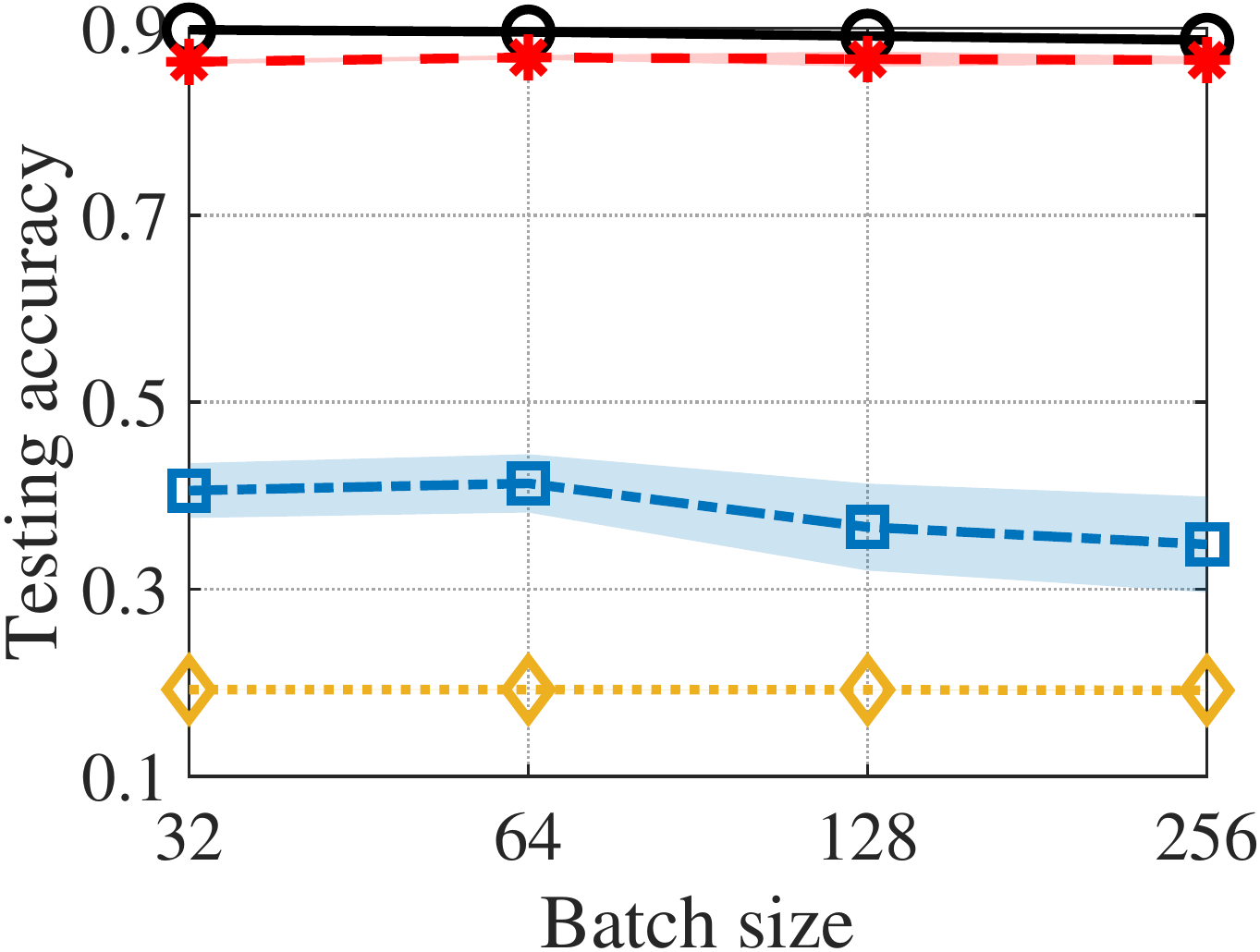}}
\caption{FL Performance on CIFAR-10 dataset with different batch sizes. The mini-batch SGD without momentum or weight decay is adopted.}
\label{fig:performance batch}
\end{figure}

Fig. \ref{fig:performance batch} compares the performance of various FL schemes on CIFAR-10 dataset across different batch sizes.
It can be observed that all FL schemes exhibit relatively stable performance despite variations in batch size.
Notably, the proposed \texttt{FedTAN} consistently performs comparably to centralized learning across all cases.
However, both \texttt{FedAvg+BN} and \texttt{FedBN} exhibit poor performance under the non-i.i.d. data distribution, regardless of the batch size value.

\section{Performance on DomainNet dataset}\label{appendix:DomainNet}

\subsection{Dataset and DNN Model}

In the experiments, we train AlexNet \cite{li2021fedbn} on DomainNet dataset \cite{peng2019moment}.
DomainNet dataset consists of natural images from six different data sources: clipart, infograph, painting, quickdraw, real, and sketch.
It includes 345 classes of objects, and we conduct our experiments using the top ten most common classes, following the approach in the paper of \texttt{FedBN} \cite{li2021fedbn}.
We assume $N=6$ clients, and each client possesses training samples from one data source.

\subsection{Parameter values}

We adopt the identical parameter setting as described in \cite{li2021fedbn}, with a batch size of 32 and $\gamma$ set to 0.01, without applying momentum or weight decay.

\subsection{Numerical Results}

We label the \texttt{FedBN} schemes updated by clients with local datasets from clipart, infograph, painting, quickdraw, real, and sketch as \texttt{FedBN-C}, \texttt{FedBN-I}, \texttt{FedBN-P}, \texttt{FedBN-Q}, \texttt{FedBN-R}, and \texttt{FedBN-S}, respectively.
The same notation applies to the \texttt{SingleNet} variants.

\begin{figure}[!h]
\begin{minipage}[!h]{1\linewidth}
\centering
\includegraphics[width= 6.5 in ]{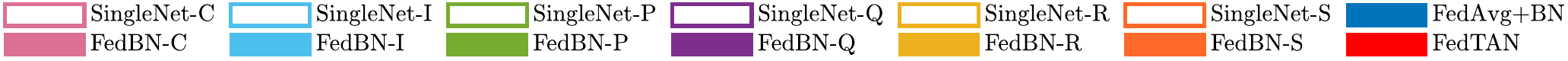}
\end{minipage}
\begin{minipage}[!h]{1\linewidth}
\centering
\subfigure[Clipart.]{
\includegraphics[width= 1.56 in ]{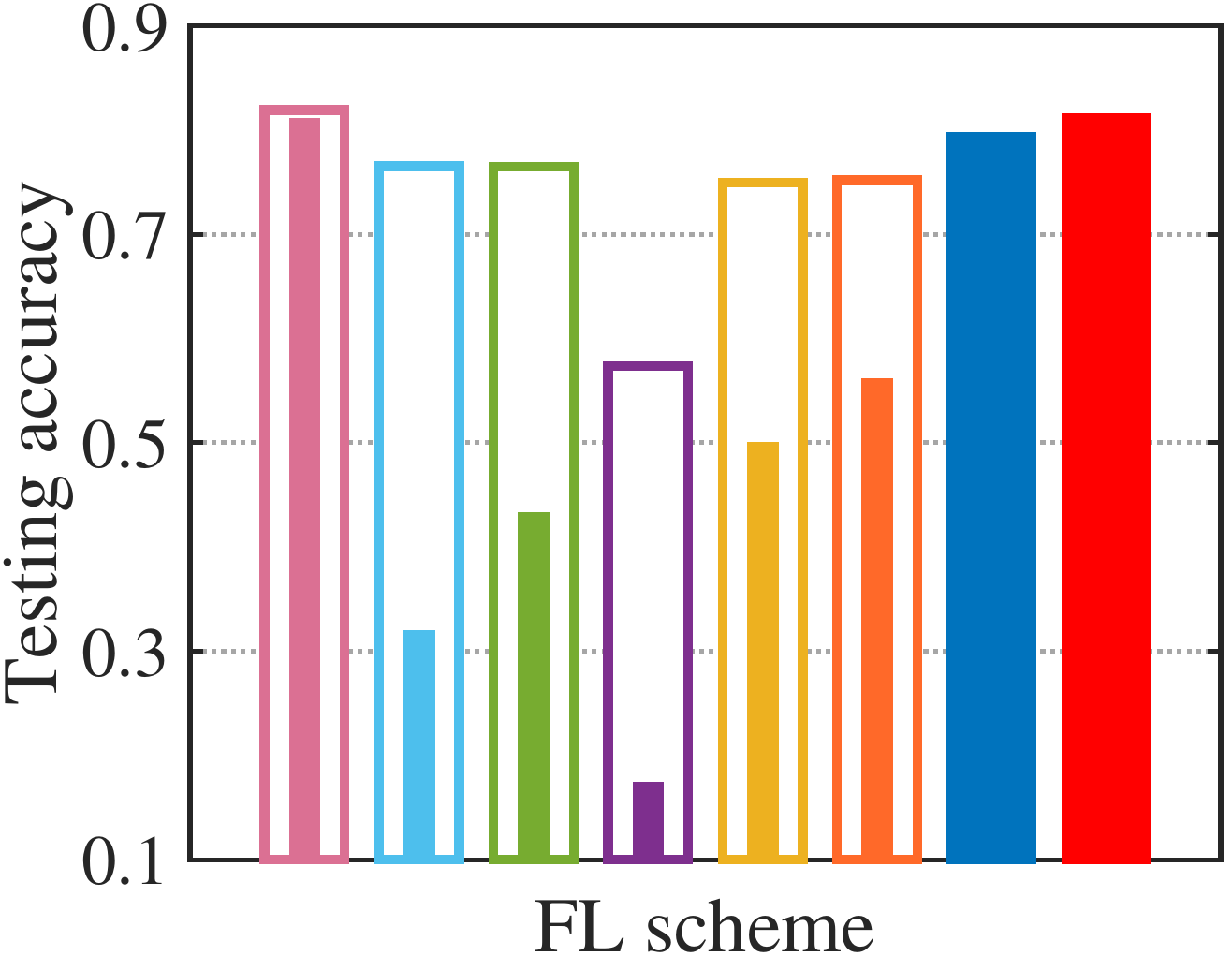}}
\subfigure[Infograph.]{
\includegraphics[width= 1.56 in ]{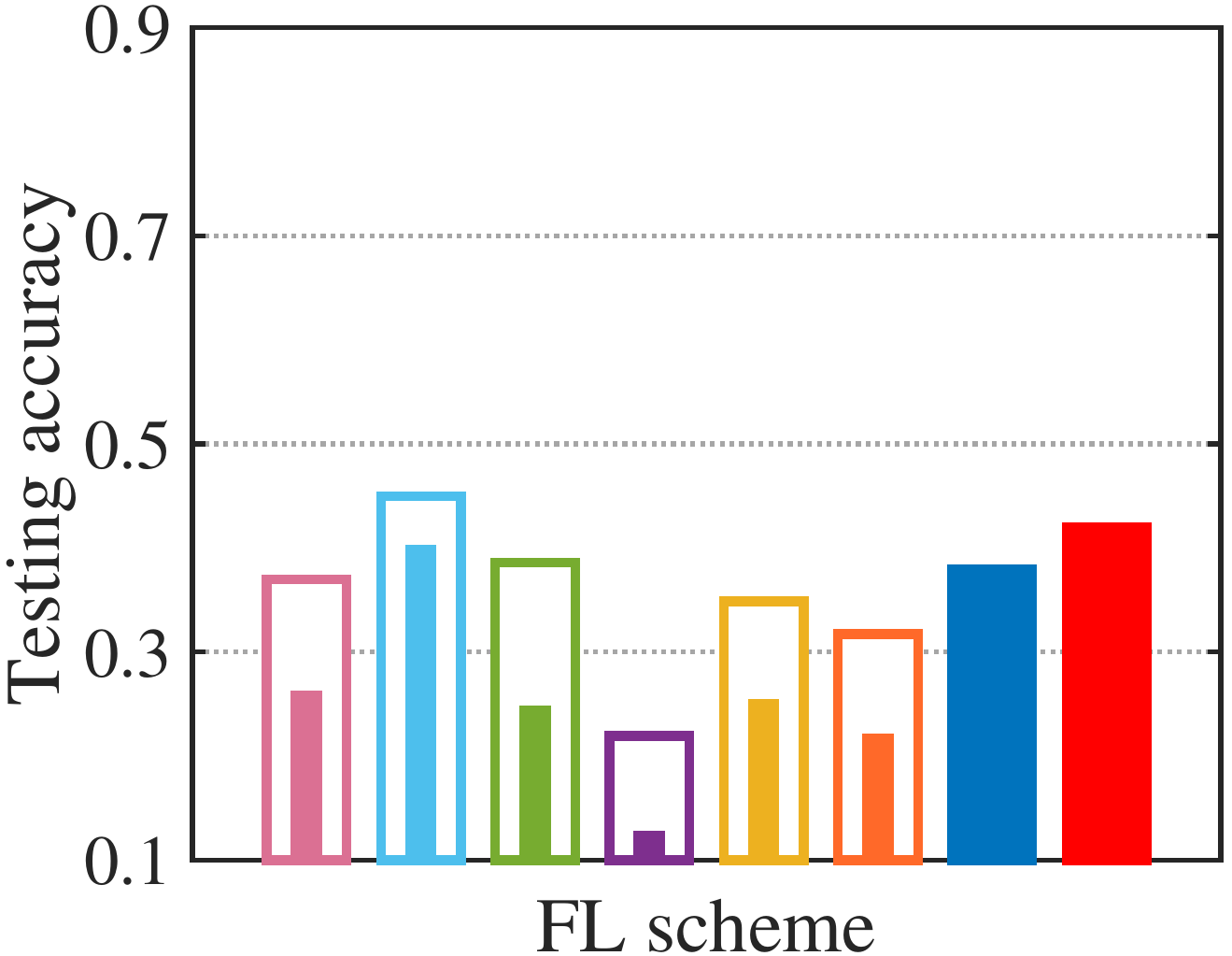}}
\subfigure[Painting.]{
\includegraphics[width= 1.56 in ]{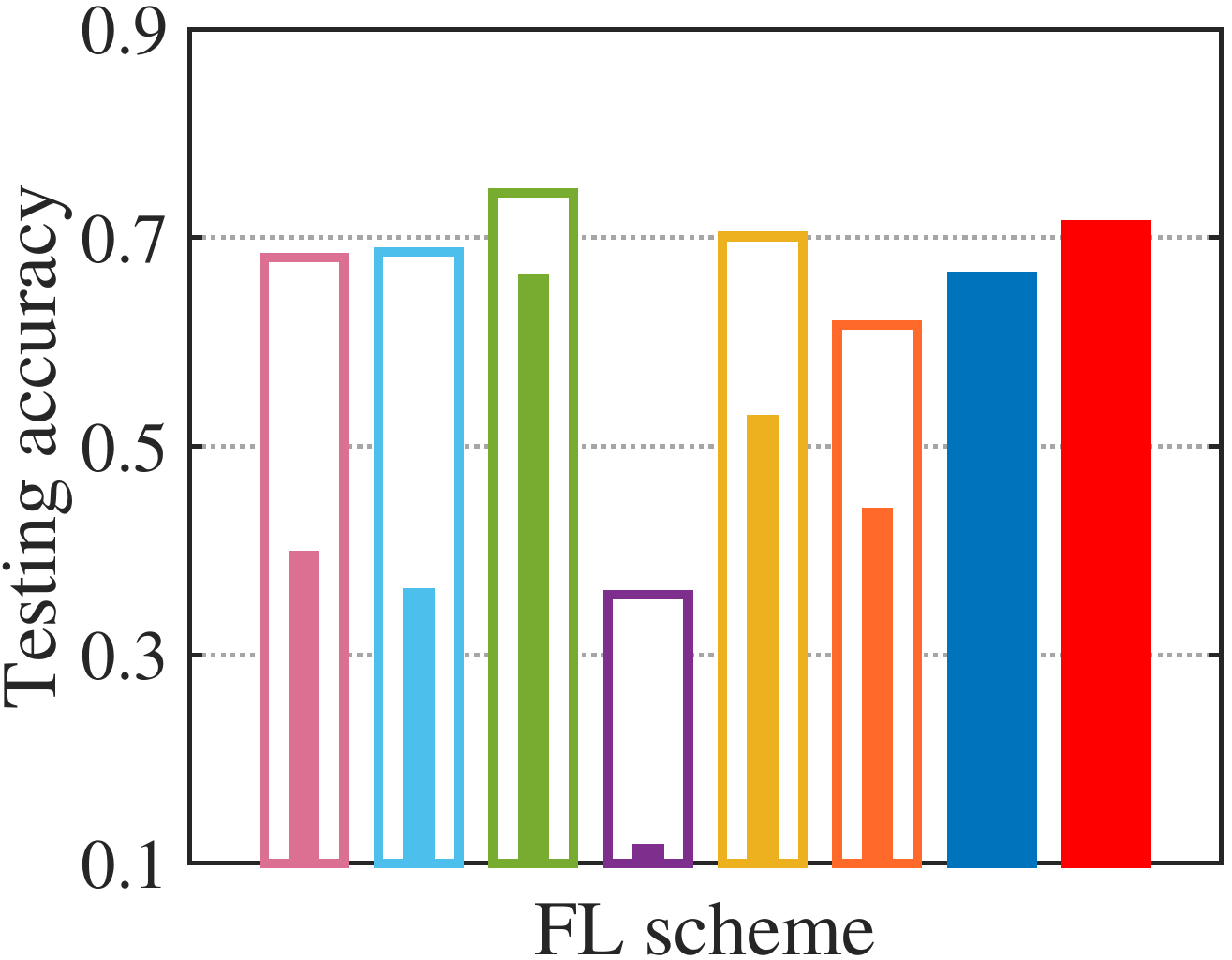}}
\subfigure[Quickdraw.]{
\includegraphics[width= 1.56 in ]{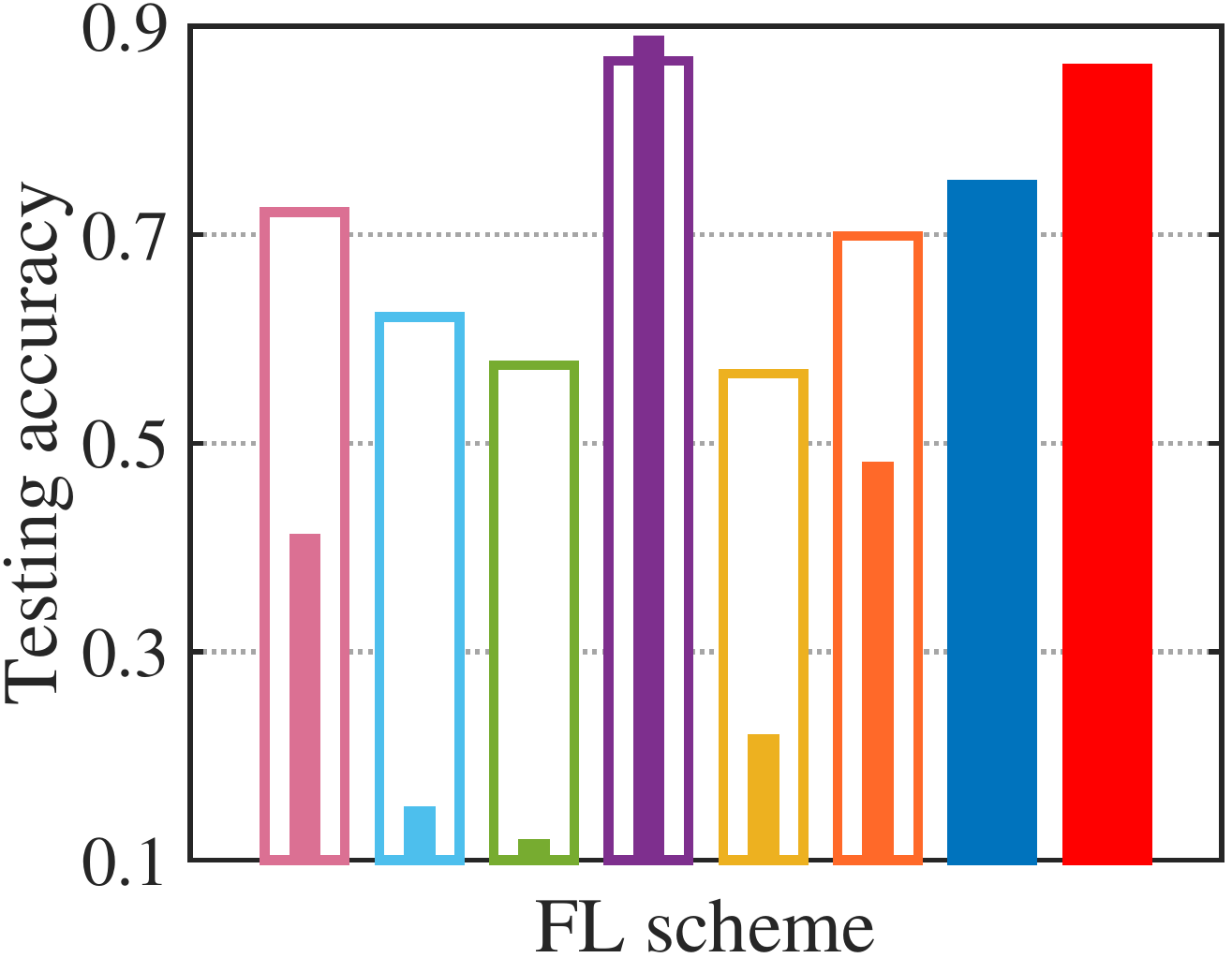}}
\subfigure[Real.]{
\includegraphics[width= 1.56 in ]{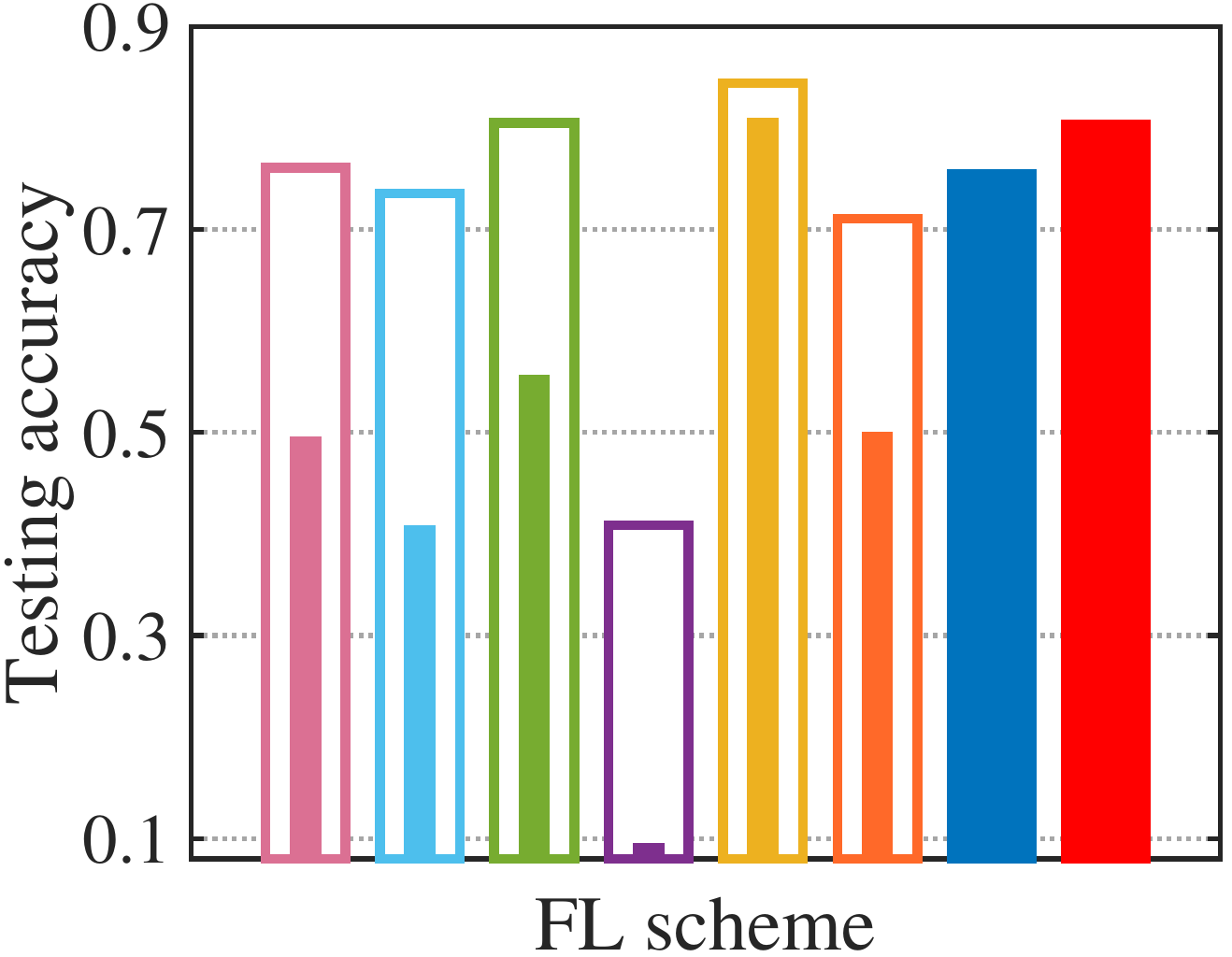}}
\subfigure[Sketch.]{
\includegraphics[width= 1.56 in ]{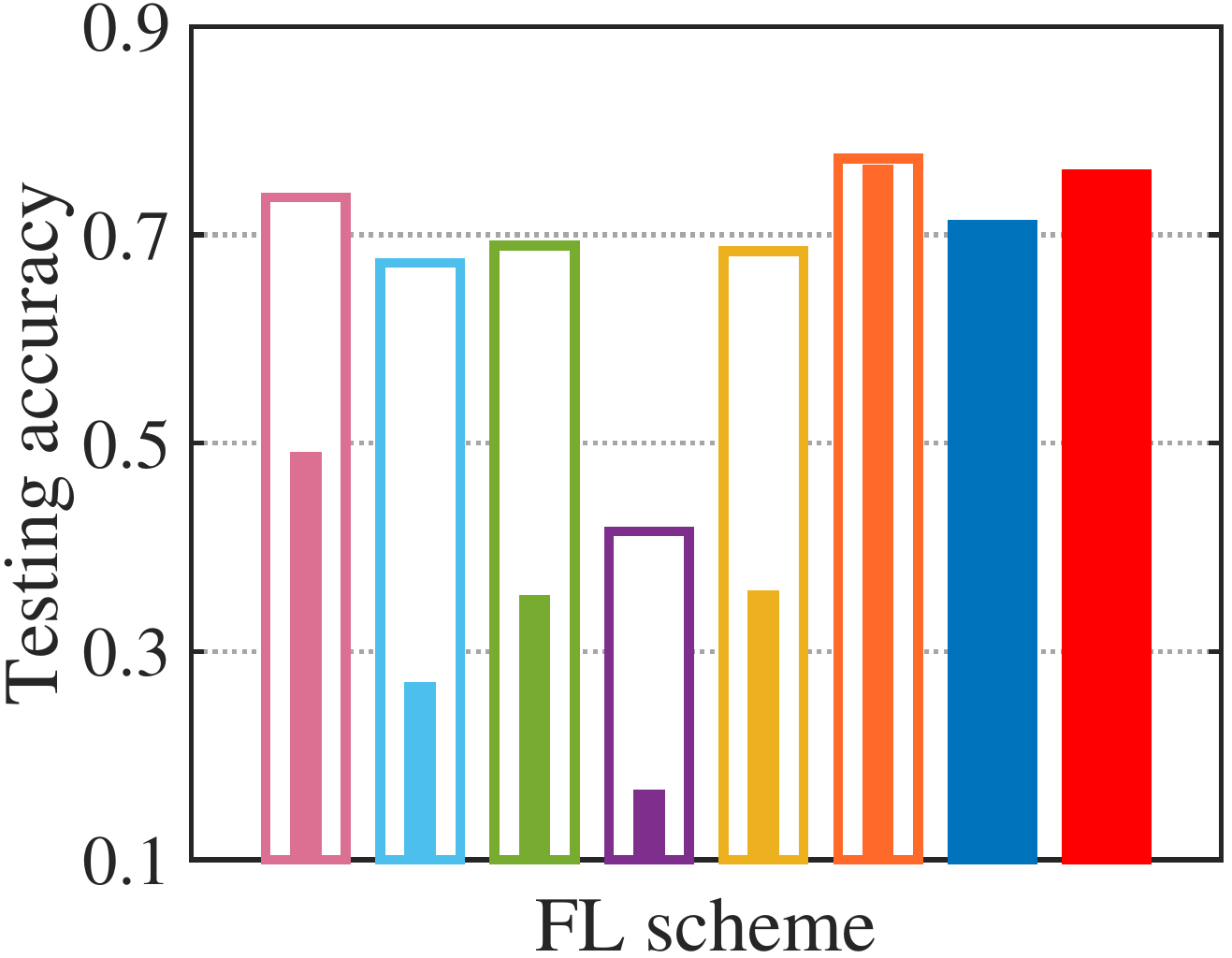}}
\subfigure[Global.]{
\includegraphics[width= 1.56 in ]{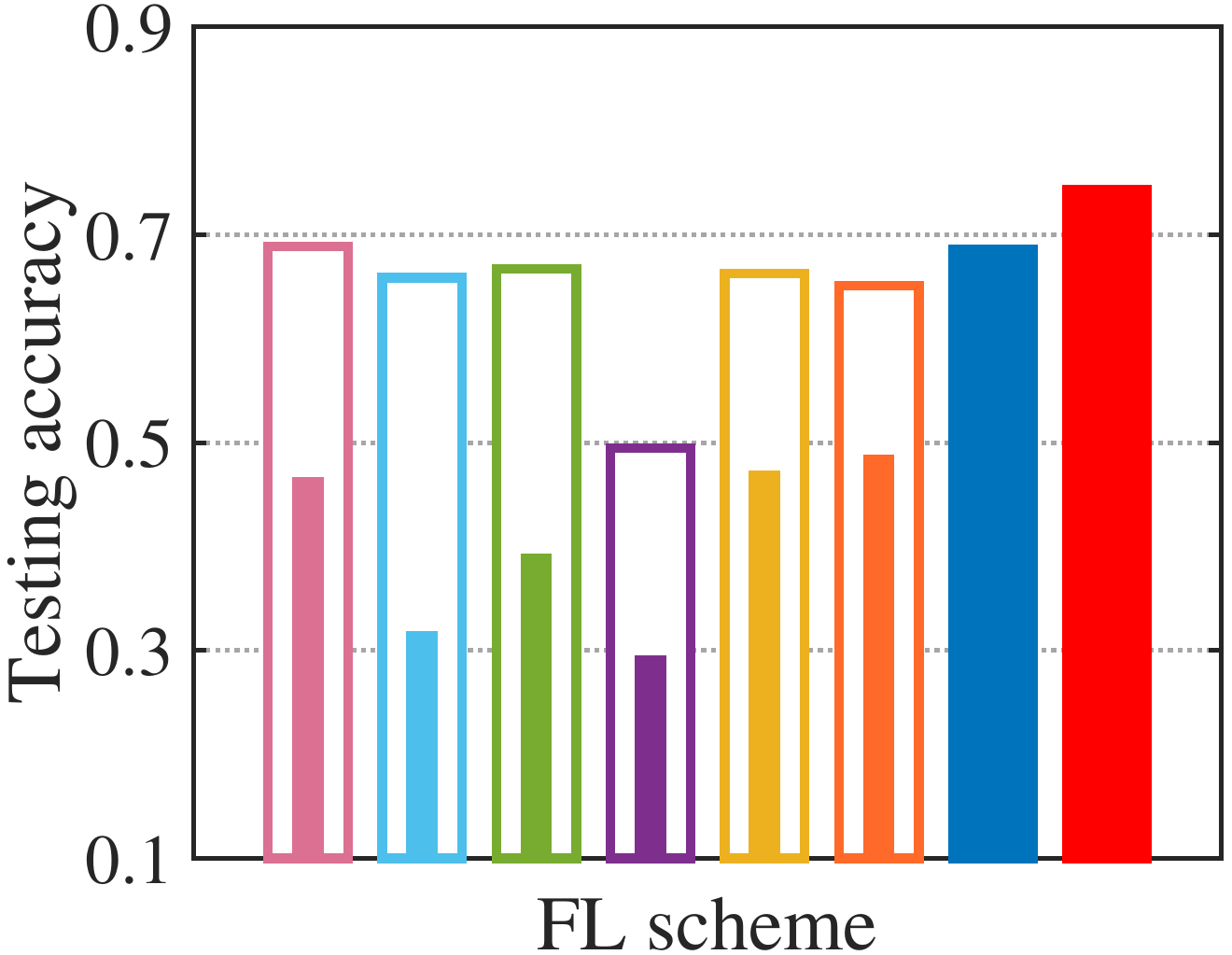}}
\caption{Performance of different FL schemes on DomainNet dataset.}
\label{fig:performance DomainNet}
\end{minipage}
\end{figure}

Fig. \ref{fig:performance DomainNet} presents a comparison of the performance of various FL schemes on testing datasets from different data sources and the whole global testing dataset.
Similar to Fig. \ref{fig:performance Office Caltech}, \texttt{FedBN} outperforms \texttt{SingleNet} when the training and testing datasets come from different sources, thanks to its improved generalization ability.
However, \texttt{FedBN} would still perform worse when datasets have different sources, such as \texttt{FedBN-Q} in Fig. \ref{fig:performance DomainNet}(a)-\ref{fig:performance DomainNet}(c), \ref{fig:performance DomainNet}(e), and \ref{fig:performance DomainNet}(f), and \texttt{FedBN-S} in Fig. \ref{fig:performance DomainNet}(a)-\ref{fig:performance DomainNet}(e).
In contrast, \texttt{FedTAN} demonstrates superior generalization ability by aggregating model parameters in the server, leading to satisfactory performance across all testing datasets.
Furthermore, \texttt{FedTAN} exhibits greater robustness compared to \texttt{FedAvg+BN} which is susceptible to gradient deviation.

\section{Transmitted data volume of different FL schemes}\label{appendix:transmitted data volume}

\subsection{\texttt{FedAvg+GN}}
As ResNet-20 with GN contains 269722 parameters, the number of DNN parameters sent by the server to five clients and the number of parameters uploaded by five clients to the server are 269722 and 269722$\times$5, respectively.
Hence, when using 32-bit (4-byte) floating point numbers to transmit DNN parameters, the total transmitted data volume between the server and clients per iteration in \texttt{FedAvg+GN} is $269722\times 6 \times 4 / 1024 / 1024 = 6.1734$ MB.

\subsection{\texttt{FedAvg+BN}}
Using the same calculation process as before, and with 271098 parameters in ResNet-20 with BN, the total transmitted data volume in \texttt{FedAvg+BN} is $271098\times 6 \times 4 / 1024 / 1024 = 6.2049$ MB per iteration.

\subsection{\texttt{FedTAN}}

ResNet-20 has 19 convolution layers, each followed by BN \cite{he2016deep}.
Then, with 16, 32, and 64 filters respectively in the first 7, middle 6, and last 6 convolution layers, respectively, the total number of statistical parameters (including batch mean and batch variance) in BN layers are $2\times ( 16 \times 7 + 32 \times 6 + 64 \times 6) = 1376$.
As a result, during the layer-wise aggregations in Algorithm \ref{algorithm:FedTAN modification forward}, the number of statistical parameters uploaded from five clients to the server and sent from the server to the clients is 1376$\times$5 and 1376, respectively.
Therefore, for each iteration, the number of extra transmitted statistical parameters caused by Algorithm \ref{algorithm:FedTAN modification forward} is 1376$\times$6, and similarly, the number of extra transmitted gradients w.r.t. the statistical parameters caused by Algorithm \ref{algorithm:FedTAN modification backward} is 1376$\times$6 as well.
Finally, the total amount of transmitted data between the server and clients per iteration in \texttt{FedTAN}, including the bits exchanged in the previous \texttt{FedAvg+BN}, is $(271098 \times 6 + 1376 \times 6 + 1376 \times 6) \times 4 / 1024 / 1024 = 6.2679$ MB.
\hfill $\blacksquare$

\section{Performance of \texttt{FedTAN-II} on MNIST Dataset}\label{appendix:FedTAN-II MNIST}

\begin{figure}[h]
\centering
\subfigure[Without momentum.]{
\includegraphics[width= 2 in ]{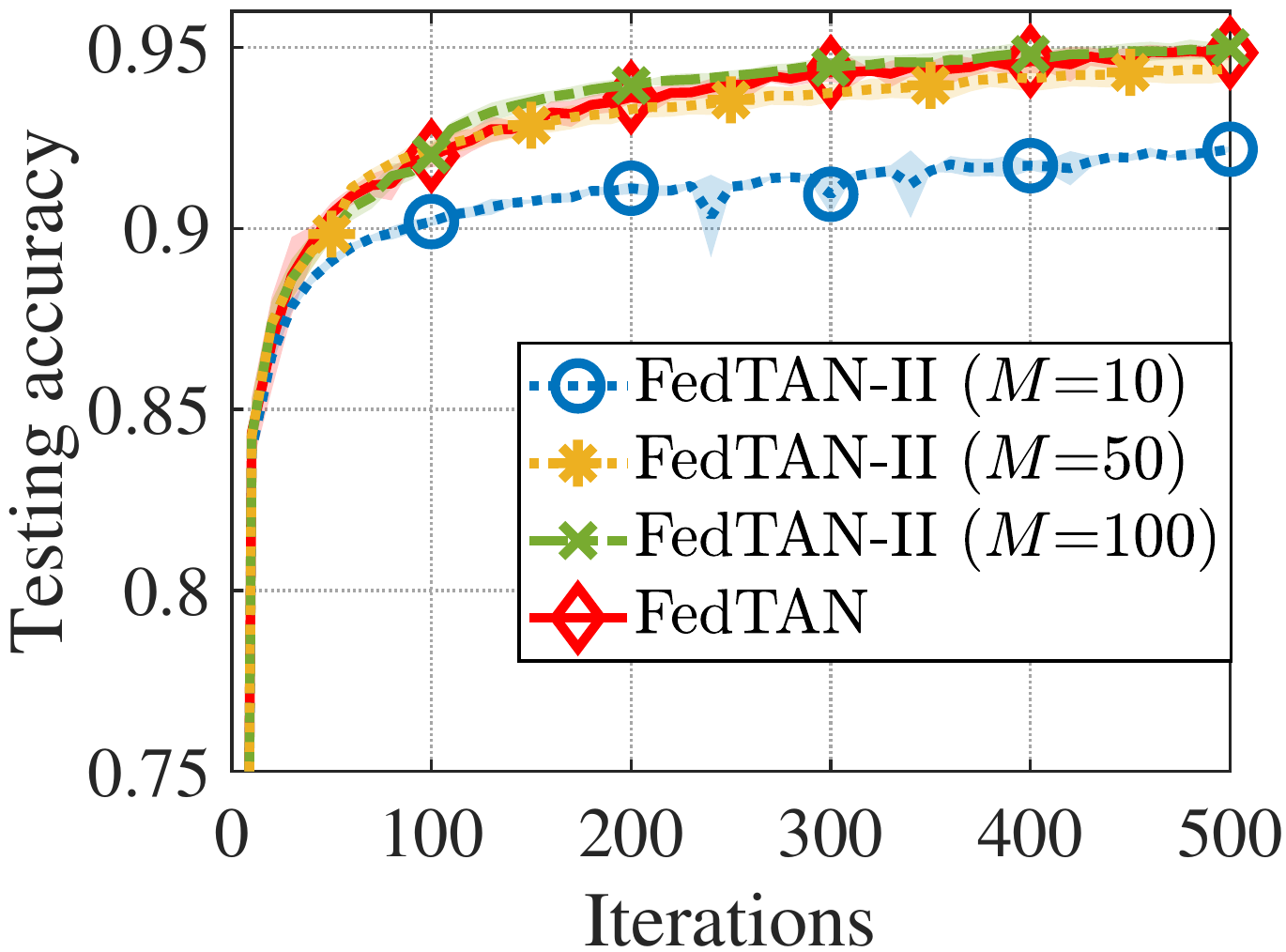}}
\subfigure[Without momentum.]{
\includegraphics[width= 2 in ]{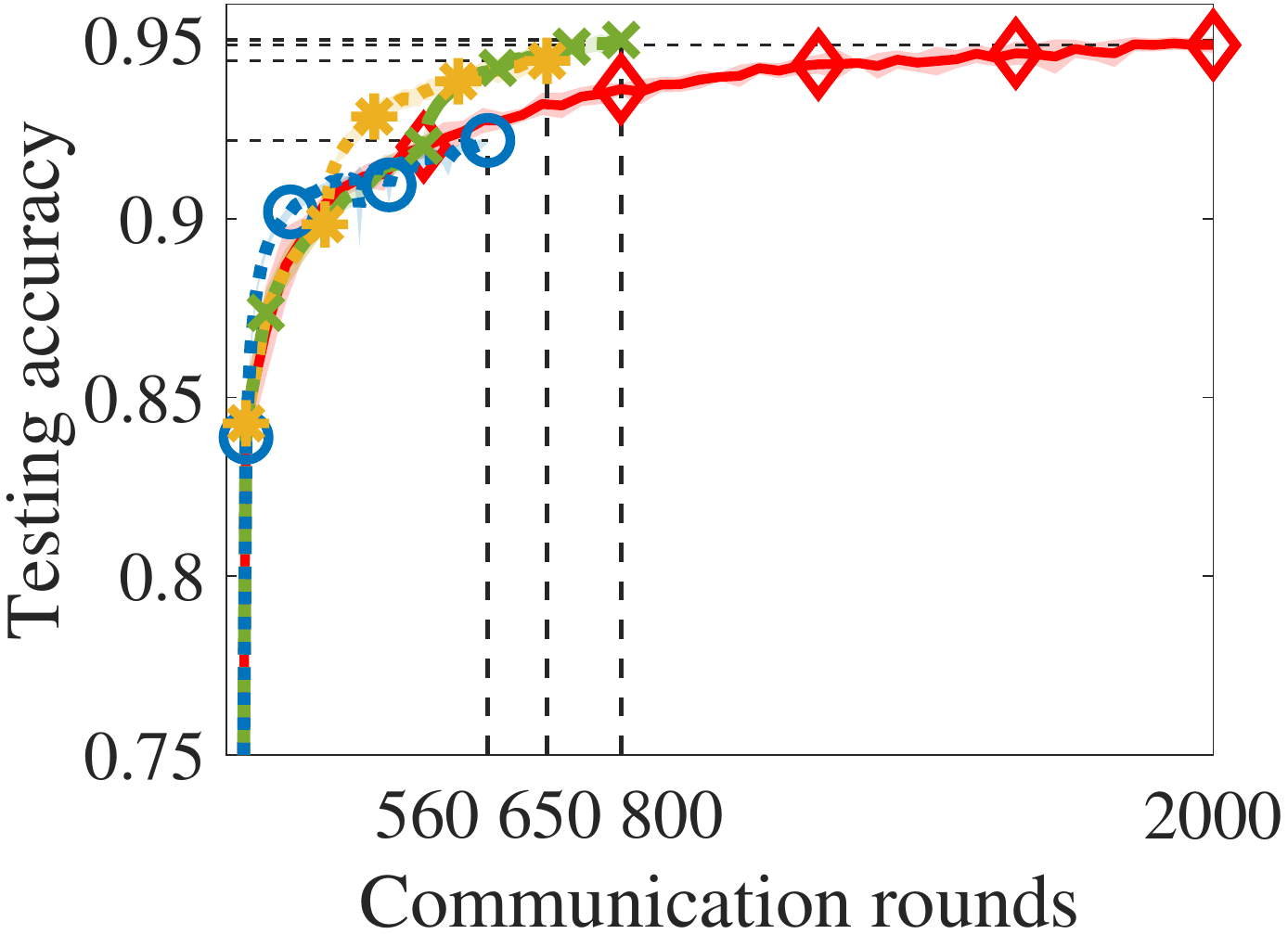}}
\caption{Testing accuracy of \texttt{FedTAN-II} on MNIST dataset.}
\label{fig:performance FedTAN-II MNIST}
\end{figure}

In this section, we demonstrate the effectiveness of \texttt{FedTAN-II} when applied to the MNIST dataset.
Fig. \ref{fig:performance FedTAN-II MNIST}(a) compares \texttt{FedTAN-II} with \texttt{FedTAN} by training a 3-layer DNN with BN on the MNIST dataset. The comparison is based on iterations.
Furthermore, Fig. \ref{fig:performance FedTAN-II MNIST} (b) compares them in terms of communication rounds.
For the \texttt{FedTAN-II} approach, we initialize the learning rate $\gamma$ at 0.5 for the first $M$ iterations, following the parameter setting of \texttt{FedTAN} in Section \ref{sec:Parameter Setting}.
Subsequently, the value of $\gamma$ is reduced to 0.05 after $M$ iterations.

As observed in Fig. \ref{fig:performance FedTAN-II MNIST}, a reduction in the value of $M$ results in a marginal decrease in accuracy, especially if $M$ falls below a certain value (e.g., $M \leq 10$ in Fig. \ref{fig:performance FedTAN-II MNIST}(a)).
This slight accuracy reduction is primarily attributed to less precise statistical parameters.
However, it is noteworthy that employing a smaller value for $M$ considerably diminishes the requisite communication rounds, as depicted in Figure \ref{fig:performance FedTAN-II MNIST}(b).

Next, In Table \ref{table:performance FedTAN-II MNIST}, we contrast the aggregate exchanged bits, the proportions of supplementary communication rounds, and the exchanged bits resulting from layer-aggregations in Algorithms \ref{algorithm:FedTAN modification forward} and \ref{algorithm:FedTAN modification backward} throughout the complete training procedure of \texttt{FedTAN-II} across different $M$ values.
As observed in Table \ref{table:performance FedTAN-II}, Table \ref{table:performance FedTAN-II MNIST} further illustrates the insubstantial variance in total exchanged bits between \texttt{FedTAN} and \texttt{FedTAN-II} across varying $M$ values, owing to the minor proportion occupied by statistical parameters in BN layers within the total model size.
Furthermore, while the additional communication rounds resulting from layer-aggregations in Algorithms \ref{algorithm:FedTAN modification forward} and \ref{algorithm:FedTAN modification backward} contribute to a specific proportion of the total exchanged bits throughout the training process, particularly evident with larger values of $M$, the accompanying extra exchanged bits remain negligible.

\begin{table}[h]
\centering
\caption{Communication overhead of \texttt{FedTAN-II} on MNIST dataset.}\label{table:performance FedTAN-II MNIST}
\begin{tabular}{c|c|cc}
\hline
\multirow{2}{*}{\textbf{FL scheme}} & \multirow{2}{*}{\textbf{Exchanged bits (GB)}} & \multicolumn{2}{c}{\textbf{Percentage of extra}}   \\
& & \textbf{Rounds} & \textbf{Bits} \\ \hline
\texttt{FedTAN-II} ($M$$=$10) & 2.1388 & 5.66\% & 0.0100\% \\
\texttt{FedTAN-II} ($M$$=$50) & 2.1397 & 23.08\% & 0.0501\% \\
\texttt{FedTAN-II} ($M$$=$100) & 2.1408 & 37.50\% & 0.1002\% \\
\texttt{FedTAN}                 & 2.1386 & 75\% & 0.4992\% \\ \hline
\end{tabular}
\end{table}




\end{document}